\documentclass[10pt,journal,compsoc]{IEEEtran}
\usepackage{booktabs}
\usepackage{graphicx}
\usepackage{amsfonts}
\usepackage{caption}
\usepackage{epstopdf}
\usepackage{epsfig}
\usepackage{cite}
\usepackage{amsmath}
\usepackage{color}
\usepackage{algorithm}
\usepackage{algorithmic}
\usepackage{url}
\usepackage{amssymb}
\usepackage{subfigure}

\newcommand{\norm}[1]{\left\Vert#1\right\Vert}
\newcommand{\abs}[1]{\left | #1 \right |}
\newcommand{\pref}[1]{(\ref{#1})}

\ifCLASSINFOpdf

\else

\fi

\begin{document}

\title{High-Dimensional Data Set Simplification by Laplace-Beltrami Operator}


\author{Chenkai Xu, Hongwei Lin, \IEEEmembership{Member, IEEE,}%
\IEEEcompsocitemizethanks{\IEEEcompsocthanksitem C. Xu and H. Lin are with School of Mathematics, State Key Lab. of CAD\&CG, Zhejiang University.}
\thanks{Email: hwlin@zju.edu.cn}}

%

\markboth{IEEE Transactions on Knowledge and Data Engineering,~Vol
.~XX, No.~XX,~XX~XXXX}{Shell \MakeLowercase{\text
it{et al.}}: A Novel Tin Can Link}

\IEEEtitleabstractindextext{%
\begin{abstract}
  With the development of the Internet and other digital technologies,
    the speed of data generation has become considerably faster than the speed of data processing.
  Because big data typically contain massive redundant information,
    it is possible to significantly simplify a big data set while maintaining the key information it contains.
  In this paper, we develop a big data simplification method based on
    the eigenvalues and eigenfunctions of the Laplace-Beltrami operator (LBO).
  Specifically,
    given a data set that can be considered as an unorganized data point set in high-dimensional space,
    a discrete LBO defined on the big data set is constructed
    and its eigenvalues and eigenvectors are calculated.
  Then, the local extremum and the saddle points of the
    eigenfunctions are proposed to be the feature points of a data set in high-dimensional space,
    constituting a simplified data set.
  Moreover, we develop feature point detection methods for the
    functions defined on an unorganized data point set in high-dimensional space,
    and devise metrics for measuring the fidelity of the simplified data set to the original set.
  Finally, examples and applications are demonstrated to validate
    the efficiency and effectiveness of the proposed methods,
    demonstrating that data set simplification is a method for processing
    a maximum-sized data set using a limited data processing capability.
\end{abstract}

}

\maketitle

\IEEEdisplaynontitleabstractindextext

\IEEEpeerreviewmaketitle

\section{Introduction}

 It is well known that big data concern large-volume, complex,
    and growing data sets with multiple sources~\cite{Xindong2013Data}.
 With the development of the Internet and other digital technologies,
    increasingly larger amounts of data are generated at a speed
    considerably greater than the data processing speed of human beings.
 Because big data typically contain massive redundant information,
    it is possible to significantly simplify a big data set while maintaining the key information it contains.
 Therefore, data set simplification is a feasible method to process
    a maximum-sized data set using a limited
    data processing capability.

 The purpose of big data simplification is to  maintain the maximum amount of
    information a big data set contains,
    while deleting the maximum amount of redundant information.
 In this paper, a big data set is considered as an unorganized point set with
    points in a high-dimensional space.
 Based on this consideration, we developed a data set simplification method by
    classifying the points in a big data set as \emph{feature} points and \emph{trivial} points,
    and then considering the feature point set as the simplification of the big data set
    because it contains a majority of the information that the original
    big data set contained.

 Feature point detection is a fundamental problem in numerous research fields
    including computer vision and computer graphics.
 These fields focus on the process of  one-dimensional (1D) and two-dimensional (2D) manifolds,
    such that the main tool for the feature point detection is the curvature,
    including both Gauss and mean curvatures.
 However, in high-dimensional space,
    the definition of curvature is complicated
    and practical computational methods for calculating the curvature of
    a high-dimensional manifold are not available.
 Therefore, it is necessary for the big data process to develop practical and
    efficient methods for detecting feature points in big data sets.

 It is well known that the eigenvalues and eigenvectors of a Laplace-Beltrami
    operator (LBO) defined on a manifold are closely related to its geometry properties.
 Moreover, whatever the dimension of the manifold,
    after discretization,
    the LBO (called a discrete LBO) becomes a matrix.
 That is, the discrete LBO is unrelated to the dimension of the
    manifold where it is defined.
 Therefore, the eigenvalues and eigenvectors of LBOs
    and the related heat kernel functions are desirable tools for studying the geometric properties of big data sets in high-dimensional space,
    including the detection of feature points.

 In this paper, we developed a data simplification method for high-dimensional big data sets
    based on feature point detection
    using the eigenvectors of LBO.
 Moreover, three metrics are devised for measuring the fidelity of the
    simplified data set to the original data set.
 Finally, the capability of the proposed big data simplification method is
    extensively discussed using several examples and applications.

 This paper is organized as follows:
  In Section~\ref{sec:related_work},
   related work is briefly reviewed,
   including the calculations and applications of eigenvalues and eigenvectors of LBO,
   feature point detection methods in computer graphics and computer vision,
   and sampling techniques on high-dimensional data.
 In Section~\ref{sec:simplification_method},
    after introducing the technique for calculating the
    eigenvalues and eigenvectors of LBO on high-dimensional big data sets,
    we develop the data simplification method by detecting feature points of the big data sets,
    and propose three metrics to measure the fidelity of the simplified data set to the original data set.
 In Section~\ref{sec:discussion_results},
    the capabilities of the proposed data simplification method
    are illustrated and discussed.
 Finally, the paper is concluded in Section~\ref{sec:conclusion}.

\section{Related work}
\label{sec:related_work}

 In this section,
    we review the related work in three aspects.
 First, we address the computation techniques and applications of
    eigenvalues and eigenvectors of LBO in dimensionality reduction (DR),
    shape descriptors,
    and mesh saliency detection.
 Secondly, the feature point detection methods are introduced.
 Finally,
    data sampling techniques on machine learning and deep learning are discussed.

 \textbf{Computation and applications of LBO.}
 To calculate the eigenvalues and eigenfunctions of an LBO defined on
    a triangular mesh or big data set,
    it must first be discretized.
 One commonly employed technique for the discretization of an LBO is the
    cotangent Laplacian~\cite{UlrichPinkall1993Computing},
    which requires only the one-ring neighbors of a vertex to construct the discrete LBO at the vertex.
 Although the cotangent Laplacian has intuitive geometric meaning,
    it does not converge in generic cases~\cite{Sun2009A}.
 In 2008, Belkin et al. proposed the mesh Laplacian
     scheme~\cite{Belkin2008Discrete},
     where the Euclidean distance between each pair of the complete data set is required for the calculation of the matrix weight.
 Furthermore, by discretizing the integration of certain continuous functions
    defined over the manifold,
    a symmetrizable discrete LBO was developed to ensure its eigenvectors were orthogonal~\cite{Liu2012Point}.

 The LBO has been extensively employed in mesh processing in the fields of
    computer graphics and computer vision.
 Based on the spectral analysis of the graph Laplacian,
    Taubin et al.~\cite{Taubin1995A} designed low-pass filters for mesh smoothing.
 Moreover,
    by first modulating the eigenvalues and eigenvectors of an LBO on a mesh
    and then reconstructing the shape of the mesh model,
    mesh editing tasks can be fulfilled, including
    mesh smoothing,
    shape detail enhancement,
    and change of model posture~\cite{Levy2006Laplace, Vallet2008Spectral}.

 Using the information in the spectral domain of the LBO,
    the mesh saliency points or regions can be detected.
 By defining a geometric energy related to the eigenvalues of an LBO,
    Hu et al. proposed a method to extracts the salient geometric feature points~\cite{Hu2009Salient}.
 Considering the properties of the log-Laplacian spectrum of a mesh,
    Song et al. developed a method for detecting mesh saliency by capturing the saliency in the frequency domain~\cite{Song2014Mesh}.

 The eigenvalues and eigenvectors of an LBO have been used in shape analysis
    and shape retrieval.
 Reuter et al.\cite{Reuter2006Laplace} used the sequence of eigenvalues of an
    LBO defined on a mesh as a numerical fingerprint or signature of the mesh,
    and called this the \emph{Shape DNA},
    which has been successfully applied in fields such as shape retrieval,
    copyright protection,
    and mesh quality assessment.
 Using the concept of heat kernel,
    which contains the information for the eigenvalues and eigenvectors of the LBO defined on a mesh,
    Sun et al.~\cite{Sun2009A} proposed the \emph{heat kernel signature} (HKS) for shape analysis.
 With the intrinsic invariant property of HKS,
     Ovsjanikov~\cite{Ovsjanikov2010One} proposed a one-point isometric matching algorithm to determine the intrinsic symmetries of shapes.
 Bronstein~\cite{Bronstein2010Scale} developed the scale-invariant
    heat kernel signatures(SI-HKS) for non-rigid shape recognition,
    based on a logarithmically sampled scale-space.

 More importantly, the LBO has been successfully applied in high-dimensional data processing.
 A well-known application of the LBO in high-dimensional data processing is Laplacian eigenmaps~\cite{Belkin2003Laplacian}.
 To appropriately represent high-dimensional data for machine learning and
      pattern recognition,
 Belkin et al.\cite{Belkin2003Laplacian} proposed a locality-preserving
    method for nonlinear DR using the LBO eigenfunctions,
    which has a natural connection to clustering,
    and can be computed efficiently.

 \textbf{Feature point detection.}
 Following the presentation of scale invariant feature
    transform(SIFT)~\cite{Lowe2004Distinctive} and histogram of oriented
    gradients(HOG)~\cite{Dalal2005Histograms},
    a series of feature point detection methods were proposed using the Difference of Gaussians(DoG)~\cite{Lowe2004Distinctive}.
 Castellani et al.~\cite{Castellani2010Sparse} designed a 3D saliency measure
    basing on DoG,
    which allows a small number of sparse salient points to characterize distinctive portions of a surface.
 Zou et al.\cite{Hua2008Surface} defined the DoG function over a curved
    surface in a geodesic scale space
    and calculated the local extrema of the DoG,
    called saliency-driven keypoints,
    which is robust and stable to noisy
    input~\cite{Johnson2002Using, Kortgen20033D, Schlattmann2008Scale}.

 Similarly, geometric function-based methods have been proposed.
 Two widely known surface descriptors based on surface geometry are 3D spin images~\cite{Johnson2002Using} and 3D shape contexts~\cite{Kortgen20033D}.
 Schlattmann et al.\cite{Schlattmann2008Scale} developed a novel
    scale space generalization to detect surface features
    based on the averaged mean curvature flow.

 Another kind of feature point detection approach is based on geometric
    diffusion equation.
 Heat kernel signature (HKS) can be used for feature point detection,
    which discerns the global feature and local feature points by adjusting
    the time parameter~\cite{Sun2009A}.
 Moreover, by integrating the eigenvalues into the eigenvectors of an LBO on
    a mesh,
    global point signature (GPS) was developed in~\cite{Rustamov2007Laplace} for feature point detection.
 For 2D surface quadrangulation,
    Shen et al.~\cite{Shen2006Spectral} used the extremes of the Laplacian eigenfunctions to build a Morse-Smale complex
    and construct a well-shaped quadrilateral mesh.

 In recent years,
    deep networks\cite{Lecun1995Convolutional} have achieved excellent success in feature point detection~\cite{Wang2017Facial}.
 Luo et al.~\cite{Luo2012Hierarchical} proposed a novel face parser approach
    by detecting faces at both the part- and component-levels,
    and then computing the pixel-wise label maps,
    where the feature points can then be easily obtained from the boundary of the label maps.
 Sun et al.~\cite{Sun2013Deep} presented an approach for detecting the
    positions of face keypoints with three-level carefully designed convolution networks.
 In addition to convolutional neural networks(CNN),
  recurrent neural networks(RNN) have also been developed on feature point detection,
  especially on facial feature point detection~\cite{Peng2018RED, Wang2017Facial, Xiao2016Robust}.

\textbf{Data sampling.}
 In the early machine learning field,
   to solve the imbalanced data problem in classification,
   over-sampling and under-sampling approaches were proposed.
 Tomek et al.~\cite{tomek1976two} proposed the 'Tomek links',
   which is computed by two examples belonging to different classes.
 'Tomek links' can determine if one of two examples is noise or borderline.
 One noted over-sampling method is SMOTE~\cite{chawla2002smote},
   which creates synthetic minority class examples by interpolating
   among several minority class examples that lie together.
 Hybrid methods have been proposed based on the above two methods,
   including 'SMOTE-Tomek links'~\cite{batista2004study},
   'Boderline-SMOTE'~\cite{bunkhumpornpat2009safe}
   and 'Safe-Level-SMOTE'~\cite{han2005borderline}.

 In the deep learning field,
   to reduce annotation effort and clean the data before training,
   following data under-sampling methods were proposed.
 In biomedical imaging applications,
   Zhou et al.\cite{zhou2018aft} presented AFT* to seek 'worthy' samples for annotation,
   which reduces the annotation cost by half compared with random selection.
 Vodrahalli et al.~\cite{vodrahalli2018all} used
   the gradient in stochastic gradient descent (SGD)
   to measure the importance of training data,
   demonstrating that a small subsample is indeed sufficient for training in certain cases.

\section{High-dimensional data set simplification}
\label{sec:simplification_method}

 In this section,
    we develop the high-dimensional data set simplification method,
    which first detects the feature points in the high-dimensional data set,
    and then uses the feature point set as the simplified data set.
 Moreover, three metrics are proposed for measuring the fidelity of the
    simplified data set to the original data set.

 Feature point detection on 1D and 2D manifolds typically depends on the
    curvature information.
 However, in high-dimensional space,
    the definition and computation of the curvature is complicated.
 Hence, feature point detection using curvature information on high-dimensional
    data sets is infeasible.

 As stated above, whatever the dimension of a space is,
    the discrete LBO defined on the space is a matrix.
 Because the eigenvectors of a discrete LBO are closely
    related to the geometric properties of the space where the LBO is defined,
    and because there are sophisticated methods for computing the eigenvectors of a matrix (which is a discrete LBO),
    the discrete LBO and its eigenvectors are desirable tools for feature point detection in high-dimensional data sets.


\subsection{Spectrum of LBO and Feature Points}
\label{subsec:spectrum_feature_point}

 Let $M$ be a Riemannian manifold (differentiable manifold with
    Riemannian metric),
   and $f\in C^{2}$ be a real-valued function defined on $M$.
 The LBO $\Delta$ on $M$ is defined as
 \begin{equation} \label{eq:lbo_org}
  \Delta f = div(grad\ f),
 \end{equation}
    where  $grad\ f$ is the gradient of $f$,
    and $div(\cdot)$ is the divergence on $M$.
 The LBO~\pref{eq:lbo_org} can be represented in local coordinates.
 Let $\psi$ be a local parametrization
 \[
    \psi : \mathbb{R}^{n}\rightarrow\mathbb{R}^{n+k},
 \]
  of a sub-manifold of $M$ with
  \begin{equation*}
  \begin{split}
  & g_{ij} = \langle \partial_{i}\psi , \partial_{j}\psi\rangle,
  \quad G = (g_{ij}), \\
  & (g^{ij})=G^{-1} ,
  \quad W=\sqrt{det G},
  \end{split}
  \end{equation*}
  where $i,j = 1,2, ...,n$ ,
  $\langle , \rangle$ is the dot product,
  and $det$ is the determinant.
 In the local coordinates,
    the LBO~\pref{eq:lbo_org} can be transformed as,
  \begin{equation}\label{eq:lbo_local}
  \Delta f = \frac{1}{W}
  \sum_{i,j} \partial_{i}(g^{ij}W\partial_{j}f).
  \end{equation}

 The eigenvalues and eigenfunctions of the
    LBO~\pref{eq:lbo_org}~\pref{eq:lbo_local} can be calculated by solving the Helmholtz equation:
  \begin{equation}\label{eq:helmholtz}
  \Delta f = -\lambda f,
  \end{equation}
  where $\lambda$ is a real number.
  The spectrum of LBO is a list of eigenvalues such as
  \begin{equation}\label{eq:eigen_value}
    0\leq\lambda_{0}\leq\lambda_{1}\leq\lambda_{2}\leq...\leq+\infty{},
  \end{equation}
  with the corresponding eigenfunctions,
  \begin{equation}\label{eq:eigen_function}
        \phi_0(x), \phi_1(x), \phi_2(x), \cdots,
  \end{equation}
    which are orthogonal.

 In the case of a closed manifold,
    the first eigenvalue $\lambda_{0}$ is always zero.
 Moreover, the eigenvalues $\lambda_i, i=0,1,2,\cdots$~\pref{eq:eigen_value}
    specify the discrete frequency domain of an LBO,
    and the eigenfunctions are the extensions of the basis functions in Fourier analysis to a manifold.
 Eigenfunctions of larger eigenvalues contain higher frequency information.
 It is well known that the eigenvalues and eigenfunctions of an LBO are
    closely related to the geometric properties of the manifold where the LBO is defined,
    including its curvature~\cite{Benko1979Eigenvalues},
    volume of $M$,
    volume of $\partial{M}$,
    and Euler characteristic of $M$~\cite{Reuter2006Laplace}.

 Specifically, consider a heat diffusion equation defined on a
    closed manifold $M$ (i.e., $\partial M = {\O}$):
 \begin{equation} \label{eq:diffusion_equation}
    \left\{
         \begin{array}{lr}
                (\Delta+ \frac{\partial}{\partial t})u(x,t) = 0,      \\
                 u(x,0)=f(x),     \\
         \end{array}
    \right.
 \end{equation}
 where $x \in M$,
    $u(x,t)$ is the heat value of the point $x$ at time $t$,
    and $f$ is the initial heat distribution on $M$.
 The solution of the heat diffusion equation~\pref{eq:diffusion_equation} can
    be expressed as:
 \begin{equation*}\label{}
              u(x,t) = \int_{M} H(x,y,t)f(y) dy
 \end{equation*}
 where, $H(x,y,t)$ is the fundamental solution of the heat diffusion
    equation~\pref{eq:diffusion_equation},
    called \emph{heat kernel}.
 The heat kernel $H(x,y,t)$ measures the amount of heat transferred from
    $x$ to $y$ within time $t$,
    and can be rewritten as,
 \begin{equation}\label{eq:heat_kernel}
    H(x,y,t)=\sum_{i=0}^{\infty} e^{-\lambda_{i}t}\phi_{i}(x)\phi_{i}(y),
 \end{equation}
 where $\lambda_i, i=0,1,\cdots$~\pref{eq:eigen_value} are the eigenvalues
    of the LBO~\pref{eq:lbo_org}~\pref{eq:lbo_local},
    and $\phi_{i}, i=0,1,\cdots$~\pref{eq:eigen_function} are the corresponding eigenfunctions.

 When $x = y$ in Eq.~\pref{eq:heat_kernel},
    $H(x,x,t)$ , called \emph{HKS},
    quantifies the change of heat at $x$ along the time $t$.
 It has been shown that~\cite{Pleijel1949Some, Berger1971Le, Benko1979Eigenvalues},
    when the time $t$ approaches $0^+$ sufficiently close,
    the heat kernel $H(x,x,t)$ can be represented as,
 \begin{equation}\label{eq:kernel_curvature}
           H(x,x,t) = \sum_{i=0}^{\infty} e^{-\lambda_{i}t}\phi_{i}^{2}(x)
           = \frac{1}{4\pi t}+\frac{K(x)}{12\pi}+O(t),
 \end{equation}
 where $K(x)$ is the scalar curvature at point $x$ on $M$.
 Especially, when $M$ is a 2-dimensional manifold,
    $K(x)$ is the Gaussian curvature at $x$.


 From Eq.~\pref{eq:kernel_curvature},
    we can see that,
    the eigenfunctions $\phi_i(x),i=0,1,\cdots$~\pref{eq:eigen_function} of a LBO are closely related to the scale curvature $K(x)$ at the point $x \in M$.

\begin{figure}[!htb]
   \centering
     \includegraphics[width=0.48\textwidth]{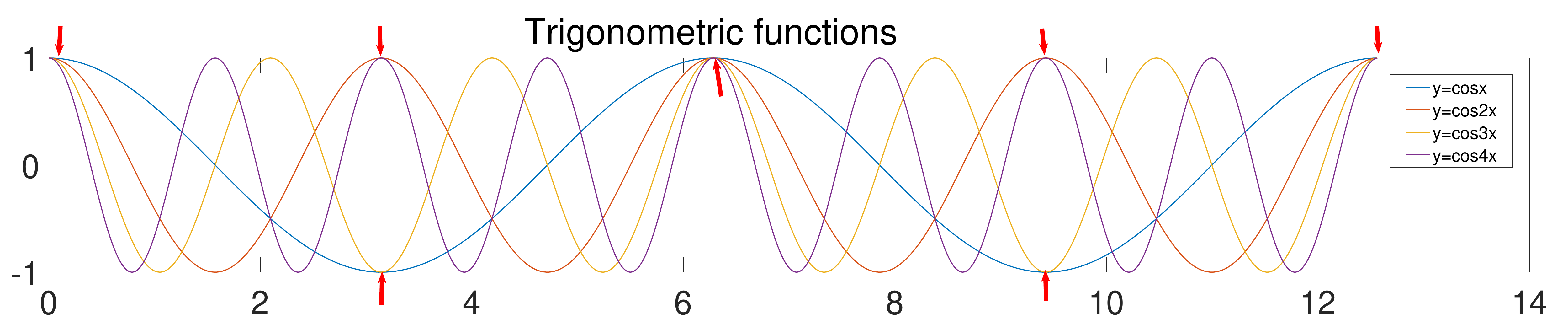}
   \caption
   {
        Local extrema (indicated by red arrows) of trigonometric functions with lower frequencies hold in those with higher frequencies.
    }
    \label{fig:trigonometric functions}
\end{figure}

\begin{figure}[!htb]
      \centering
      \includegraphics[width=0.45\textwidth]{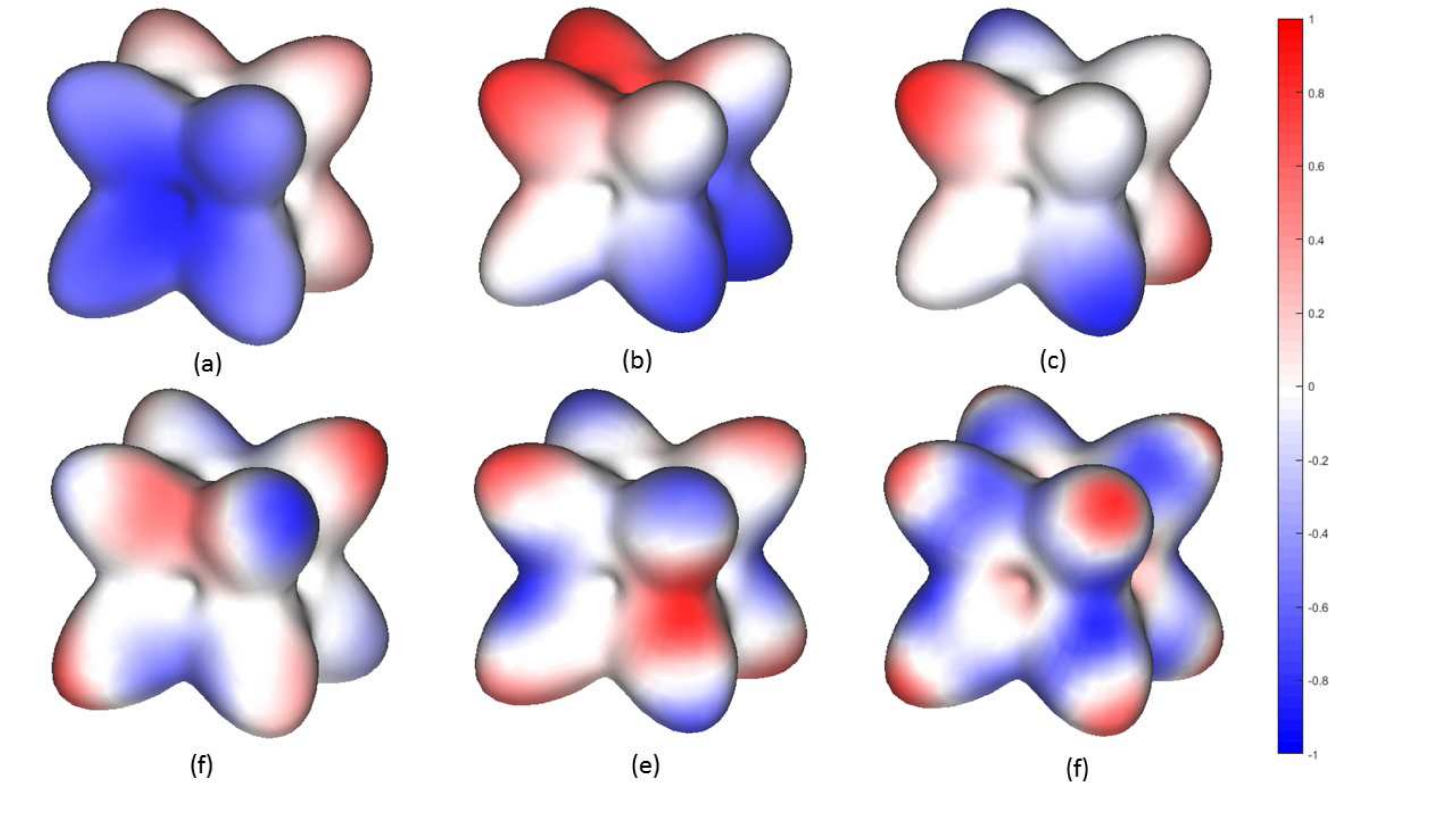}
      \caption
      {
        Local extrema (the centres of blue and red regions) of the eigenvectors~\pref{eq:eigen_function} of a discrete LBO defined on a 2D manifold.
        (a)-(f) Eigenvectors $\phi_2(x)$, $\phi_3(x)$, $\phi_6(x)$, $\phi_{15}(x)$, $\phi_{19}(x)$, $\phi_{31}(x)$.
      }
      \label{fig:cube8}
\end{figure}

 As stated above,
    the eigenfunctions $\phi_i(x),i=0,1,\cdots$~\pref{eq:eigen_function}
    are the extension of the trigonometric functions.
 In the trigonometric function system,
    the local extrema of a trigonometric function with a lower frequency
    hold in the trigonometric function with a higher frequency (Fig.~\ref{fig:trigonometric functions}).
 This property is preserved by the eigenfunctions
    $\phi_i(x),i=0,1,\cdots$~\pref{eq:eigen_function} on manifolds.
 In Fig.~\ref{fig:cube8},
    eigenfunctions of an LBO defined on a 2D manifold
    are illustrated with their local extrema.
 We can observe that,
    the local extrema in the lower frequency eigenfunctions
     remain the local extrema in the higher frequency eigenfunctions (Fig.~\ref{fig:cube8}).
 Because the local extrema of $\phi_i(x)$ hold in $\phi_j(x), j < i$,
    and the heat kernel~\pref{eq:kernel_curvature} is a weighted sum of $\phi_i^2(x), i=0,1,\cdots$,
    the local extrema of $\phi_i(x)$,
    especially the eigenfunctions with low frequency,
    are the candidates of the local extrema of the heat kernel $H(x,x,t)$~\pref{eq:kernel_curvature},
    and the scale curvature $K(x)$~\pref{eq:kernel_curvature}.
 In conclusion, the local extrema of the eigenfunctions
    $\phi_i(x), i=0,1,\cdots$ contain the features of the manifold where the LBO is defined.
 Therefore, in this paper,
    we use the local extrema of the eigenfunctions of the LBO defined
    on a high-dimensional data set as its \emph{feature points},
    which constitute the simplified data set.

\subsection{Discrete Computation}
\label{subsec:discrete_computation}

 To solve the Helmholtz equation~\pref{eq:helmholtz} for calculating the
    eigenvalues and eigenfunctions of the LBO~\pref{eq:lbo_local},
    the Helmholtz equation~\pref{eq:helmholtz} must be discretized into a linear system of equations.
 Traditionally, Eq.~\pref{eq:helmholtz} is discretized on a mesh with
    connectivity information between two mesh vertices~\cite{Belkin2008Discrete,Belkin2003Laplacian,UlrichPinkall1993Computing}.
 However,
    in this paper, we are addressing an unorganized data point set in
    a high-dimensional space,
    where the connectivity information is missing.
 Hence, we use the discretization method developed
    in~\cite{Belkin2003Laplacian} to construct the discrete LBO.
 Because only the Euclidean distances between points are involved in the
    construction of the discrete LBO,
    it can be employed for LBO discretization in high-dimensional space.

 The LBO discretization method~\cite{Belkin2003Laplacian} contains
    the two steps:
    \begin{itemize}
      \item  adjacency graph construction,
      \item  weight computation,
    \end{itemize}
    which are explained in detail in the following.

 \textbf{Adjacency graph construction.}
  There are two methods for constructing an adjacency graph:
   (a) $k$ nearest neighbors (\emph{KNN}).
  For any vertex $v$, its $k$ nearest neighbors $v_i, i=1,2,\cdots,k$ are
    determined using the \emph{KNN} algorithm~\cite{muja2009flann}.
  Then, the connection between the vertex $v$ and each neighbor
    $v_i, i=1,2,\cdots,k$ is established,
    thus constructing the adjacent graph.
  Because the $k$ nearest neighbors are not symmetric,
    i.e., it is possible that the vertex $v_i$ is in the $k$ nearest neighbors of the vertex $v_j$,
    yet, $v_j$ is not in the $k$ nearest neighbors of $v_i$,
    the weight matrix (see below) constructed by the \emph{KNN} method is not guaranteed to be symmetric.
    (b) $\epsilon$ -neighbors.
  For a vertex $v$,
    a hypersphere is first constructed,
    which uses $v$ as the center,
    and $\epsilon$ as its radius.
  The adjacency graph is constructed by connecting the vertex $v$ and
    each vertex lying in the hypersphere.
  Although the weight matrix constructed based on the
    $\epsilon$ -neighbors is symmetric,
    there is a serious problem in that the value of $\epsilon$ is difficult to choose.
  An overly small $\epsilon$ will cause certain rows of the weight matrix $W$
    to be zero;
       and overly large $\epsilon$ will create a matrix that is excessively dense.
  Therefore, in our implementation,
    we employ the $k$ nearest neighbors to construct the adjacency matrix.

  \textbf{Weight computation.}
  Using the adjacency graph for each point $v_i$ in the given data set,
    the connections between point $v_i$ and the other points in the data set are established.
  Based on the adjacency graph,
    the weights between point $v_i$ and each of the other points can be calculated as follows:
     \begin{equation}\label{eq:weight_element}
      w_{ij}=
     \left\{
             \begin{array}{lr}
             -e^{-\frac{\norm{v_{i}-v_{j}}^2}{t}
                },      &$if $i,j$ are adjacent,$ \\
              \sum_{k \neq i} -w_{ik} ,     &$if $ i = j, \\

              0,   &$otherwise$.\\
             \end{array}
    \right.
    \end{equation}
  Because only the Euclidean distance $\norm{v_{i}-v_{j}}$ is required in
    the computation of $w_{ij}$,
    it is appropriate for addressing high-dimensional data sets.

  As stated above, the weight matrix $\bar{W}=[w_{ij}]$
    \pref{eq:weight_element} constructed by the \emph{KNN} method is not symmetric,
    so we take the following matrix:
    \[
        W = \frac{\bar{W} + \bar{W}^T}{2}
    \]
    as the weight matrix,
    which is a symmetric matrix.
  Moreover, we construct a diagonal matrix
  \[
        A = diag(a_1, a_2, \cdots, a_n),
  \]
    where $a_i = w_{ii}$ (refer to Eq.~\pref{eq:weight_element}).
  Then the LBO can be discretized into the matrix,
  \[
        L = A^{-1} W,
  \]
  whatever the dimension of the space the LBO defined on.
  Meanwhile, the Helmholtz equation~\pref{eq:helmholtz} is discretized as,
  \[
        L \varphi = A^{-1} W \varphi = \lambda \varphi,
  \]
    where $\lambda$ is the eigenvalue of the Laplace matrix $L$,
    and $\varphi$ is the corresponding eigenvector.
  It is equivalent to,
  \begin{equation}\label{eq:discrete_helmholtz}
    W \varphi =\lambda A \varphi.
  \end{equation}
  Because the matrix $W$ is symmetric,
    and the matrix $A$ is diagonal,
    it has been shown that the eigenvalues $\lambda$ are all non-negative real numbers satisfying ~\cite{Sun2009A}:
    \[
        0=\lambda_1 \leq \lambda_2 \leq \cdots \leq \lambda_n,
    \]
  and the corresponding eigenvectors $\varphi_i, i=1,2,\cdots,n$
    are orthogonal,
  i.e.,
  \[
    <\varphi_i, \varphi_j> = \varphi_i^T \varphi_j = 0,\ i \neq j.
  \]

%
%

\subsection{Data Simplification by Feature Point Detection}
\label{subsec:data_simplification}

 By solving Eq.~\pref{eq:discrete_helmholtz},
    the eigenvalues $\lambda_1 \leq \lambda_2 \leq \cdots \leq \lambda_n$
    and the corresponding eigenvectors $\varphi_i, i=1,2,\cdots,n$ are determined.
 The local maximum, minimum, and saddle points of the eigenvectors
    are used as the feature points.
 As stated above, the eigenvalue $\lambda_i$ measures the 'frequency' of
    its corresponding eigenvector $\varphi_i$.
 The larger the eigenvalue,
    the higher the frequency of the eigenvector $\varphi_i$,
    and then the greater number of feature points in the eigenvector $\varphi_i$.
 The data simplification algorithm first detects the feature points of the
    eigenvector $\varphi_1$,
    and adds them to the simplified data set.
 Then, it detects and adds the feature points of $\varphi_2$ to
    the simplified data set.
 This procedure is performed iteratively
    until the simplified data set satisfies a preset fidelity threshold to the original data point set,
    according to the metrics defined in Section~\ref{subsec:metrics}.

 Because each of the eigenvectors $\varphi_i, i=1,2,\cdots,n$ is a scalar
    function defined on an unorganized point set in a high-dimensional space,
    the detection of the maximum, minimum,
    and saddle points on the defined function is not straightforward.
 Suppose $\omega(x)$ is a scalar function defined on each point of
    an unorganized data point set.
 For each point $x$ in the data set,
    we construct its neighbor $N_x$ using the \emph{KNN} algorithm,
    discussed previously in Section3.2.
 The methods for detecting the maximum, minimum,
    and saddle points of the function $\omega(x)$ are elucidated in the following.

 \textbf{Maximum and minimum point detection:}
  For a data point $x$ in the data set
    and any data point $y \in N_x$,
    if $\omega(y) < \omega(x)$, $\forall y \in N_x$,
    the data point $x$ is a local maximum point of $\omega(x)$.
  Conversely, if $\omega(y) > \omega(x)$,  $\forall y \in N_x$,
    the data point $x$ is a local minimum point of $\omega(x)$.

\begin{figure}[!htbp]
   \begin{minipage}[t]{0.5\textwidth}
   \centering
   \subfigure[]
   {
        \includegraphics[width=0.38\textwidth]
            {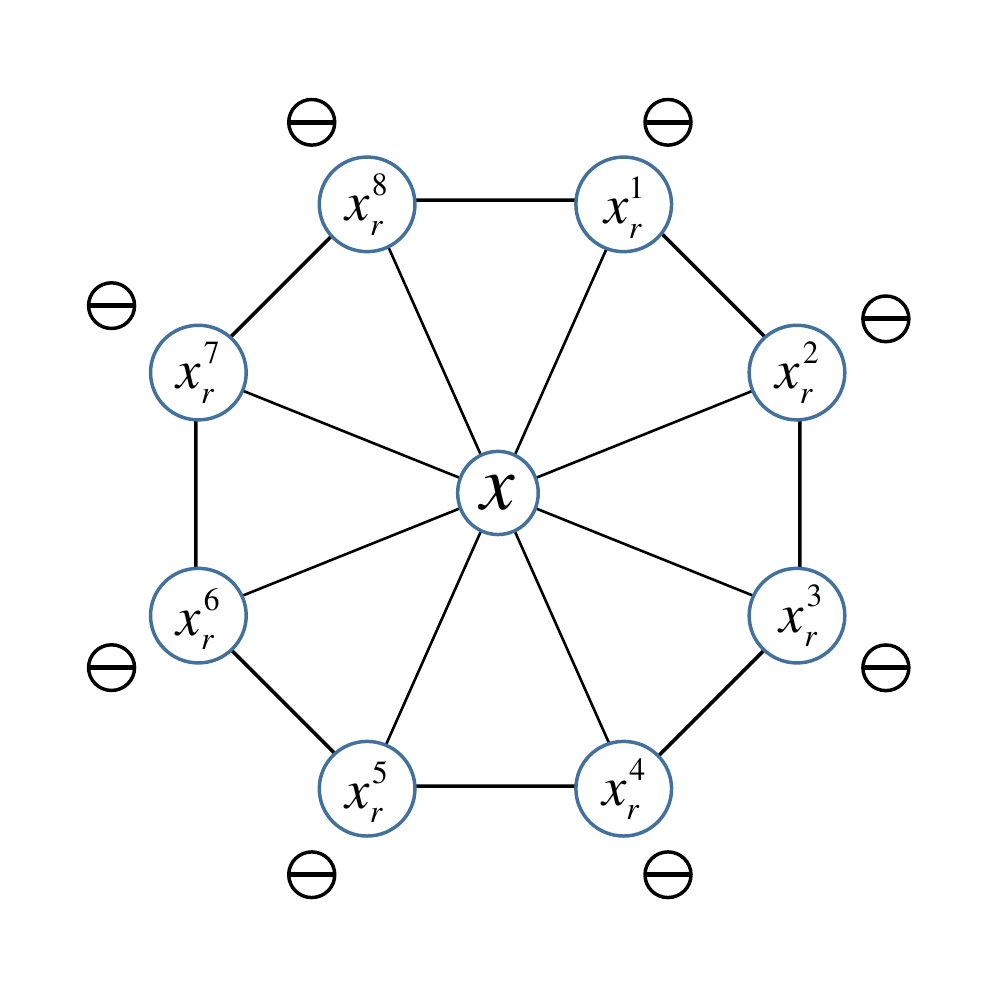}
   }
   \subfigure[]
   {
        \includegraphics[width=0.38\textwidth]
            {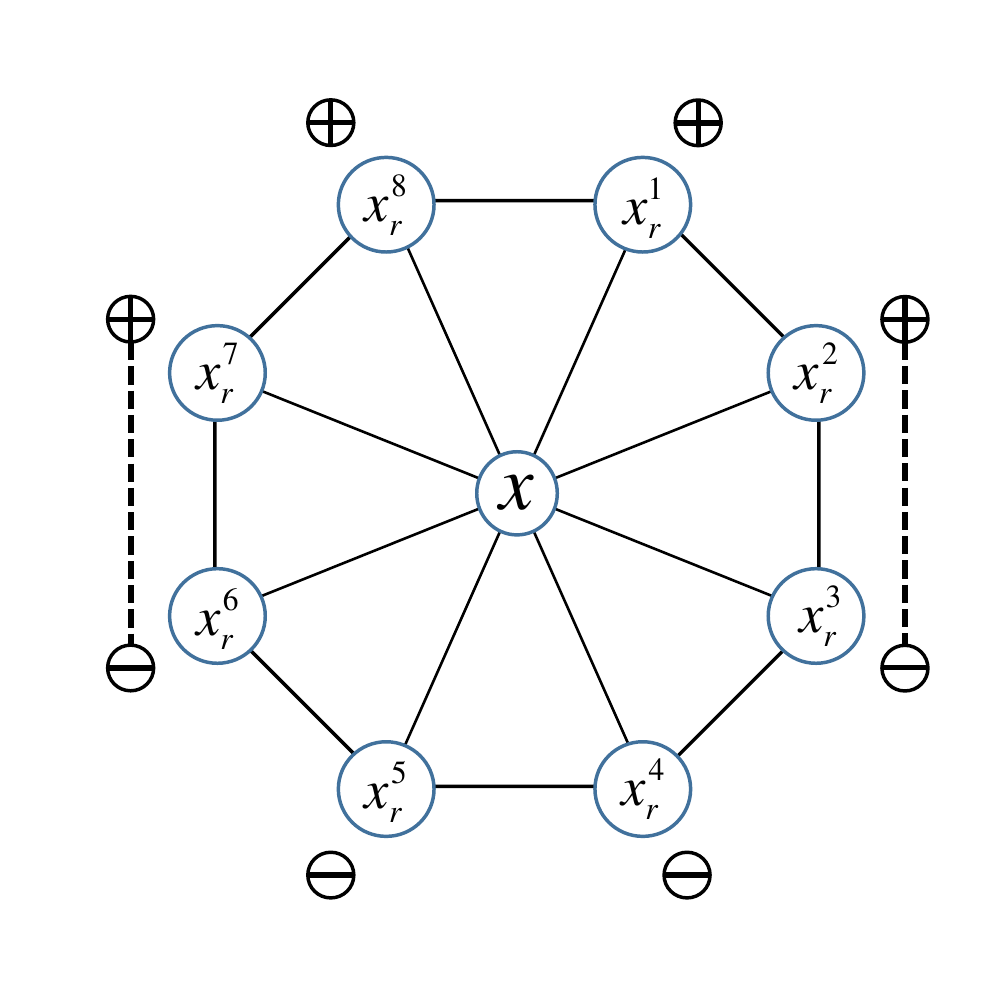}
   }
   \subfigure[]
   {
        \includegraphics[width=0.38\textwidth]
            {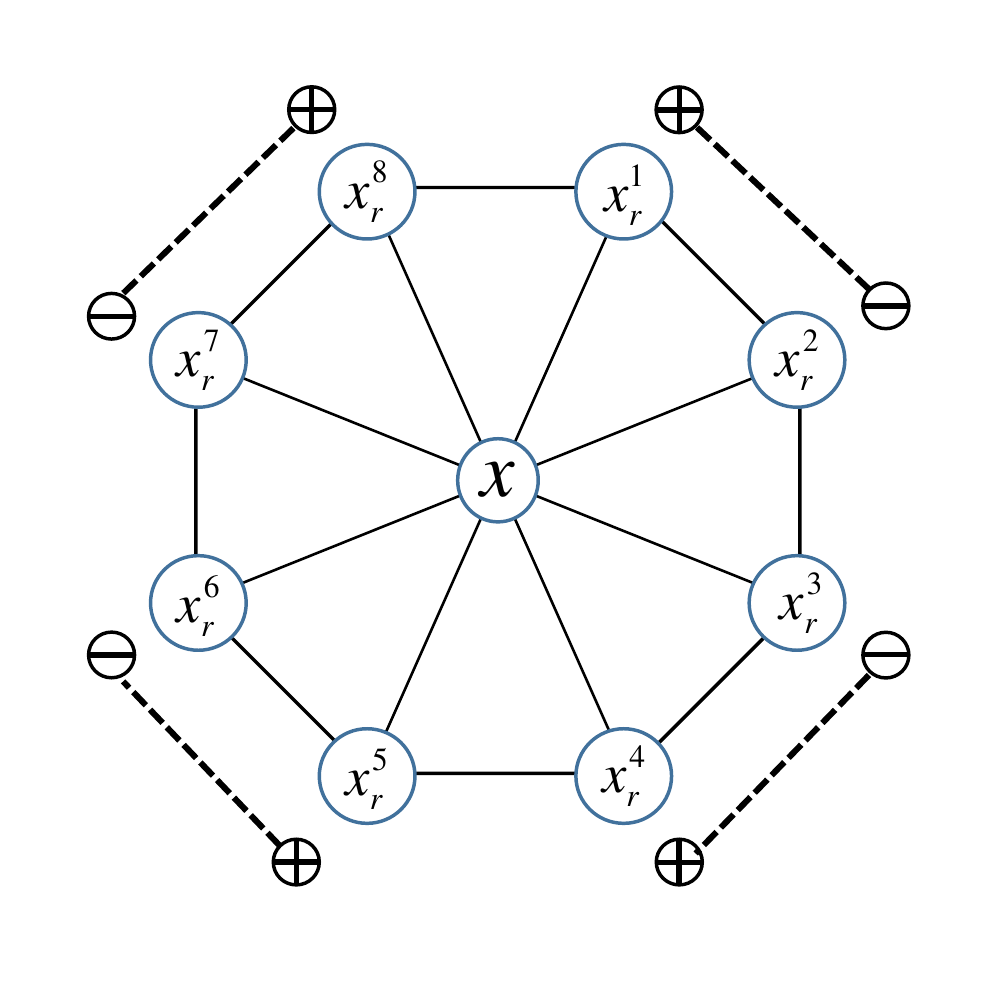}
   }
   \subfigure[]
   {
        \includegraphics[width=0.38\textwidth]
            {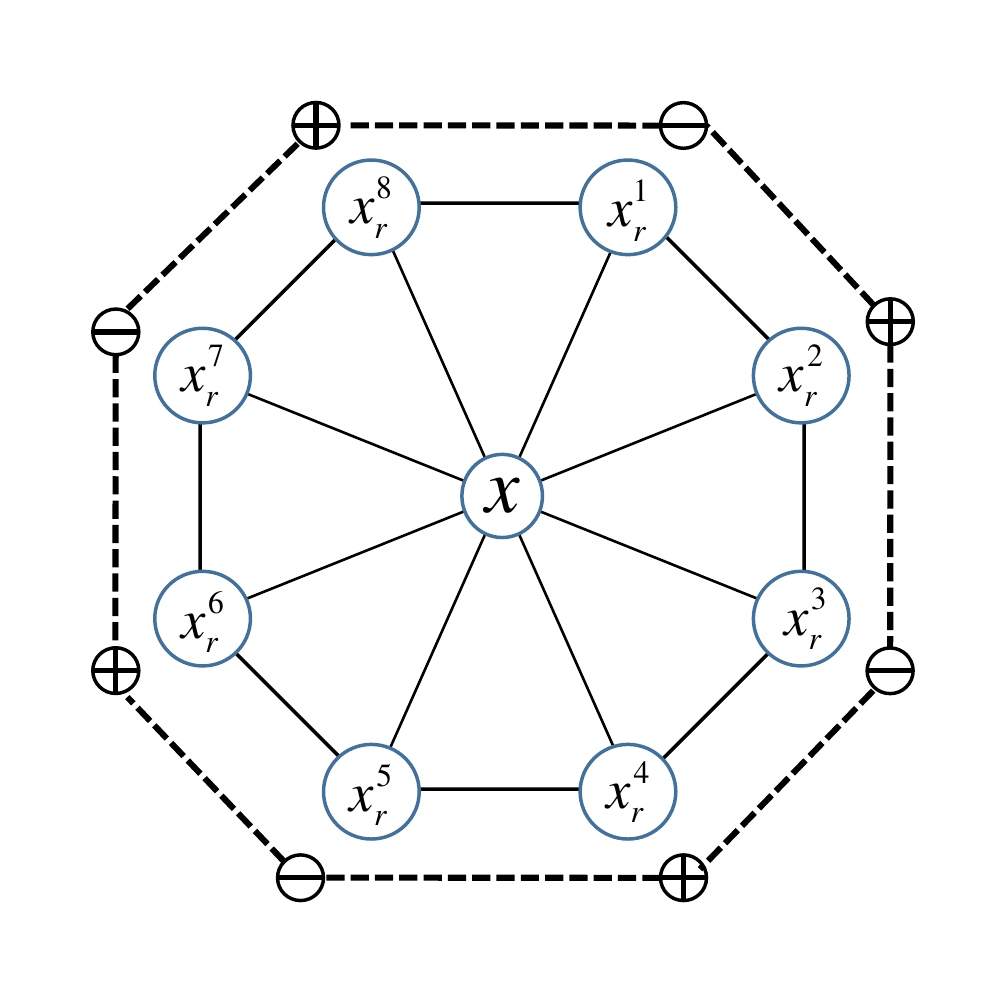}
   }
   \caption
   {
        Saddle point detection.
        Dotted line indicates the change of signs between neighbor points.
        While the points $x$ in (a) and (b) are not saddle point,
        the points  $x$ in (c) and (d) are saddle point.
   }
   \label{fig:saddle}
   \end{minipage}
 \end{figure}

 \textbf{Saddle point detection:}
 To detect the saddle points of $\omega(x)$ defined on
    an unorganized high-dimensional data point set,
    the points of the set are projected into a 2D plane,
    using the DR methods~\cite{Belkin2003Laplacian}.
 The data point projected into a 2D plane continues to be denoted as $x$.
 For a point $x$,
    the points in its \emph{KNN} neighbor $N_x$ comprise its one-ring neighbor $R_x$ (see Fig.~\ref{fig:saddle}).
 Each point $x_r$ in $R_x$ has a sole preceding point $x_r^p$ and a
    sole succeeding point $x_r^n$ in a counter-clockwise direction.
 As illustrated in Fig.~\ref{fig:saddle},
    if $\omega(x) \geq \omega(x_r)$,
    the point $x_r$ is labeled as $\ominus$;
    otherwise, it is labeled as $\oplus$.
 Now, we initialize a counter $sum = 0$
    and traverse the one-ring neighbor $R_x$ of point $x$ from an arbitrary point.
 If the signs of two neighbor points in $R_x$ are different
    (marked with dotted lines),
    $sum = sum + 1$.
 Finally, if the counter $sum \geq 4$,
    point $x$ is a saddle point;
    otherwise, it is not a saddle point (Fig.~\ref{fig:saddle}).
    Examples of saddle points and non-saddle points are
    demonstrated in Fig.~\ref{fig:saddle}.

\subsection{Measurement of Fidelity of the Simplified Data Set}
\label{subsec:metrics}

 As stated in Section~\ref{subsec:data_simplification},
    the data simplification method detects the feature points of the eigenvectors $\varphi_i, i=1,2,\cdots,n$,
    and adds them to the simplified data set iteratively,
    until the simplified data set satisfies a preset threshold of fidelity to the original data set.
 In this section, three metrics are devised to measure the fidelity of the
    simplified data set $\mathcal{S}$ to the original high-dimensional data set $\mathcal{H}$;
    these are the Kullback-Leibler divergence (KL divergence)~\cite{Perez2008Kullback}-based metric,
    Hausdorff distance~\cite{dubuisson1994modified},
    and determinant of covariance matrix~\cite{Sharma2010Shape}.
 They measure the fidelity of the simplified data set $\mathcal{S}$ to the
    original data set $\mathcal{H}$ from three aspects:
  \begin{itemize}
    \item entropy of information (KL-divergence-based metric),
    \item distance between two sets (Hausdorff distance),and
    \item 'volume' of a data set (determinant of covariance matrix).
  \end{itemize}

 \textbf{KL-divergence-based metric.}
    KL-divergence~\cite{Perez2008Kullback},
    also called relative entropy,
    is typically employed to measure the difference between two probability distributions $P = \{p_{i}\}$ and $Q = \{q_{i}\}$, i.e.,
    \begin{equation}\label{eq:KL devergence}
       D_{KL}(P\|Q) = \sum_{i} p_{i} log\left(\frac{p_{i}}{q_{i}}\right).
     \end{equation}

 To define the KL-divergence-based metric,
    the original data set $\mathcal{H}$ and simplified data set $\mathcal{S}$ are represented as an $n \times d$ matrix $M_{\mathcal{H}}$,
    and an $m \times d$ matrix $M_{\mathcal{S}}$,
    where $n, m, (n > m)$ are the number of data points in the data sets $\mathcal{H}$ and $\mathcal{S}$, respectively.
 Each row of the matrices $\mathcal{H}$ and $\mathcal{S}$ stores a data point
    in the $d-$dimensional Euclidean space.
 First, we calculate the probability distribution
    $P^{(k)} = \{p^{(k)}_i\}, k = 1,2,\cdots,d$
    for each dimension of the data points in the original data set $\mathcal{H}$,
    corresponding to the $k^{th}$ column of matrix $M_{\mathcal{H}}$.
 To do this,
    the $k^{th}$ column elements of matrix $M_{\mathcal{H}}$ are normalized into the interval $[0,1]$,
    which is divided into $l$ subintervals,
    (in our implementation, we use $l=100$).
 By counting the number of elements of the $k^{th}$ column of matrix
    $M_{\mathcal{H}}$
    whose values lie in the $l$ subintervals,
    the value distribution histogram is generated,
    which is then used as the probability distribution
    $P^{(k)} = \{p^{(k)}_i, i=1,2,\cdots,l\}$
    of the $k^{th}, k=1,2,\cdots,d$ column of matrix $M_{\mathcal{H}}$.
 Similarly, the probability distribution
    $Q^{(k)} = \{q^{(k)}_i, i=1,2,\cdots,l\}$
    of the $k^{th}, k=1,2,\cdots,d$ column of matrix $M_{\mathcal{S}}$ can be calculated.
 Then, the KL-divergence for the $k^{th}, k=1,2,\cdots,d$ column elements of
    matrix $M_{\mathcal{H}}$ and $M_{\mathcal{S}}$  is
    \begin{equation*}
       D_{KL}(P^{(k)}\|Q^{(k)}) = \sum_{i} p^{(k)}_{i} log\left(\frac{p^{(k)}_{i}}{q^{(k)}_{i}}\right).
     \end{equation*}
 And the KL-divergence-based metric is defined as
 \begin{equation} \label{eq:kl_metric}
       d_{KL}(\mathcal{H}, \mathcal{S}) =
        \sqrt{\sum_{k=1}^d (D_{KL}(P^{(k)}\|Q^{(k)}))^2}.
 \end{equation}

 The KL-divergence-based metric~\pref{eq:kl_metric} transforms the
    high-dimensional data set into its probability distribution,
    and measures the difference between the probability distributions of the original and simplified data sets.
 As the simplified data set $\mathcal{S}$ becomes increasingly closer to the original data set $\mathcal{H}$,
    the KL-divergence-based metric $d_{KL}$ becomes increasingly smaller.
 When $\mathcal{S} = \mathcal{H}$, $d_{KL} = 0$.

 \textbf{Hausdorff distance.}
 The Hausdorff distance~\cite{dubuisson1994modified} measures how far two
    sets are away from each other.
 Suppose $\mathcal{X}$ and $\mathcal{Y}$ are two data set in
    $d-$dimensional space,
    and denote
    \begin{equation*}
        D(\mathcal{X},\mathcal{Y}) =
        \sup_{x\in \mathcal{X}}\inf_{y\in \mathcal{Y}} d(x,y),
    \end{equation*}
    the maximum distance from an arbitrary point in set $\mathcal{X}$
        to set $\mathcal{Y}$,
    where $d(x,y)$ is the Euclidean distance between two points $x$ and $y$.
 The Hausdorff distance between the original data set $\mathcal{H}$ and
    simplified data set $\mathcal{S}$ is defined as
    \begin{equation*}
       \begin{aligned}
          d_{H}(\mathcal{H},\mathcal{S})
            & =
              max \{D(\mathcal{H},\mathcal{S}),D(\mathcal{S},\mathcal{H})\}\\
            &= max \{\sup_{x\in \mathcal{H}}\inf_{y \in \mathcal{S}} d(x,y),
                     \sup_{y\in \mathcal{S}}\inf_{x\in \mathcal{H}} d(x,y)\}.
       \end{aligned}
    \end{equation*}
 Note that the simplified data set $\mathcal{S}$ is a subset of the original
    data set $\mathcal{H}$, and then, $D(\mathcal{S},\mathcal{H}) = 0$.
 Therefore,
   the Hausdorff distance between $\mathcal{H}$ and $\mathcal{S}$
    can be calculated by
    \begin{equation}\label{eq:Hausdorff_distance}
          d_{H}(\mathcal{H},\mathcal{S}) =
            D(\mathcal{H},\mathcal{S})\\
            = \sup_{x\in \mathcal{H}}\inf_{y \in \mathcal{S}} d(x,y).
    \end{equation}
 That is,
    the Hausdorff distance is the maximum distance from an arbitrary point
    in the original data set $\mathcal{H}$ to the simplified data set $\mathcal{S}$.

 \textbf{Determinant of covariance matrix.}
 The covariance matrix is frequently employed in statistics and probability theory
   to measure the joint variability of several variables~\cite{Sharma2010Shape}.
 Suppose
    \begin{equation*}
     X_i = (x_{i1}, x_{i2}, \cdots, x_{id}), i = 1,2,\cdots,m,
    \end{equation*}
    are $m$ samples in a $d-$dimensional sample space $X$,
    and denote
    \begin{equation*}
        \bar{X}_j = \frac{\sum_{i=1}^m x_{ij}}{m},\ j=1,2,\cdots,d,
    \end{equation*}
    where $\bar{X}_j$ is the mean value of the $j$-th random variable.
 The covariance $c_{ij}$ between the $i$-th and $j$-th random variables is
    \begin{equation*}
        c_{ij} = \frac{1}{m} \sum_{k=1}^m (X_{ki}-\bar{X}_i)(X_{kj}-\bar{X}_j),
    \end{equation*}
    which is the $(i,j)^{th}-$element of the covariance matrix $Cov(X)$ of the data set $X$.
 According to ~\cite{Sharma2010Shape},
    the determinant of the covariance matrix of $X$,
     i.e., $det(Cov(X))$,
     can express the 'volume' of the data set $X$.
 When the simplified data set $\mathcal{S}$ approaches the original data
    set $\mathcal{H}$,
    they are expected to have increasingly 'volumes',
    namely, the determinants of covariance matrices.
 Therefore, the difference of the determinants of the covariance matrices of
    $\mathcal{H}$ and $\mathcal{S}$:
    \begin{equation} \label{eq:covariance}
        d_{COV}(\mathcal{H}, \mathcal{S}) =
        \abs{det(Cov(\mathcal{H})) - det(Cov(\mathcal{S}))},
    \end{equation}
    can measure the fidelity of the simplified data set $\mathcal{S}$ to the original data set $\mathcal{H}$.

%
\section{Results, Discussion, and Applications}
\label{sec:discussion_results}

\begin{figure}[!htb]
   \centering
    \subfigure[]
    {
        \label{subfig:swiss}
        \includegraphics[width=0.22\textwidth]
        {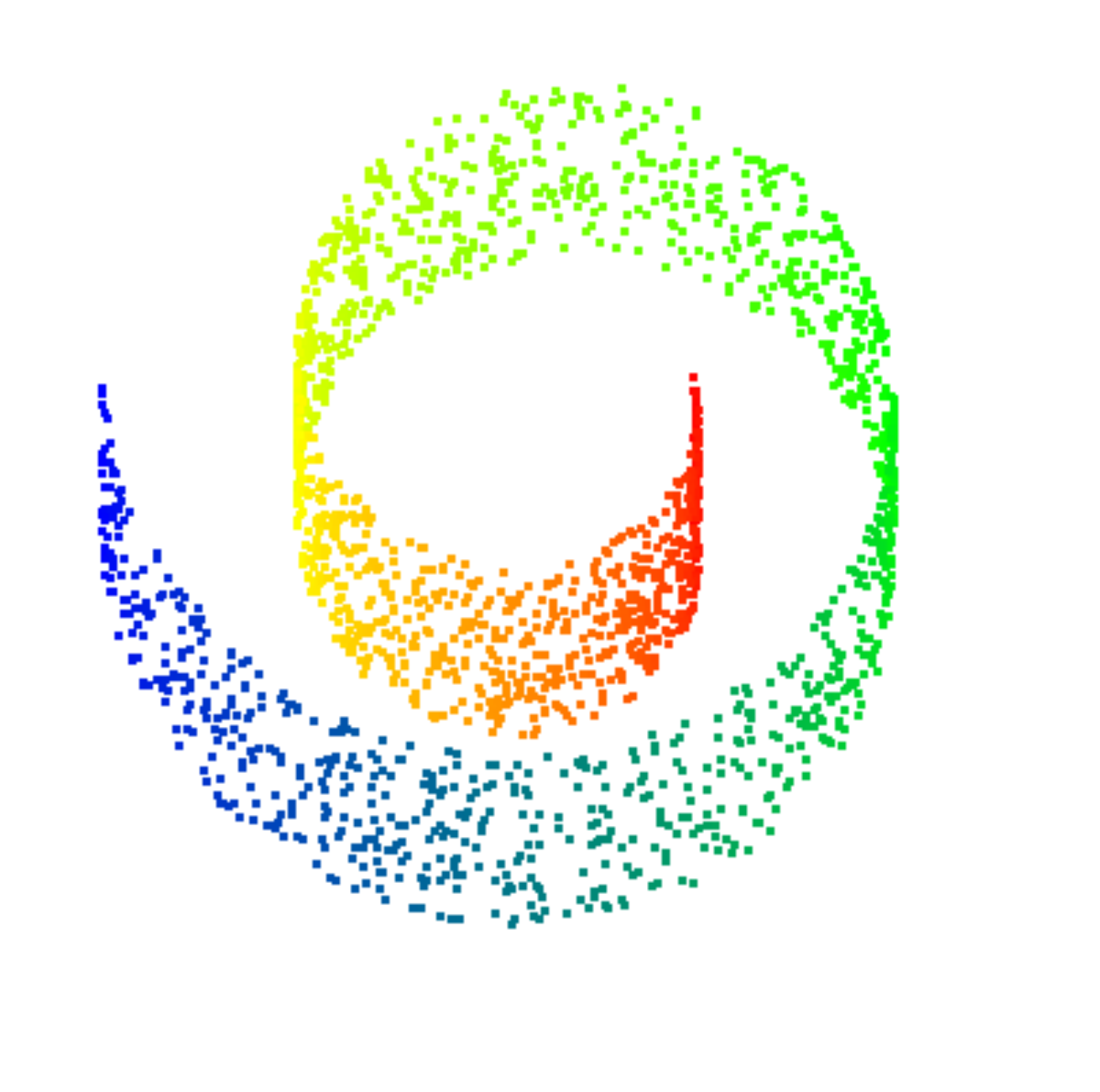}
     }
    \subfigure[]
    {
        \label{subfig:mnist}
        \includegraphics[width=0.22\textwidth]
        {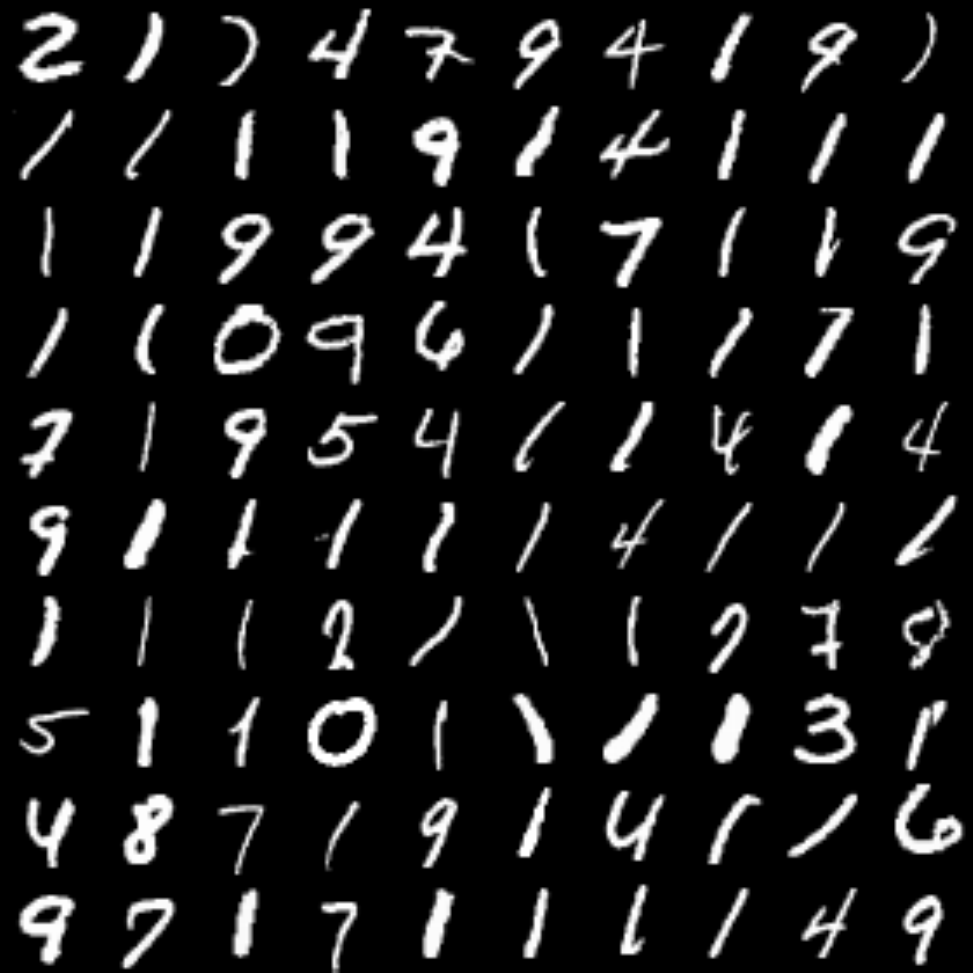}
    }
    \subfigure[]
    {
        \label{subfig:face}
        \includegraphics[width=0.22\textwidth]
        {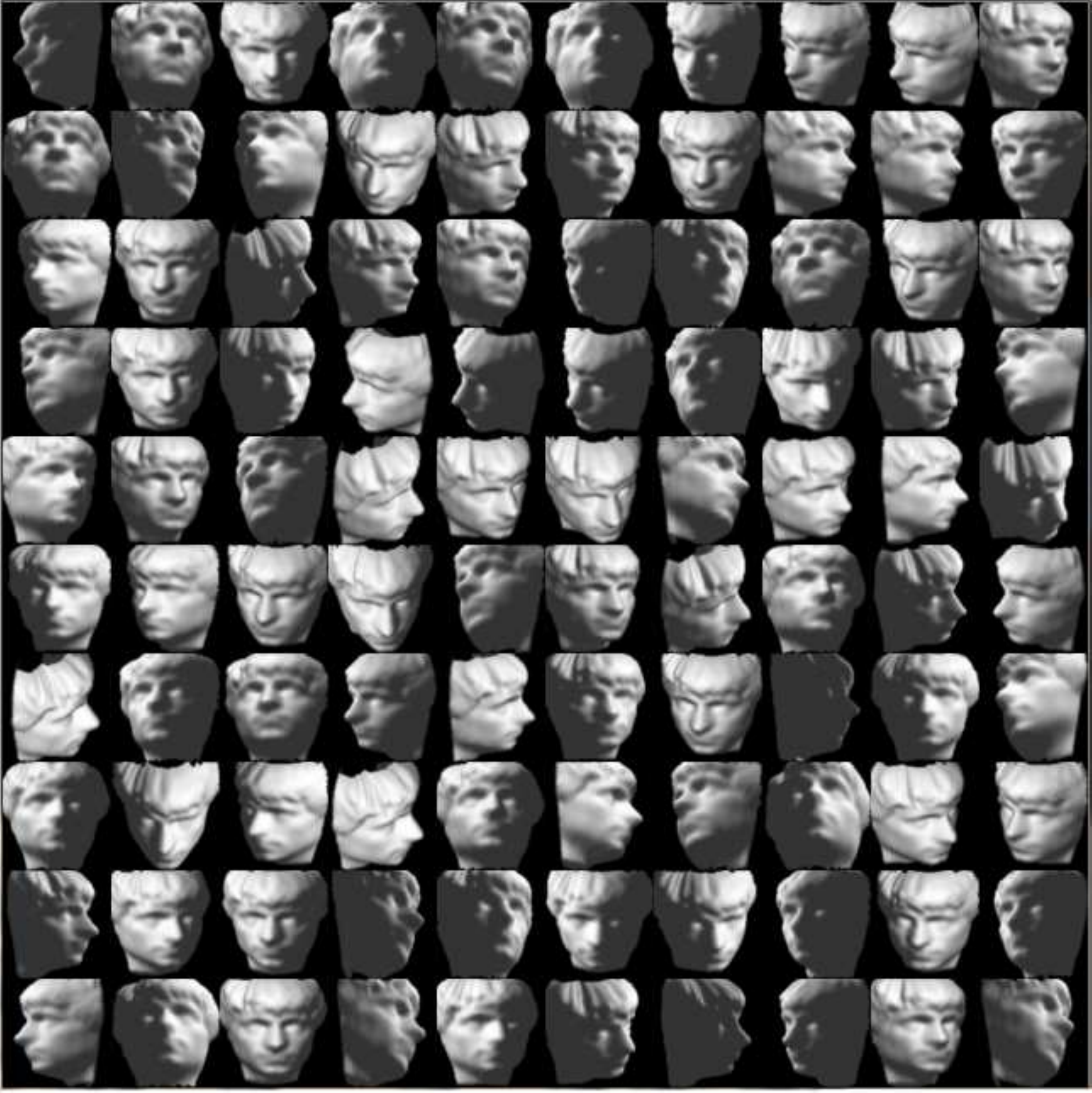}
    }
    \subfigure[]
    {
        \label{subfig:cifar10}
        \includegraphics[width=0.22\textwidth]
        {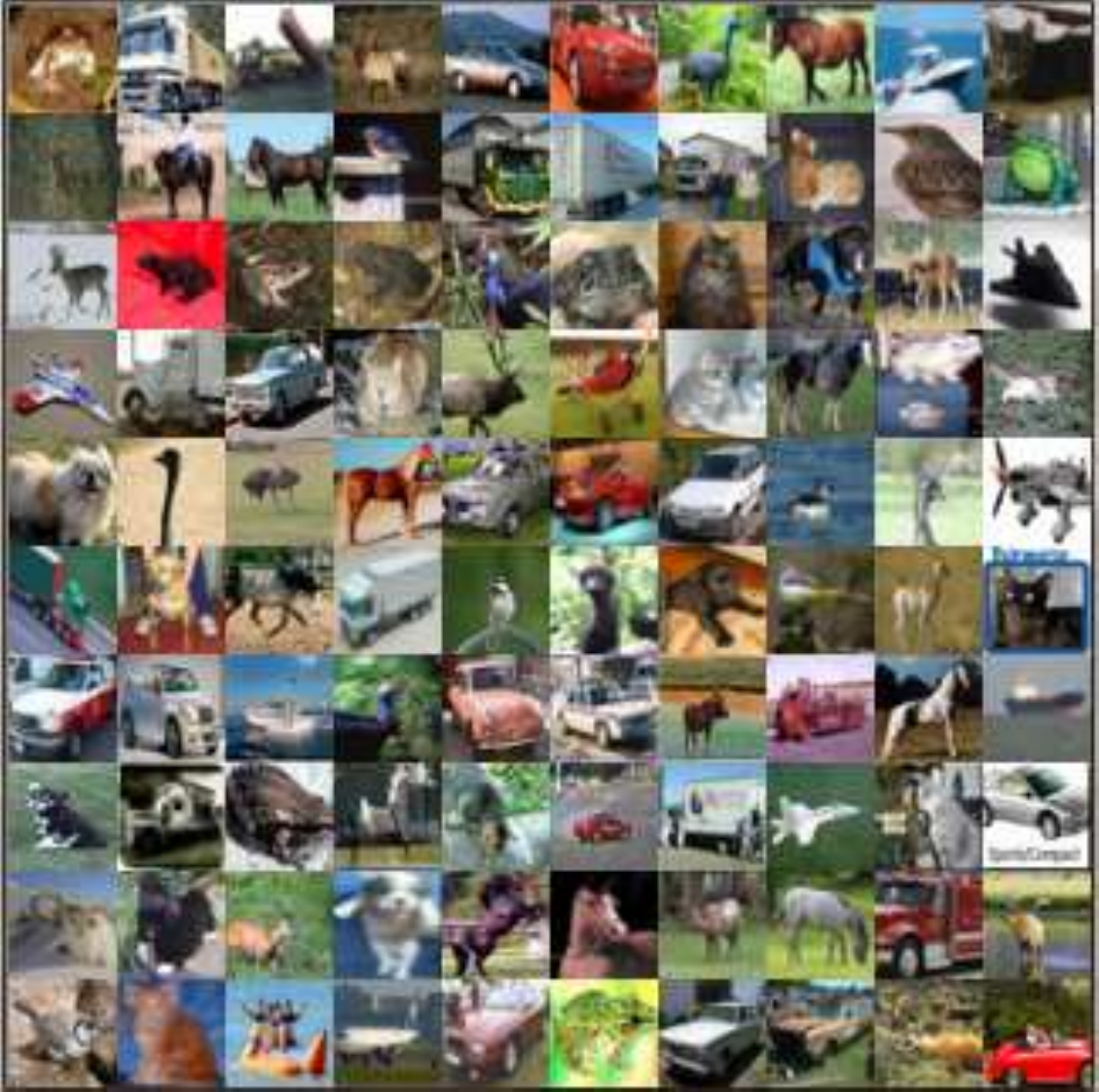}
    }
      \caption
      {
        Snapshots of the data sets employed in this paper.
        (a) \emph{Swiss roll}.
        (b) \emph{MNIST}.
        (c) \emph{Human face}.
        (d) \emph{CIFAR-10}.
      }
      \label{fig:data}
   \end{figure}

 In this section, the proposed method is employed to simplify four
    well-known high-dimensional data sets (refer to Fig.~\ref{fig:data}),
    including
    the \emph{swiss roll} data set~\cite{Tenenbaum2000A} (Fig.~\ref{subfig:swiss})
        with $2000$ points in 3D space,
    \emph{MNIST} handwritten digital data set~\cite{lecun1998mnist} (Fig.~\ref{subfig:mnist})
        which contains 60000 images ($28 \times 28$) of handwritten digits from '0' to '9',
    \emph{human face} data set~\cite{Tenenbaum2000A} (Fig.~\ref{subfig:face})
        which includes 698 images($64 \times 64$) of a plaster head sculpture,
    and \emph{CIFAR-10} classification data set~\cite{krizhevsky2014cifar} (Fig.~\ref{subfig:cifar10})
        with $50000$ RGB-images($32 \times 32 \times 3$).
    The fidelity of the simplified data sets to the original data sets are
      measured with the three metrics described in Section~\ref{subsec:metrics}.
    Moreover,
      two applications of a simplified data set are also demonstrated
      including the speedup of DR
      and a training data simplification in supervised learning.

 The proposed method is implemented using C++ with OpenCV and
   ARPACK~\cite{lehoucq2007arpack},
   and executes on a PC with a $3.6$ GHz Intel Core $i7-4790$ CPU,
   GTX $1060$ GPU,
   and $16G$ memory.
 We employed the OpenCV function 'cv::flann' to perform
    the \emph{KNN} operation~\cite{muja2009flann}
    and ARPACK to solve the sparse symmetric matrix eigen-equation~\pref{eq:discrete_helmholtz}.

%
%

 \subsection{Simplification Results}
 \label{subsec:results_and_applications}

 In this section, we will illustrate several data simplification results.

 \begin{figure}[!htb]
   \centering
   \subfigure[]
   {
     \label{subfig:swiss_5}
     \includegraphics[width=0.14\textwidth]{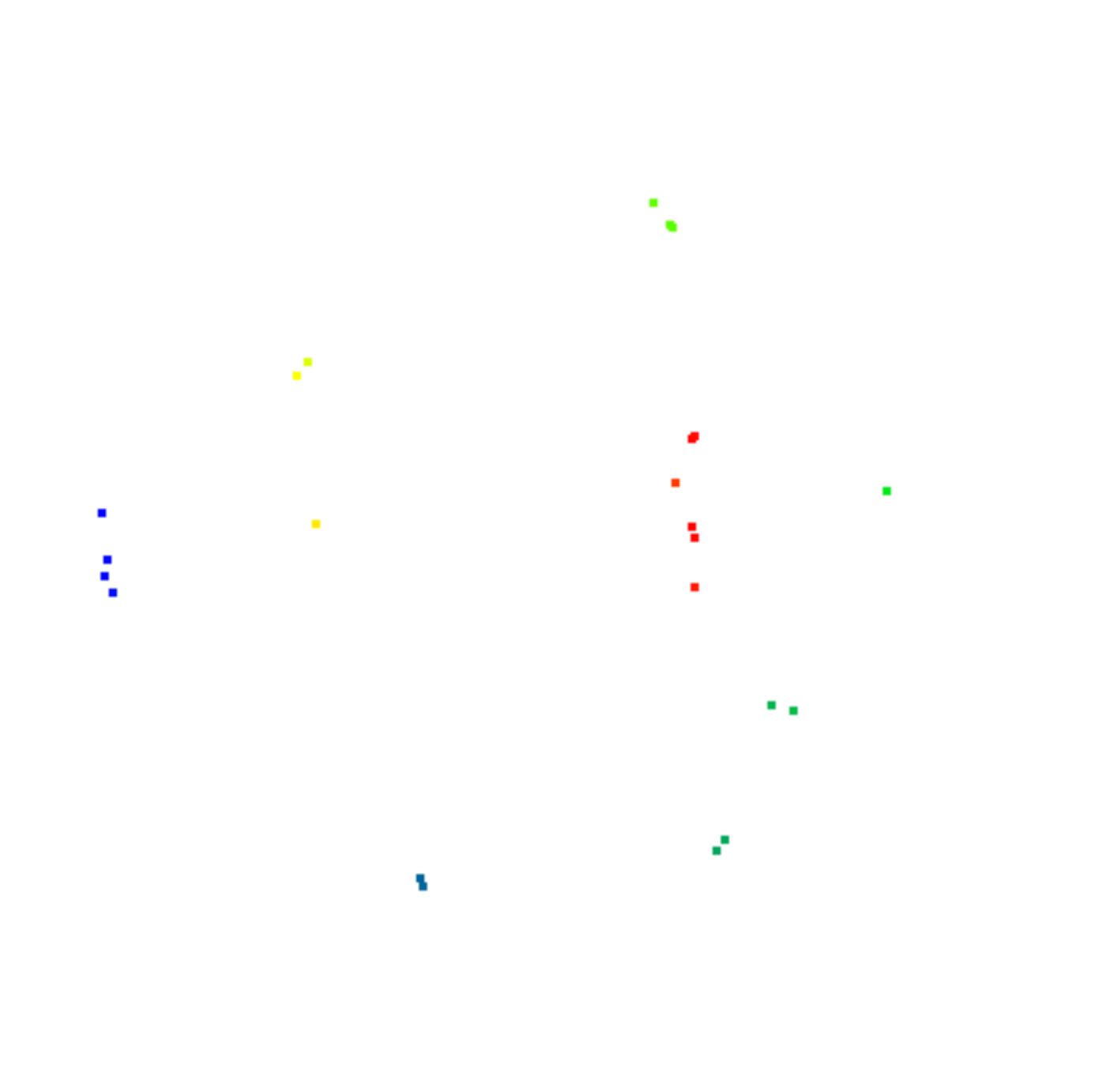}
   }
   \subfigure[]
   {
      \label{subfig:swiss_10}
      \includegraphics[width=0.14\textwidth]{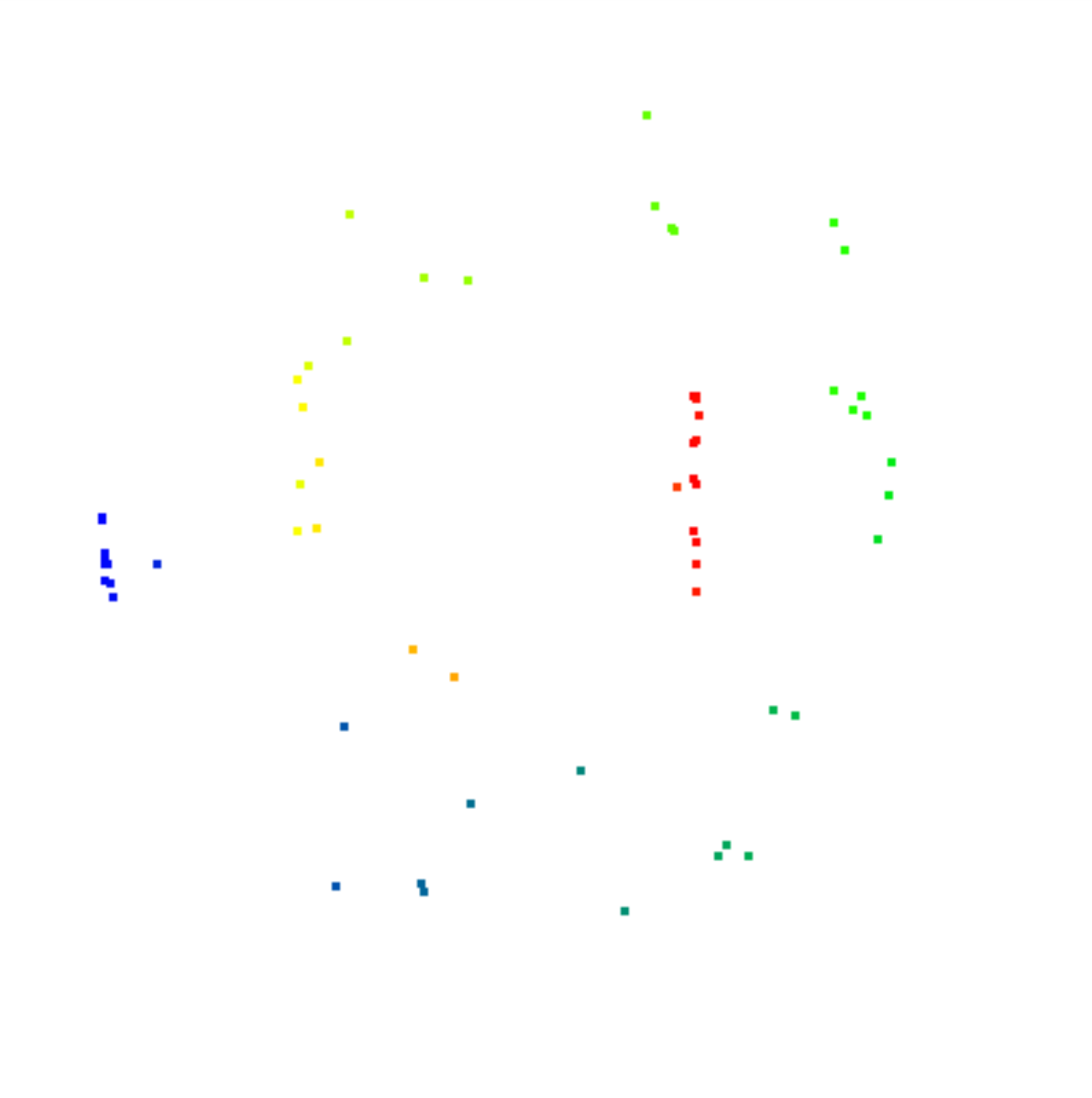}
   }
   \subfigure[]
   {
      \label{subfig:swiss_30}
      \includegraphics[width=0.14\textwidth]{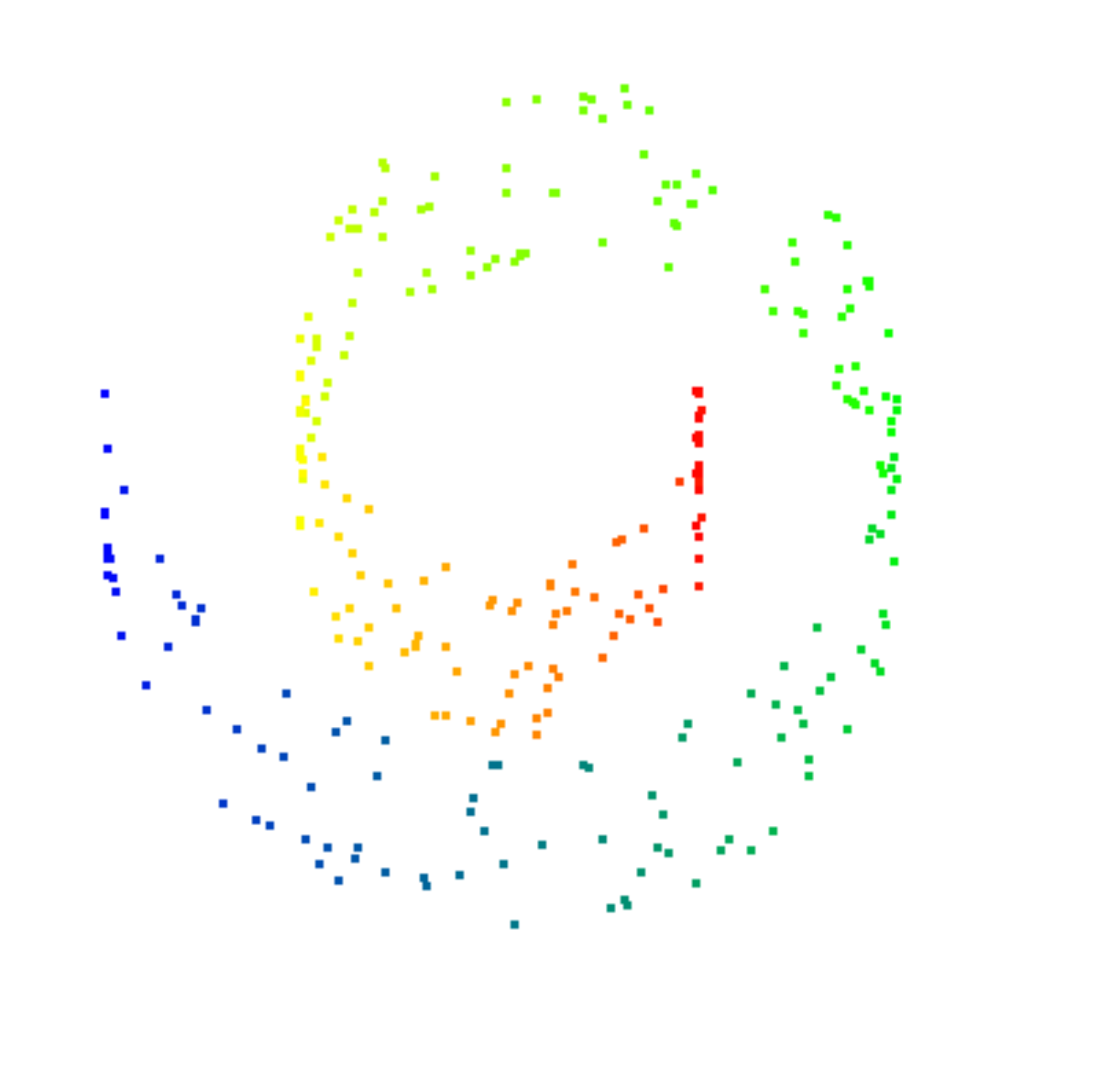}
   }
   \subfigure[]
   {
     \label{subfig:swiss_50}
     \includegraphics[width=0.14\textwidth]{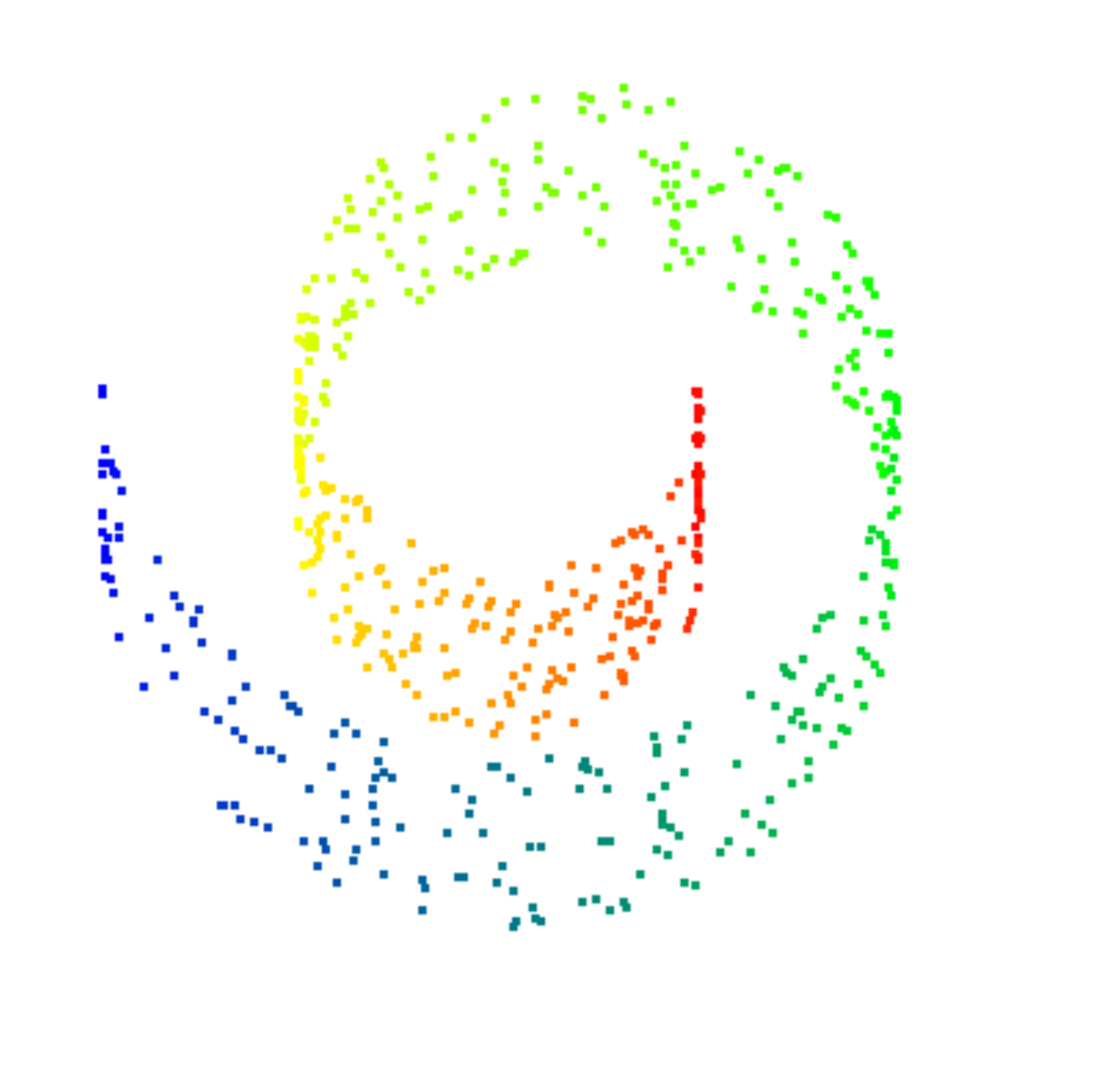}
   }
   \subfigure[]
   {
      \label{subfig:swiss_100}
      \includegraphics[width=0.14\textwidth]{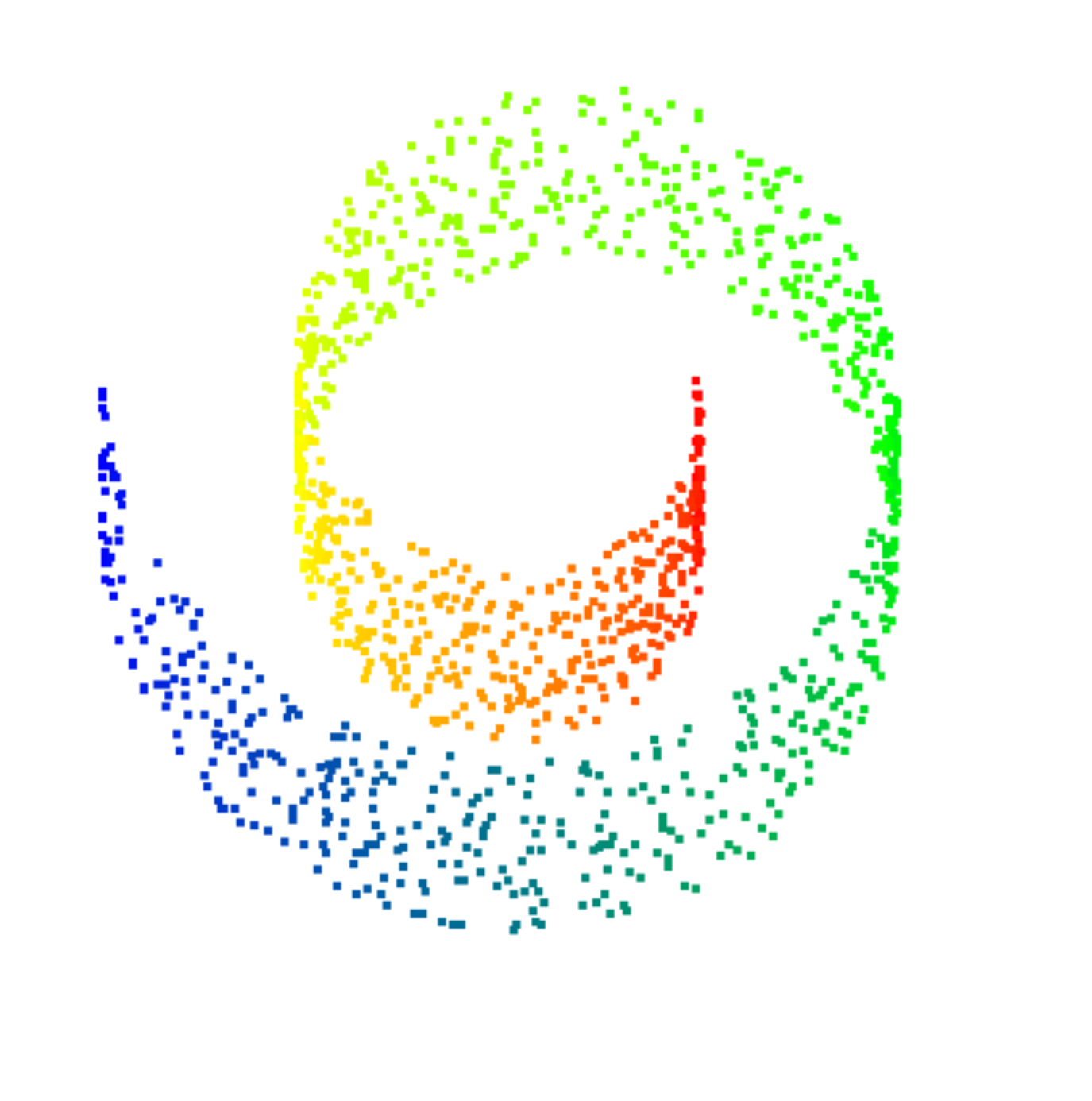}
   }
   \subfigure[]
   {
      \label{subfig:swiss_original}
      \includegraphics[width=0.14\textwidth]{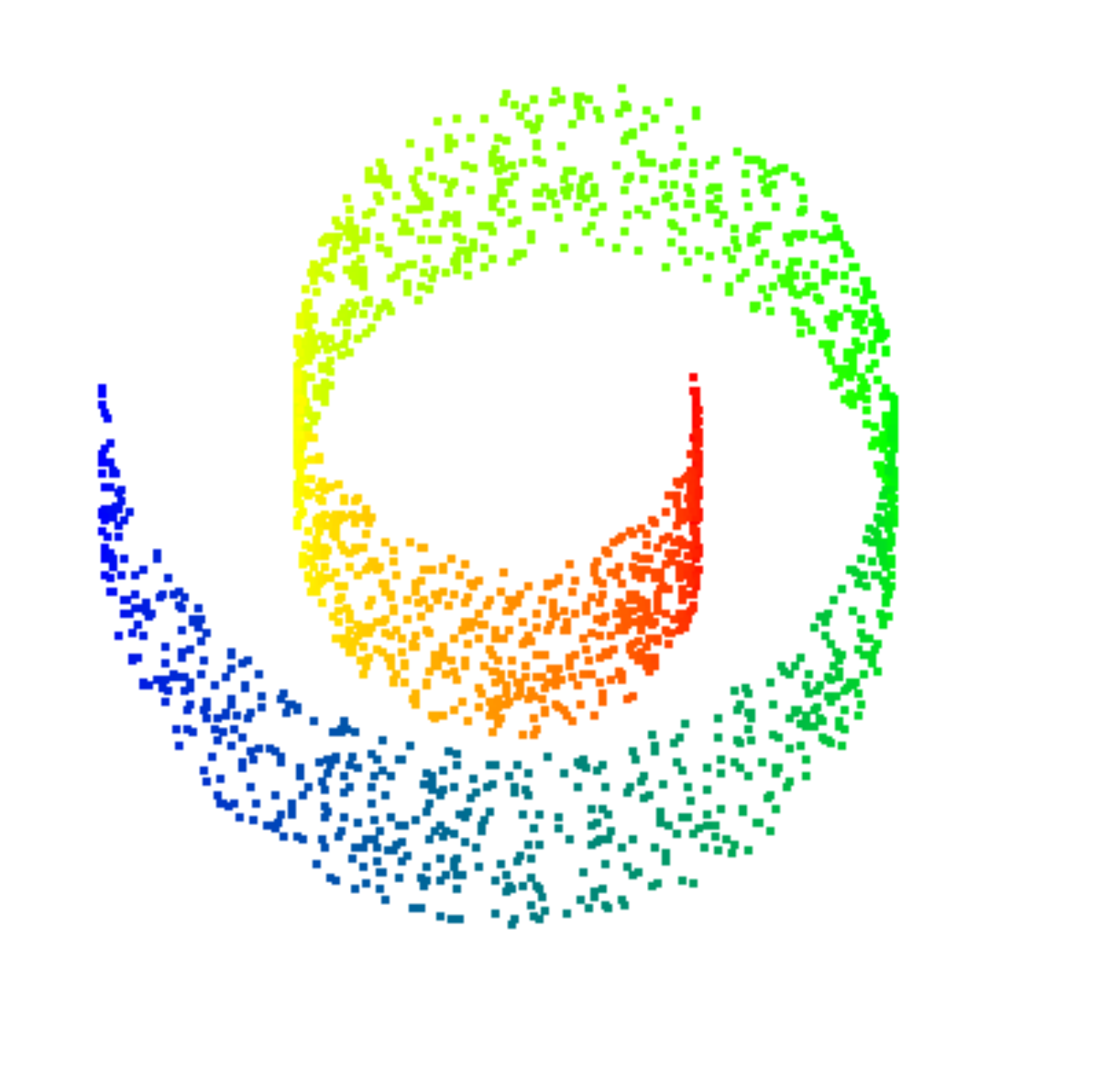}
   }
   \caption
   {
        The intuitive data simplification procedure of the \emph{swiss roll}
            data set.
        In (a)-(e),
            the simplified data sets consist of $24, 62, 397, 612, 1286$ data points,
            generated by detecting the feature points of the first $5, 10, 30, 50, 100$ LBO eigenvectors, respectively.
       (f) is the original data with 2000 points.
   }
   \label{fig:swissdetection}
   \end{figure}

 \textbf{Swiss roll}:
 In Fig.~\ref{fig:swissdetection},
    we use the \emph{swiss roll} data set in 3D space
    to illustrate the intuitive simplification procedure by the proposed data simplification algorithm.
 In Fig~\ref{subfig:swiss_5},
    the simplified data set is generated by detecting the feature points of the first five eigenvectors,
    which captures the critical points at the head and tail positions of the roll (marked in red and blue colors).
 In Figs.~\ref{subfig:swiss_10}-~\ref{subfig:swiss_100},
    by detecting higher frequency eigenvectors,
    increasingly more feature points are added into the simplified data set and
    it approaches increasingly closer to the original data set
    (Fig.~\ref{subfig:swiss_original}).
 Note that
    the simplified data set in Fig.~\ref{subfig:swiss_30},
    which contains $397$ data points (approximately $20\%$ points of the entire data set),
    includes the majority of the information of the original data set.
 Please refer to Section~\ref{subsec:quantitative_analysis}
    for the quantitative fidelity analysis of the simplified data set to the original data set
    measured by the metrics
    $d_{KL}$~\pref{eq:kl_metric},
    $d_{H}$~\pref{eq:Hausdorff_distance},
    and $d_{COV}$~\pref{eq:covariance}.

\begin{figure}[!htb]
   \centering
    \subfigure[]
    {
        \label{subfig:hand_5}
        \includegraphics[width=0.18\textwidth]{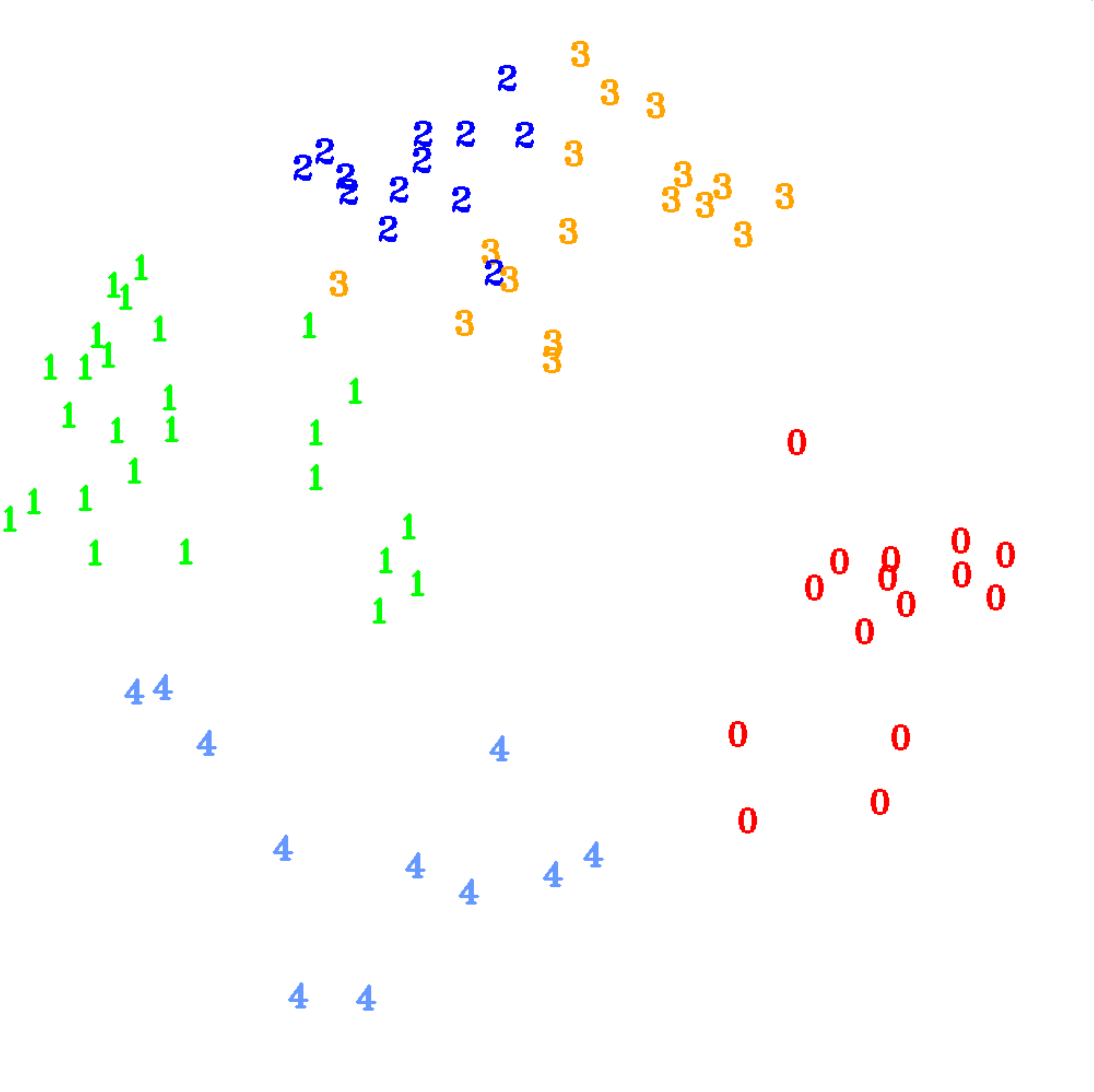}
    }
    \subfigure[]
    {
        \label{subfig:hand_10}
        \includegraphics[width=0.18\textwidth]{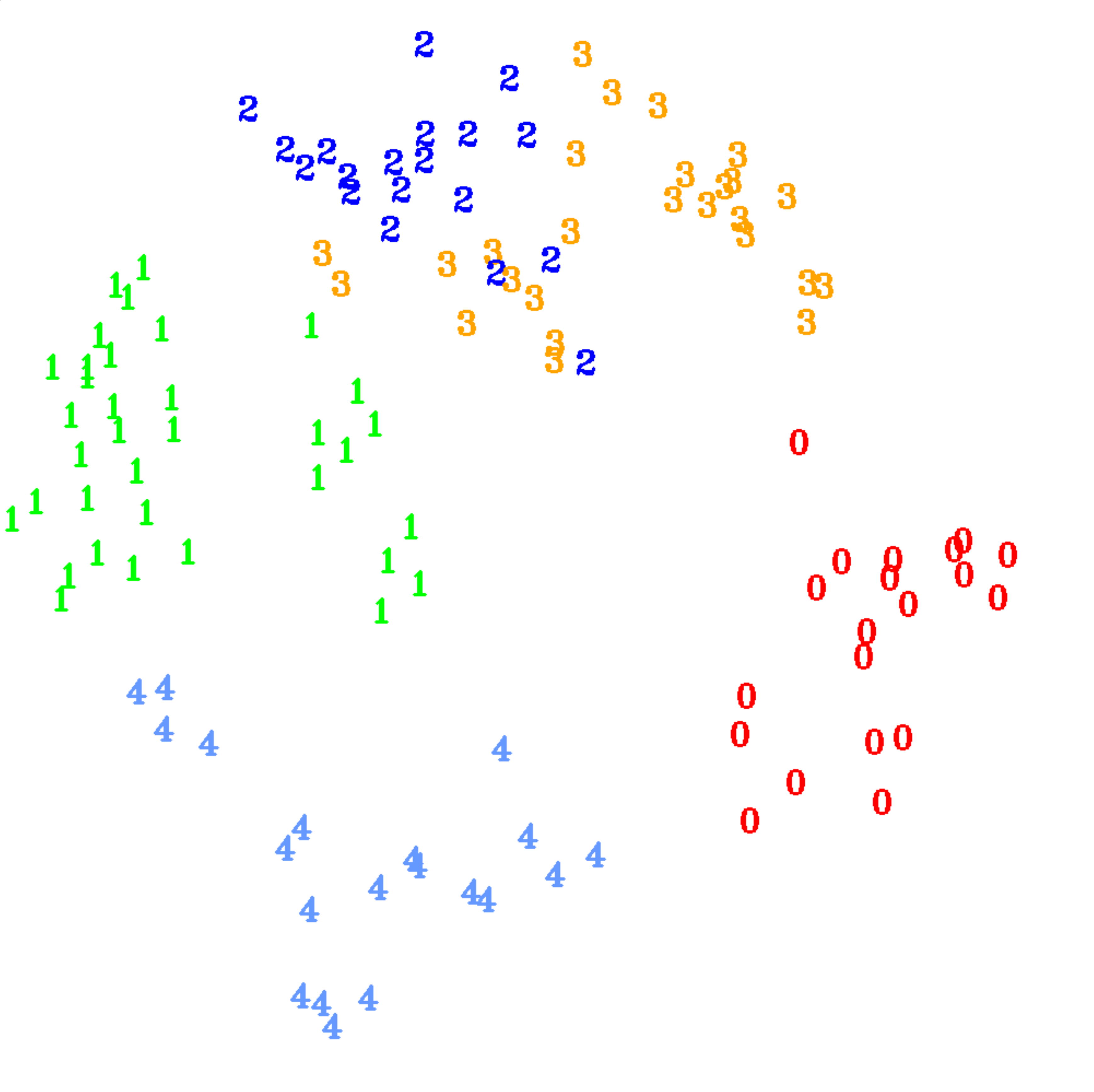}
    }
    \subfigure[]
    {
        \label{subfig:hand_20}
        \includegraphics[width=0.18\textwidth]{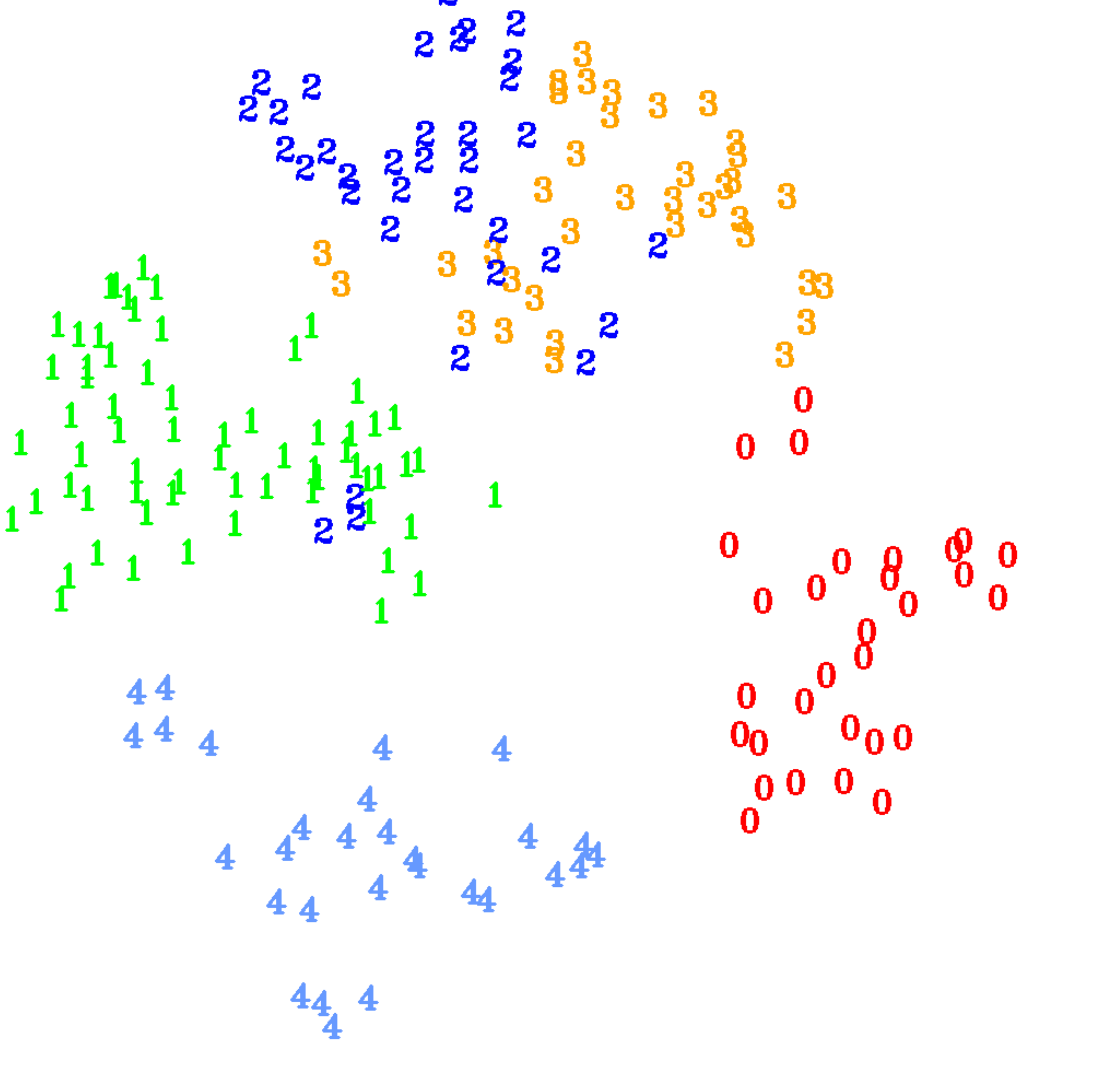}
    }
    \subfigure[]
    {
        \label{subfig:hand_30}
        \includegraphics[width=0.18\textwidth]{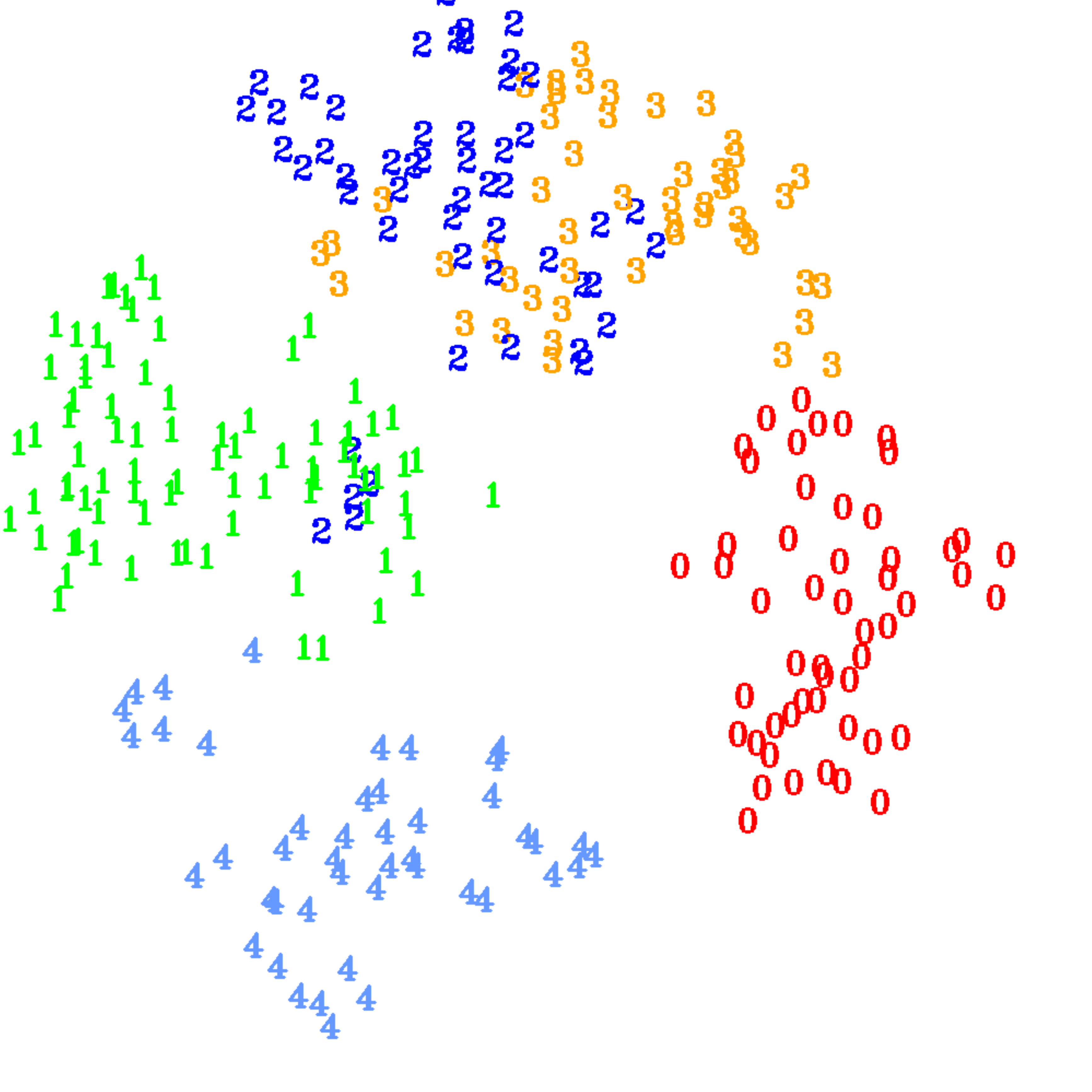}
    }
    \subfigure[]
    {
        \label{subfig:hand_color_replaced}
        \includegraphics[width=0.18\textwidth]{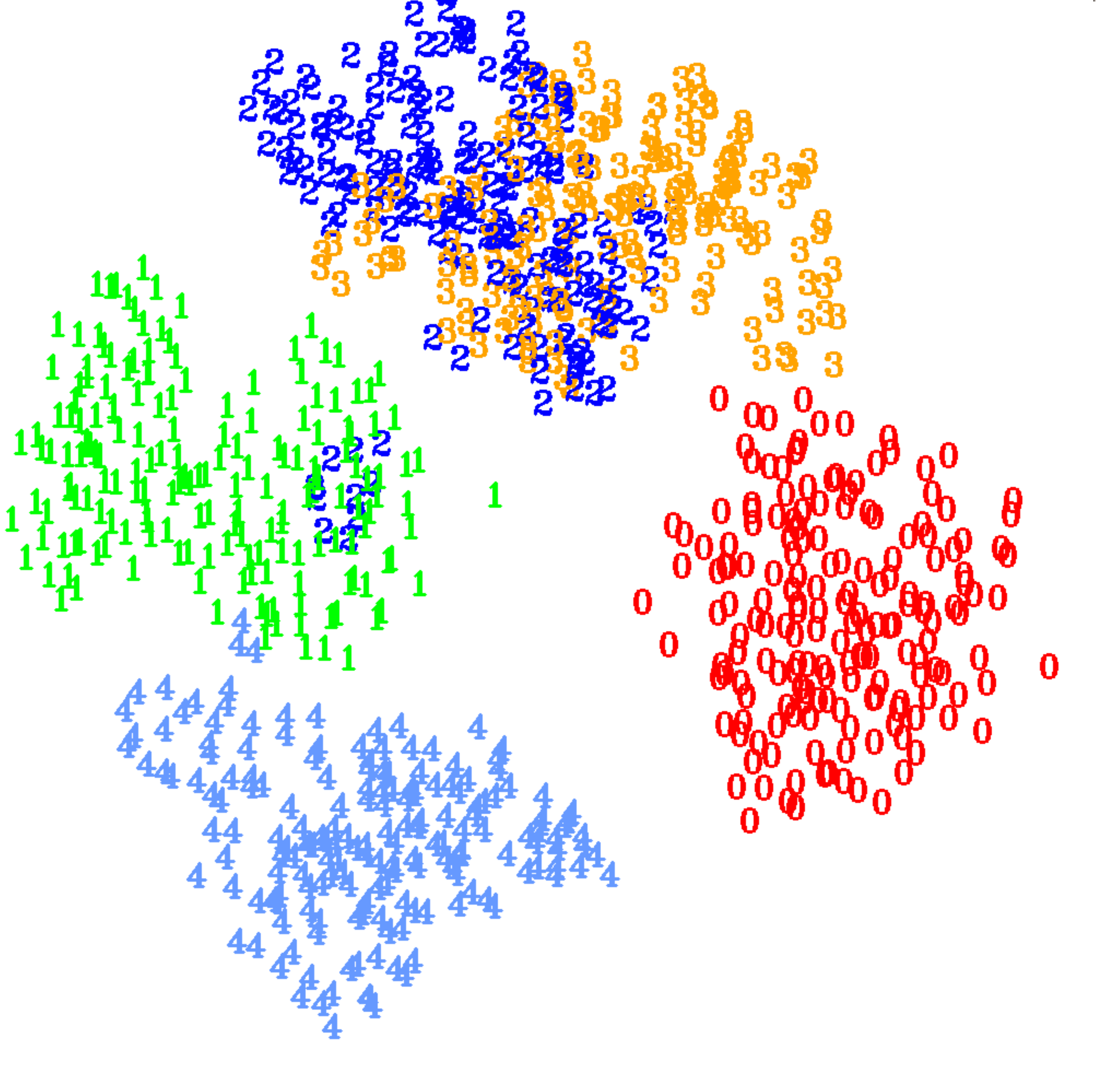}
    }
    \subfigure[]
    {
      \label{subfig:hand_image_replaced}
      \includegraphics[width=0.18\textwidth]{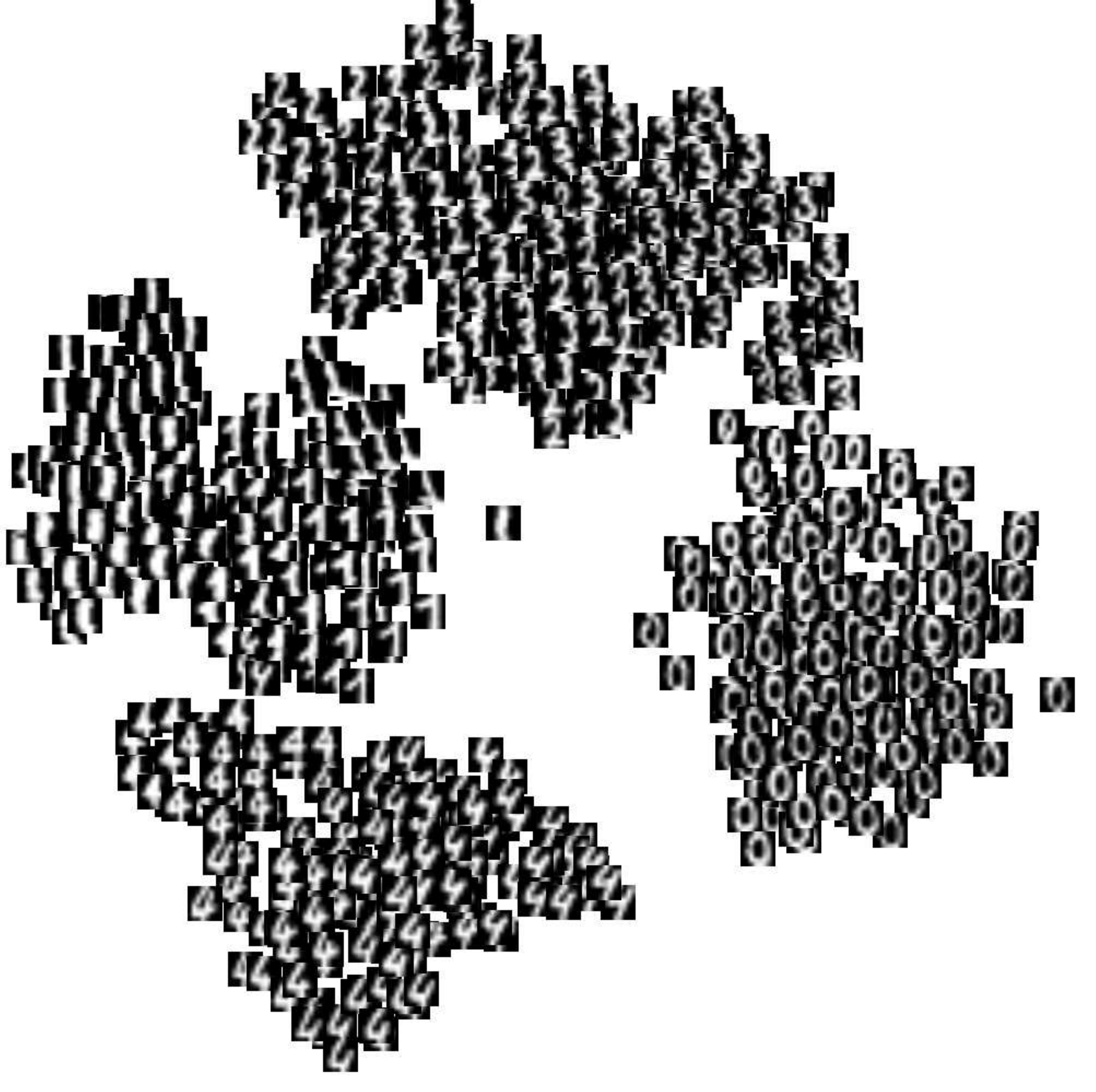}
    }
    \caption
    {
        Simplification of the data set \emph{MNIST}.
        (a-d): The simplified data sets by detecting the first
            $5, 10, 20, 30$ LBO eigenvectors,
            which consist of $82, 120, 196, 280$ data points.
        (e,f): The DR result
            (using T-SNE~\cite{maaten2008visualizing}) of the original data set,
            illustrated by colorful digits (e) and original images (f).
   }
   \label{fig:hand}
\end{figure}

 \textbf{MNIST:}
 For clarity of the visual analysis,
     we selected $900$ units of the gray images for the digits '0' to '4',
     from the \emph{MNIST} data set,
     and resized each image as $8 \times 8$.
 The selected handwritten digit images were projected into
     the 2D-plane using the DR method T-SNE~\cite{maaten2008visualizing}.
  Whereas in Fig.~\ref{subfig:hand_color_replaced},
    colorful digits are placed at the projected 2D points,
    in Fig.~\ref{subfig:hand_image_replaced},
    the images themselves are located at the projected points.
  The simplified data sets with $82,120, 196, 280$ data points,
    which were generated by detecting the feature points of the first $5, 10, 20, 30$ LBO eigenvectors
    are displayed in Fig.~\ref{subfig:hand_5}-~\ref{subfig:hand_30}.
  For illustration,
    their positions in the 2D-plane
    are extracted from the dimensionality reduced data set (Fig.~\ref{subfig:hand_color_replaced}).
  In the simplified data set in Fig.~\ref{subfig:hand_5},
    which has $82$ data points (approximately $9\%$ of the points of the original data set),
    all five categories of the digits are represented.
  In the simplified data set in Fig.~\ref{subfig:hand_20},
    which has $196$ data points (approximately $22\%$ of the points of the original data set),
    virtually all of the key features of the data set are represented,
    including the subset of digit '2' in the class of digit '1'.
  In Fig.~\ref{subfig:hand_30},
    the simplified data set contains $280$ data points (approximately $31\%$ of the points of the original data set).
  We can observe that the boundary of each class of data points has formed
    (Fig.~\ref{subfig:hand_30}),
    and the geometric shape of each class is similar to that of the corresponding class in the original data set (Fig.~\ref{subfig:hand_color_replaced}).

\begin{figure}[!htb]
   \centering
    \subfigure[]
    {
        \label{subfig:face_5}
        \includegraphics[width=0.22\textwidth]{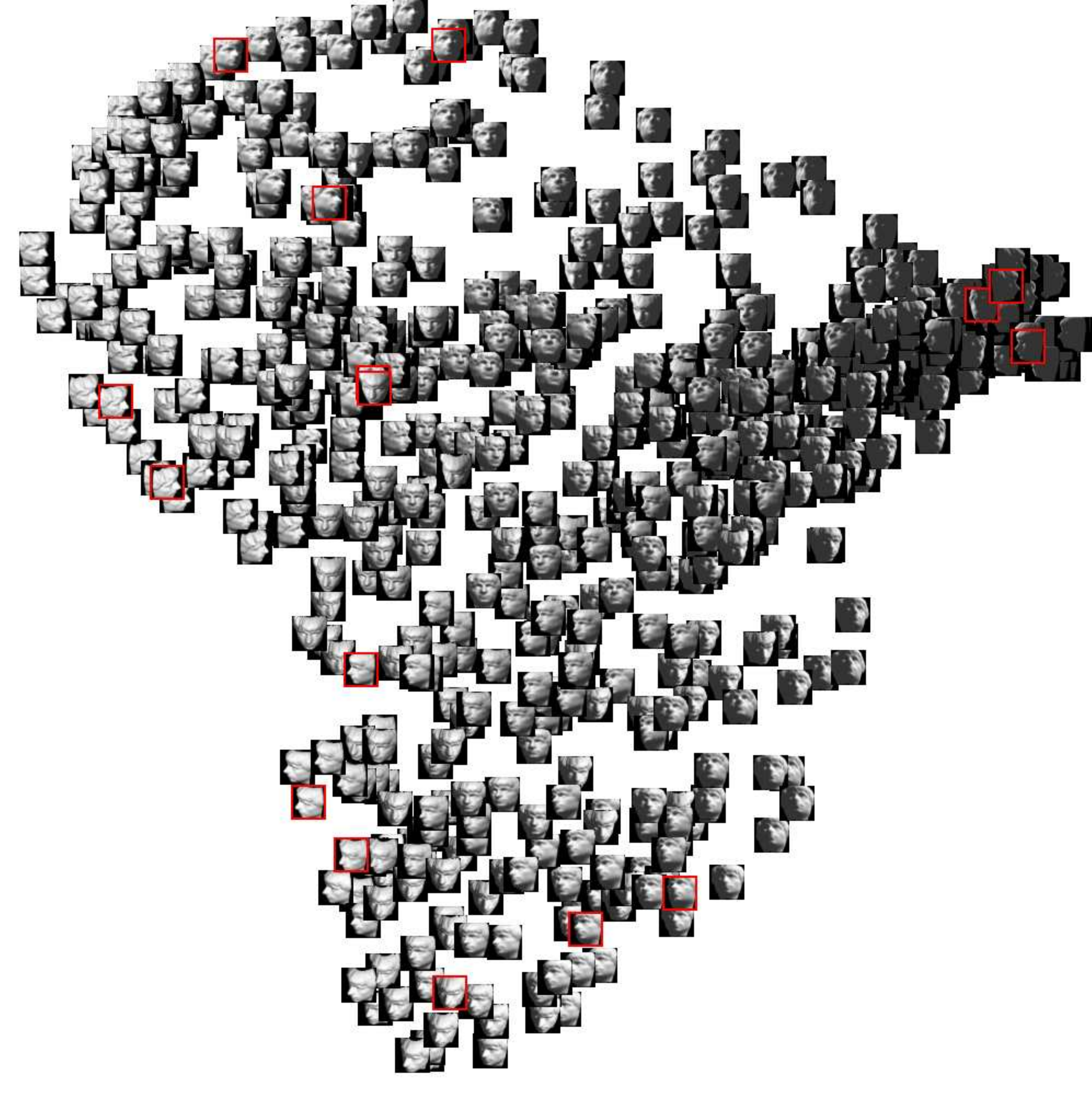}
    }
    \subfigure[]
    {
        \label{subfig:face_10}
        \includegraphics[width=0.22\textwidth]{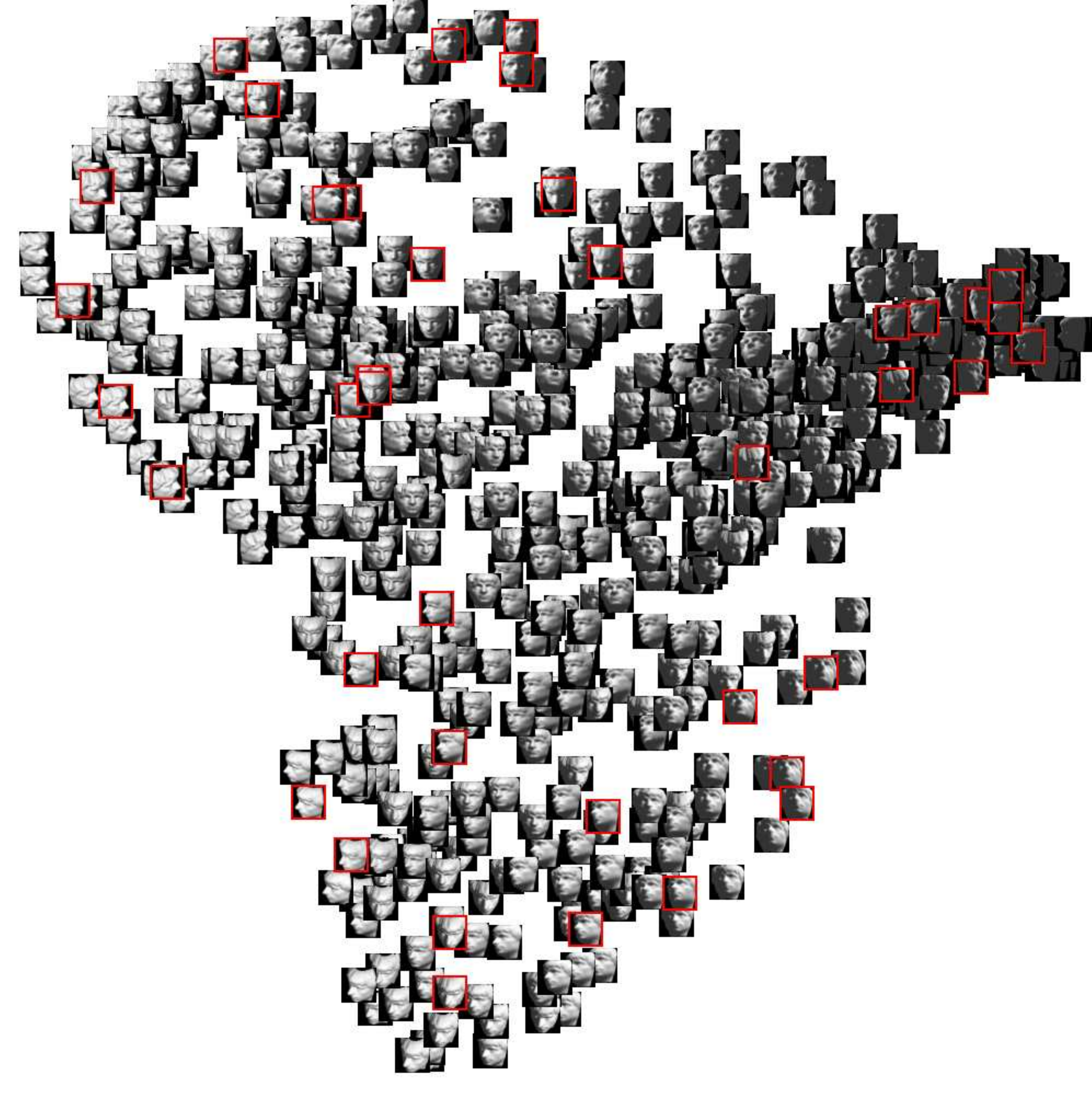}
    }
    \subfigure[]
   {
        \label{subfig:face_20}
        \includegraphics[width=0.22\textwidth]{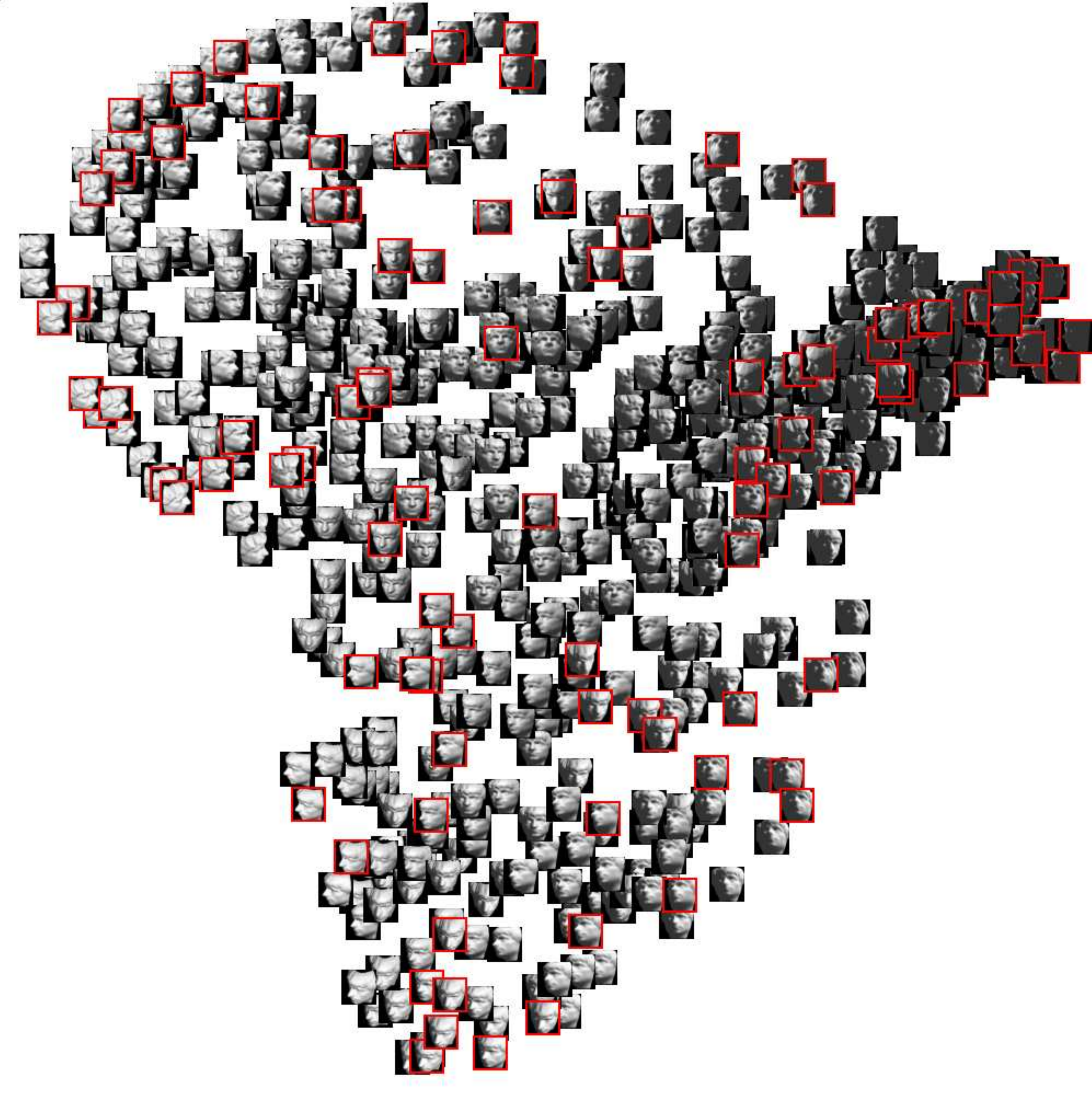}
   }
   \subfigure[]
  {
        \label{subfig:face_30}
        \includegraphics[width=0.22\textwidth]{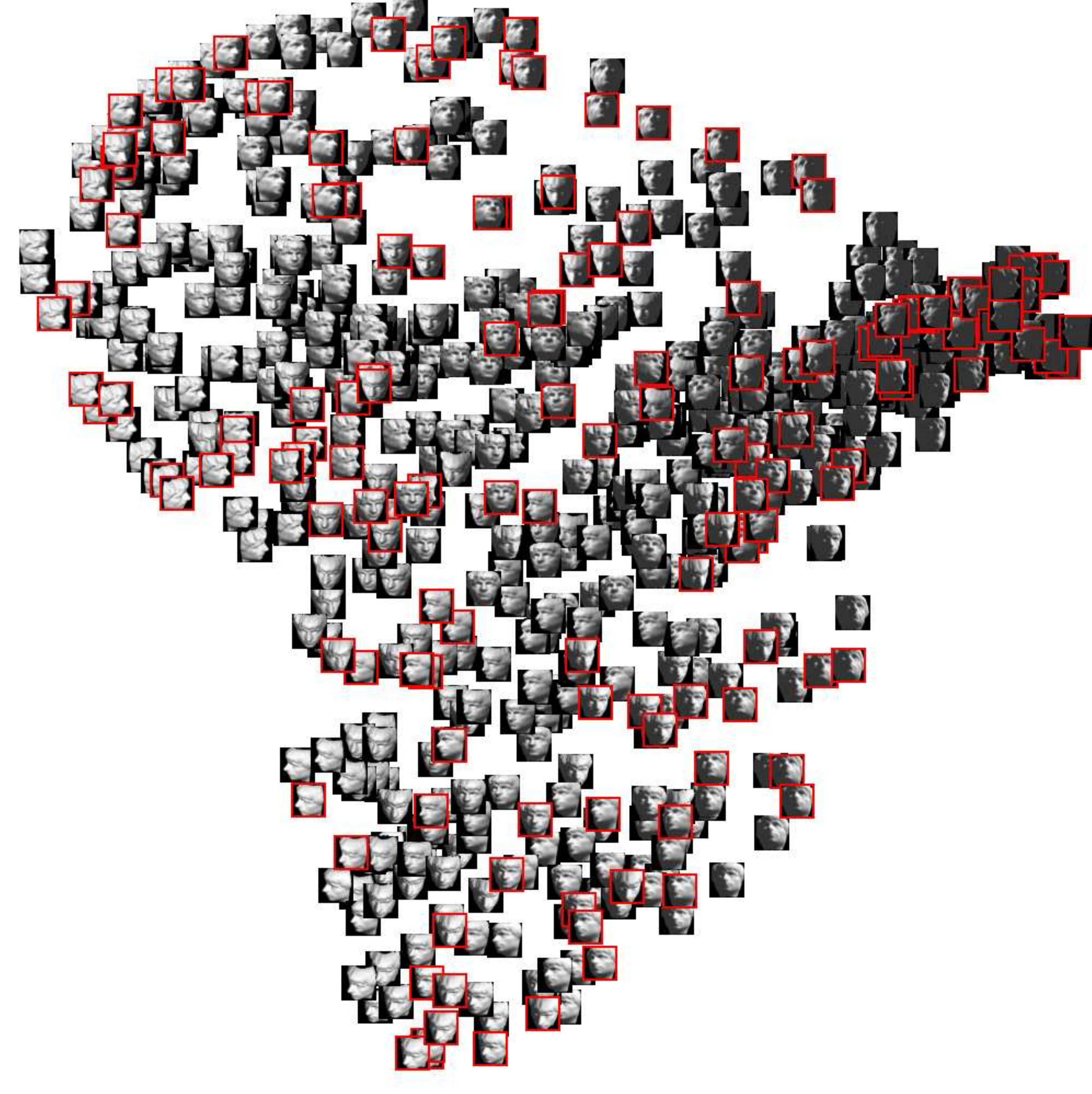}
  }
  \subfigure[]
  {
        \label{subfig:face_50}
        \includegraphics[width=0.22\textwidth]{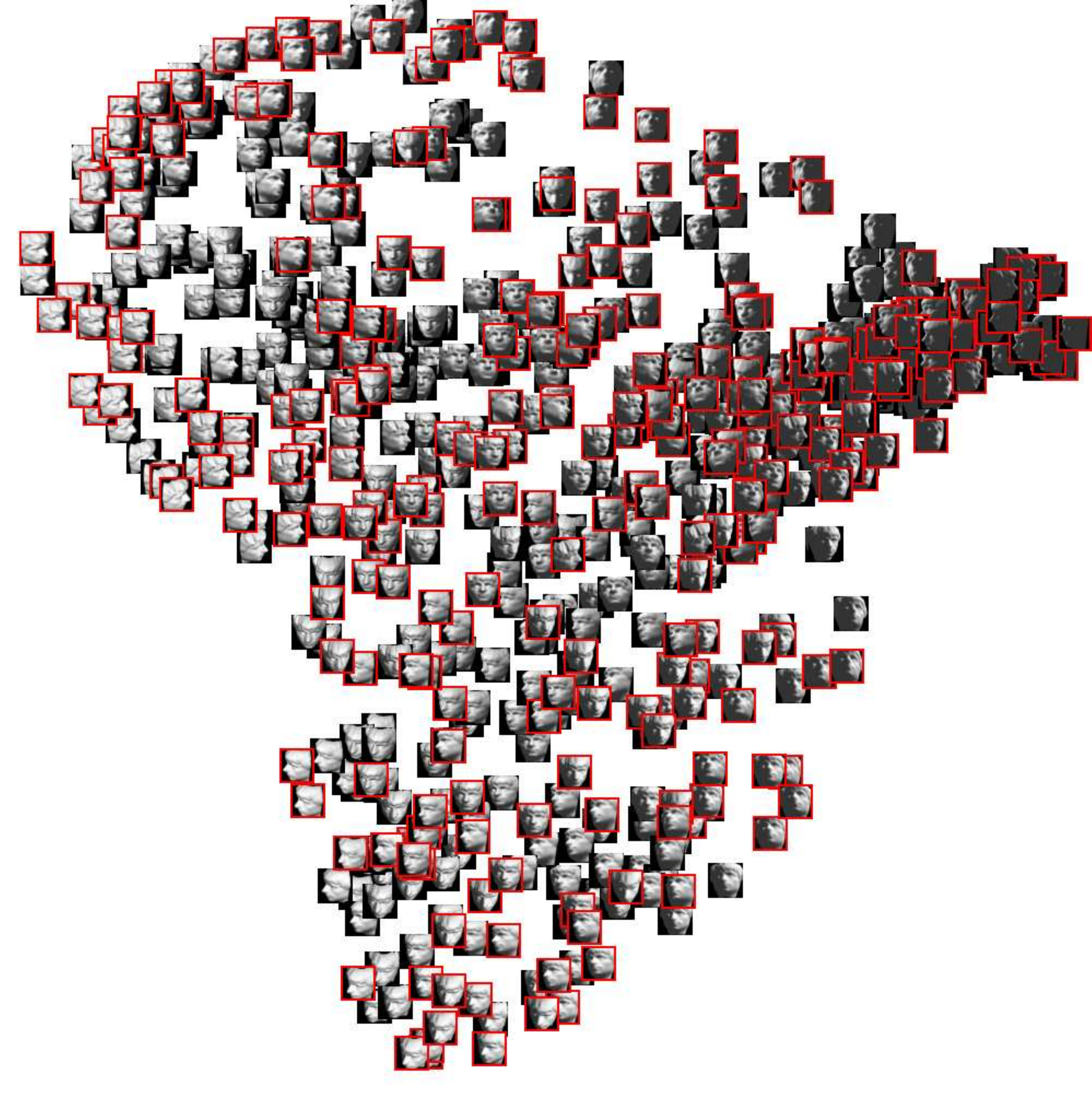}
  }
  \subfigure[]
  {
        \label{subfig:face_eigen}
        \includegraphics[width=0.22\textwidth]{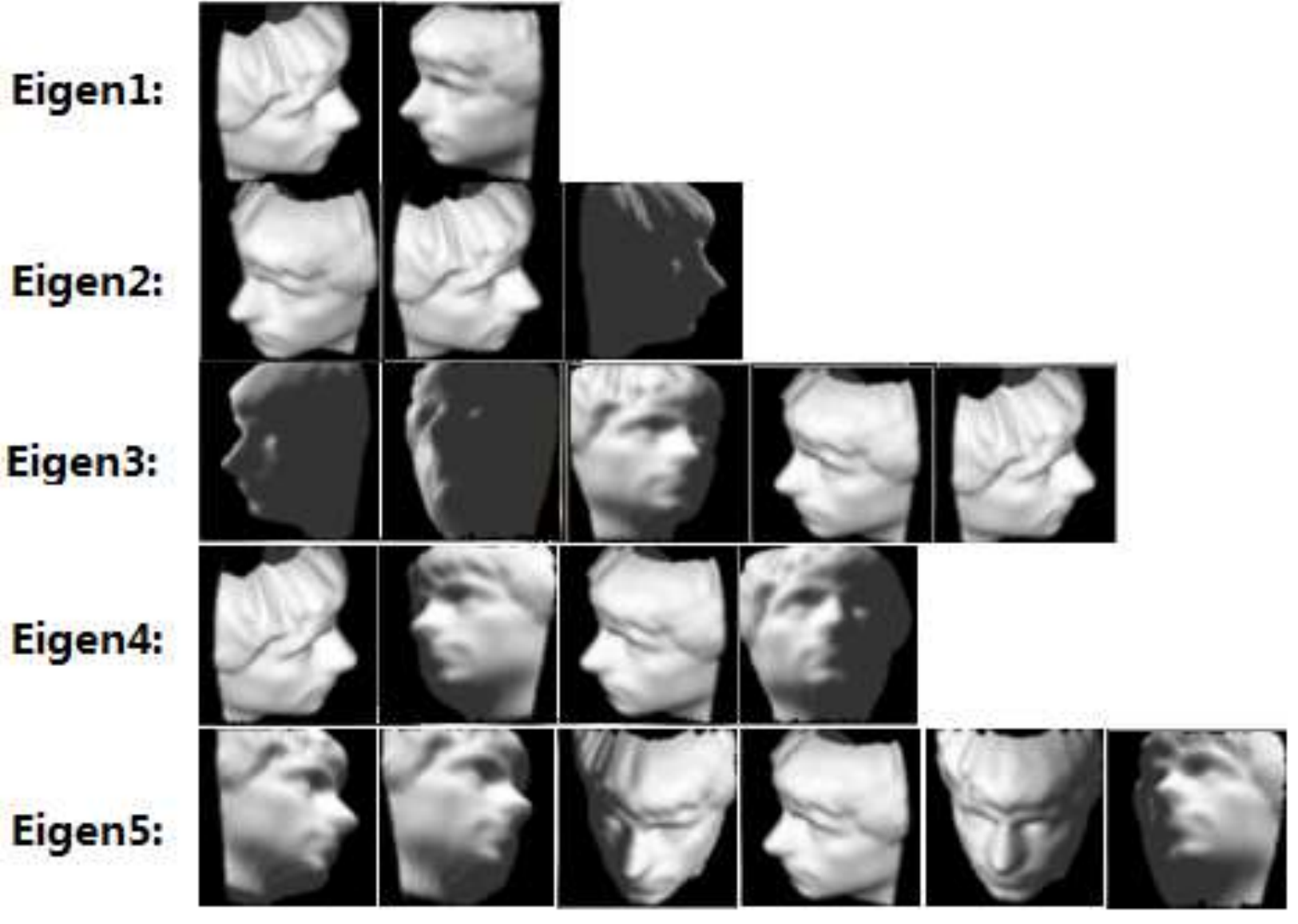}
  }
  \caption
  {
    The simplified data sets of the \emph{human face} data set,
        after DR by PCA~\cite{jolliffe2011principal}.
    Each simplified data set is comprised of the images
        marked with red frames.
    (a-e) The simplified data sets by detecting the first $5, 10, 20, 30, 50$ LBO eigenvectors,
        which consist of $16, 40, 95, 150, 270$ data points, respectively.
    (f) The feature points of the first $5$ eigenvectors.
  }
  \label{fig:facedetection}
\end{figure}

 \textbf{Human face}:
 The \emph{human face} data set (Fig.~\ref{subfig:face})
    contains $698$ gray-images($64 \times 64$) of a plaster head sculpture.
 The images of the data set were captured by adjusting three parameters:
    up-down pose, left-right pose, and lighting direction.
 Therefore, the images in the data set can be regarded as
    distributing on a 3D-manifold.
 Thus, as illustrated in Fig.~\ref{fig:facedetection},
    after the \emph{human face} data set is dimensionally reduced into
    a 2D-plane using PCA~\cite{jolliffe2011principal},
    it can observed that
    the longitudinal axis represents the sculpture's rotation in the right-left direction,
    and the horizontal axis indicates the change of lighting.

 In Fig.~\ref{fig:facedetection},
    the images marked with red frames constitute the simplified data set.
 The simplified data sets in Figs.~\ref{subfig:face_5}-~\ref{subfig:face_50}
    which consist of $16, 40, 95, 150, 270$ data points,
    are generated by detecting the feature points of
    the first $5, 10, 20, 30, 50$ LBO eigenvectors, respectively.
 In the simplified data set in Fig.~\ref{subfig:face_5},
    the critical points at the boundary are captured.
 In Figs.~\ref{subfig:face_10}-~\ref{subfig:face_30},
    increasingly more critical points at the boundary and interior are added into the simplified data sets.
 In Fig.~\ref{subfig:face_50},
    virtually all of the images at the boundary are added into the simplified set.
 For clarity,
    we display the feature points of the first five LBO eigenvectors
    in Fig.~\ref{subfig:face_eigen},
    which indeed capture several 'extreme' images in the \emph{human face} data set.
 Specifically,
    the two feature points of the first eigenvector, $\phi_1$,
    are two images with 'leftmost pose + most lighting' and 'rightmost pose + most lighting'.
 In the feature point set of the second eigenvector, $\phi_2$,
    an image with 'leftmost pose + fewest lighting' is added.
 Similarly,
    in the feature point sets of the eigenvectors $\phi_3, \phi_4$, and $\phi_5$,
    there are 'extreme' images of the pose and lighting (Fig.~\ref{subfig:face_eigen}).

 \textbf{CIFAR-10}:
 The data set CIFAR-10 (Fig.~\ref{subfig:cifar10}) was also simplified by
    the proposed method.
 Similarly,
    increasingly more critical points of the data set were added into the
    simplified data set as a consequence of the simplification procedure.

 The fidelity of the simplified data sets was quantitatively measured as per
    Section~\ref{subsec:quantitative_analysis}.
 In Table~\ref{tab:simplification},
    The running time in seconds of the proposed method is listed.
 It can be seen that
    the operations of \emph{KNN} and eigen-solving consume the majority of the time,
    and the data simplification time ranged from $0.044$ to
    $4.759$ seconds.

 \begin{table}[htbp]
 \caption{\label{tab:simplification}Running time (seconds) of the developed method}
 \begin{tabular}{ccccc}
    \toprule
    Data set   & Data size  & \emph{KNN}    & Eigen    & Simplification  \\
    \midrule
    Swiss roll & $2000 \times 3$          & 5.535    & 6.986       &0.044\\
    MNIST      & $60000 \times 784$       & 57.579   & 178.933     &6.544\\
    Human face & $698 \times 4096$        & 3.604    & 1.079       &0.035\\
    CIFAR-10    & $50000 \times 3072$      & 115.919  & 146.113     &4.759\\
   \bottomrule
  \end{tabular}
 \end{table}
 \subsection{Quantitative Measurement of the Fidelity}
 \label{subsec:quantitative_analysis}

\begin{figure}[!htb]
   \centering
     \includegraphics[width=0.45\textwidth]{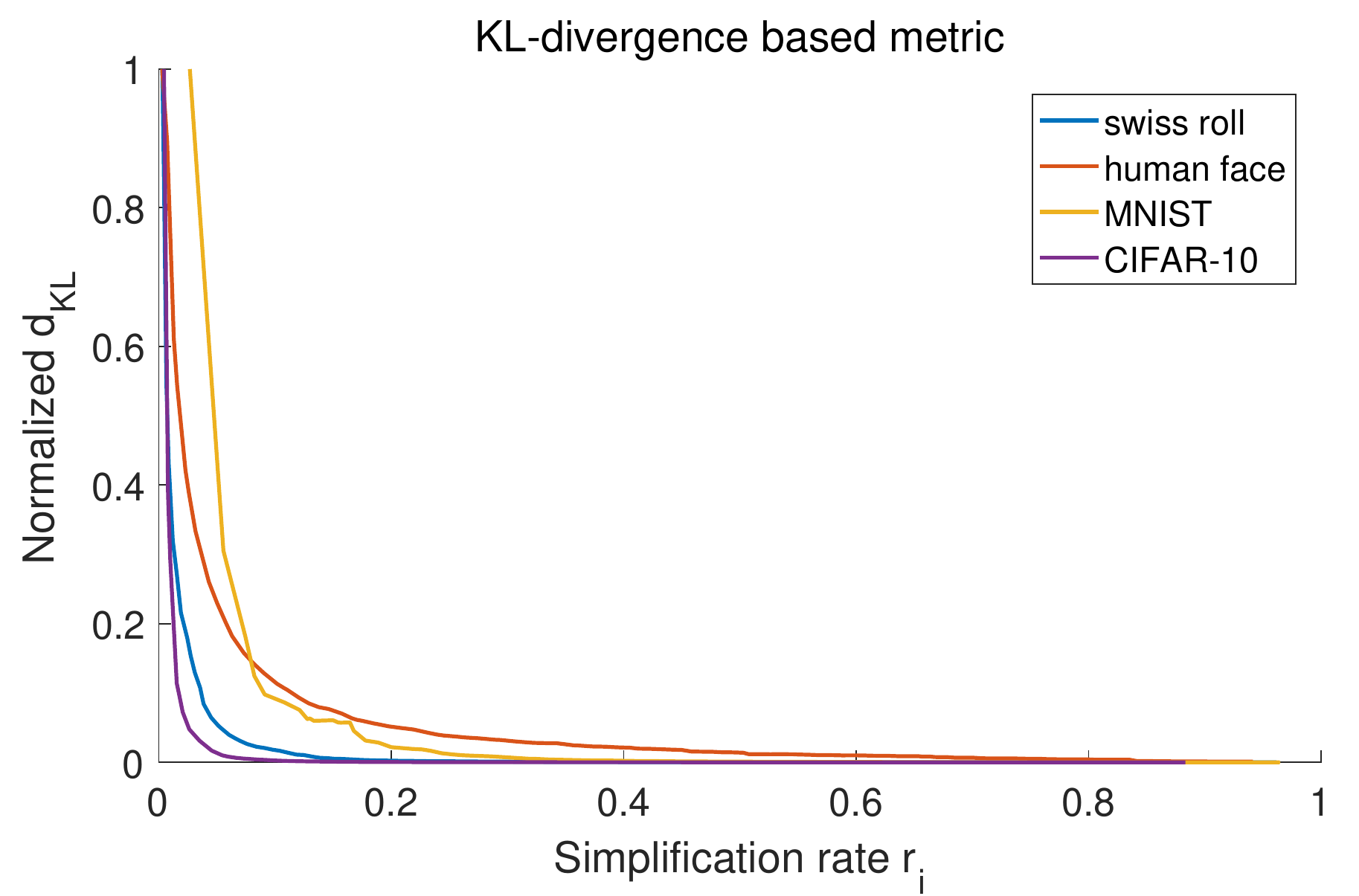}
   \caption
   {
        Diagram of $r_i = \frac{Card(\mathcal{S}_i)}{Card(\mathcal{H})}$ v.s. the normalized $d_{KL}(\mathcal{H},\mathcal{S}_i), i=1,2,\cdots,n$.
    }
    \label{fig:KLD}
\end{figure}

\begin{figure}[!htb]
      \centering
      \includegraphics[width=0.45\textwidth]{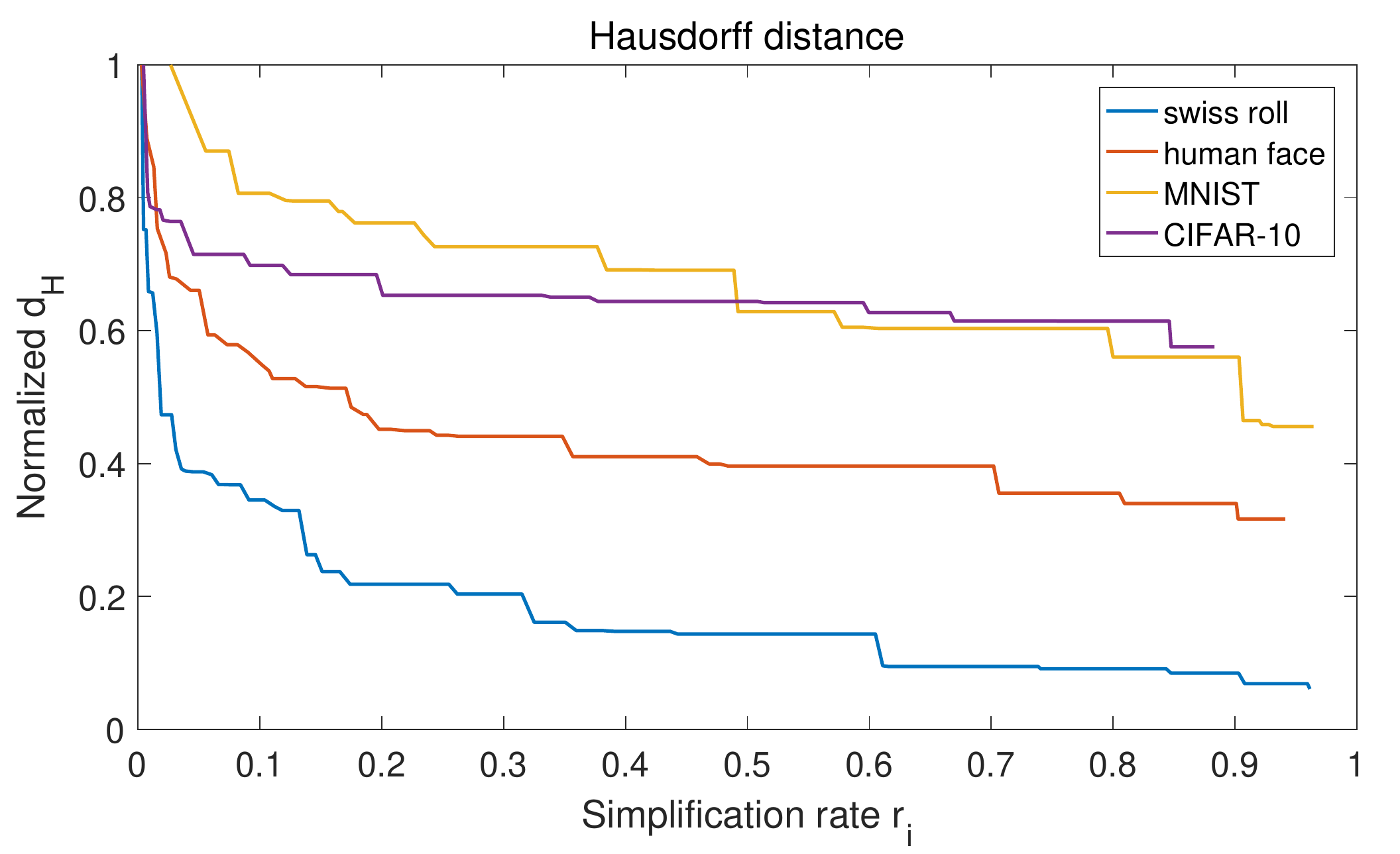}
      \caption
      {
        Diagram of $r_i = \frac{Card(\mathcal{S}_i)}{Card(\mathcal{H})}$ v.s. the normalized $d_{H}(\mathcal{H},\mathcal{S}_i), i=1,2,\cdots,n$.
      }
      \label{fig:HD}
\end{figure}

\begin{figure}[!htb]
      \centering
      \includegraphics[width=0.45\textwidth]{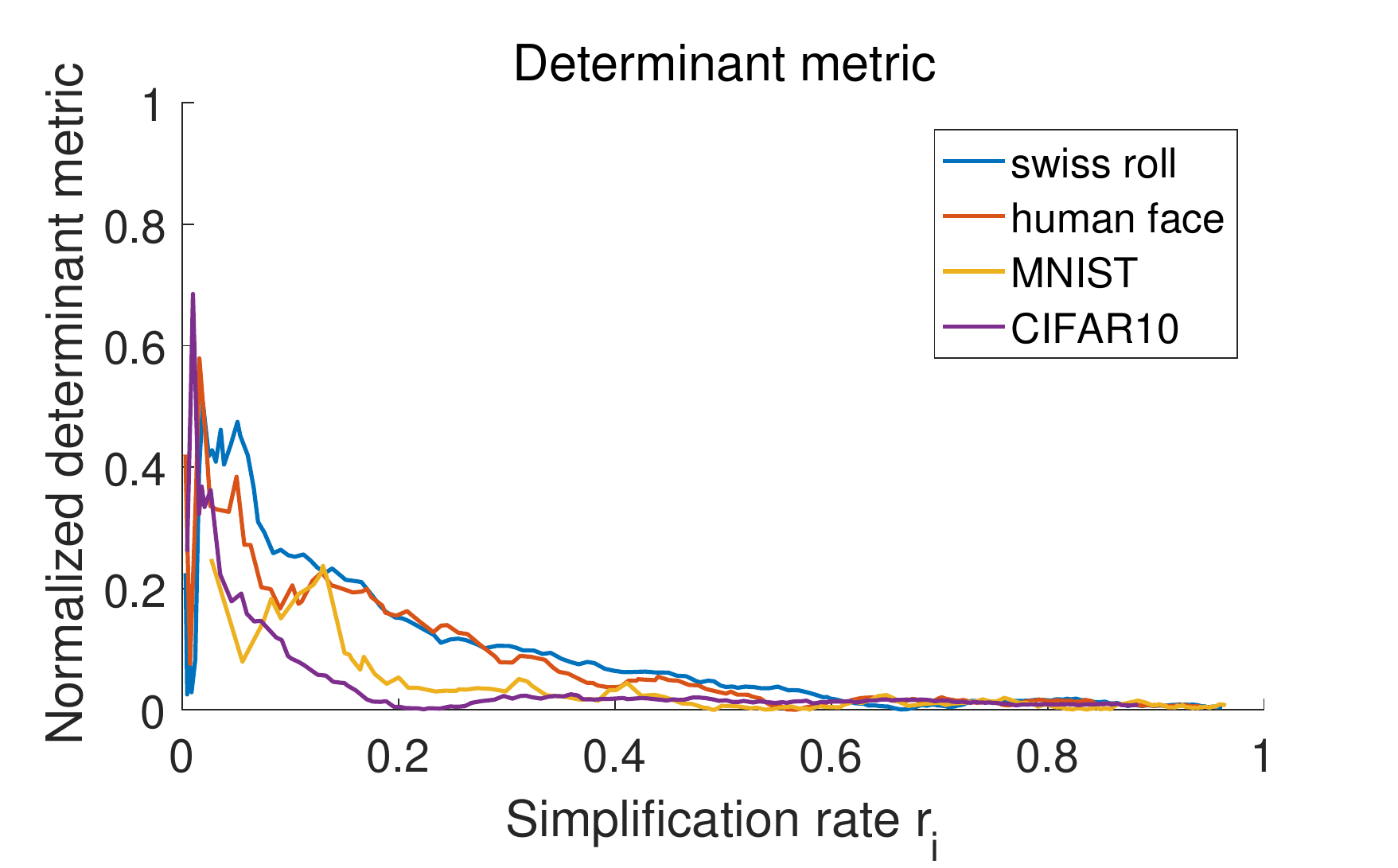}
      \caption
      {
        Diagram of $r_i = \frac{Card(\mathcal{S}_i)}{Card(\mathcal{H})}$ v.s. the normalized $d_{COV}(\mathcal{H},\mathcal{S}_i), i=1,2,\cdots,n$.
      }
      \label{fig:CO}
\end{figure}

 In this section, the fidelity of the simplified data set $\mathcal{S}$
    to the original data set $\mathcal{H}$ is quantitatively measured by the three metrics developed in Section~\ref{subsec:metrics}, i.e.,
    KL-divergence-based metric $d_{KL}(\mathcal{H},\mathcal{S})$~\pref{eq:kl_metric},
    Hausdorff distance $d_{H}(\mathcal{H},\mathcal{S})$~\pref{eq:Hausdorff_distance},
    and determinant of covariance matrix $d_{COV}(\mathcal{H},\mathcal{S})$~\pref{eq:covariance}.
 Specifically,
    for the data simplification procedure of
    each of the four data sets mentioned above,
    a series of simplified data sets $\mathcal{S}_i, i = 1,2,\cdots,n$ was first generated.
 Then, we calculated the \emph{simplification rate} $r_i$,
    i.e., the ratio between the cardinality of each data set
    $\mathcal{S}_i$ and that of the original data set $\mathcal{H}$,
    i.e.,
 \begin{equation} \label{eq:ratio}
    r_i = \frac{Card(\mathcal{S}_i)}{Card(\mathcal{H})},
 \end{equation}
    and the three metrics
 \begin{equation*}
    d_{KL}(\mathcal{H},\mathcal{S}_i), d_{H}(\mathcal{H},\mathcal{S}_i),\
    \text{and}\ d_{COV}(\mathcal{H},\mathcal{S}_i), i=1,2,\cdots,n,
 \end{equation*}
    which were normalized into $[0,1]$.
 Finally, the diagrams of
    $r_i$ v.s. $d_{KL}(\mathcal{H},\mathcal{S}_i)$,
    $r_i$ v.s. $d_{H}(\mathcal{H},\mathcal{S}_i)$,
    and  $r_i$ v.s. $d_{COV}(\mathcal{H},\mathcal{S}_i), i=1,2,\cdots,n$
    were plotted as displayed in Figs.~\ref{fig:KLD}, \ref{fig:HD}, and \ref{fig:CO},
    respectively.

 \textbf{KL-divergence-based metric}:
    $d_{KL}(\mathcal{H},\mathcal{S})$~\pref{eq:kl_metric} measures the similarity of the distributions of the simplified data set $\mathcal{S}$ and the original data set $\mathcal{H}$.
 If $\mathcal{S}$ is exactly the same as $\mathcal{H}$,
    then $d_{KL}(\mathcal{H},\mathcal{S}) = 0$.

 Fig.~\ref{fig:KLD} displays the diagrams of
    $r_i = \frac{Card(\mathcal{S}_i)}{Card(\mathcal{H})}$ v.s.
    the normalized $d_{KL}(\mathcal{H},\mathcal{S}_i), i=1,2,\cdots,n$ of the four data sets.
 It can be observed that
    in the simplification procedures of all of four data sets,
    the metric $d_{KL}(\mathcal{H},\mathcal{S})$ is rapidly reduced to a small value (less than 0.1),
    when $r_i = 0.1$ , approximately,
    i.e., the simplified data set contains approximately $10\%$ of the data points of the original data set.
 Using the data set \emph{swiss roll} as an example (blue line)
    when the simplified data set $\mathcal{S}$ is composed of $5\%$ of the data points of the original data set $\mathcal{H}$
    (i.e., $r_i = 0.05$),
    the normalized $d_{KL}(\mathcal{H},\mathcal{S})$ is less than $0.1$;
    when $r_i = 0.2$ ($\mathcal{S}$ contains $20\%$ of the data points of $\mathcal{H}$),
    the normalized $d_{KL}(\mathcal{H},\mathcal{S})$ is close to $0$,
    meaning that the simplified data set $\mathcal{S}$ contains virtually all of the information embraced by the original data set $\mathcal{H}$.

 \textbf{The Hausdorff distance}
    $d_H(\mathcal{H},\mathcal{S})$~\pref{eq:Hausdorff_distance} measures the geometric distance between two sets.
 In Fig.~\ref{fig:HD},
    the diagram between
    \begin{equation*}
        r_i =  \frac{Card(\mathcal{S}_i)}{Card(\mathcal{H})}\ \text{v.s. the normalized}\ d_{H}(\mathcal{H},\mathcal{S}_i), i=1,2,\cdots,n
    \end{equation*}
    is illustrated.
 Because the Hausdorff distances between the original data set $\mathcal{H}$
    and two adjacent simplified data sets,
    e.g., $d_{H}(\mathcal{H},\mathcal{S}_i)$ and $d_{H}(\mathcal{H},\mathcal{S}_{i+1}), i=1,2,\cdots,n-1$,
    are possibly the same,
    the diagrams demonstrated in Fig.~\ref{fig:HD} are step-shaped.
 Based on Fig.~\ref{fig:HD},
    in the data simplification procedures of the four data sets,
    the Hausdorff distances $d_{H}(\mathcal{H},\mathcal{S})$ decrease when
    increasingly more data points are contained in the simplified data sets.
 Moreover,
    the convergence speed of $d_H(\mathcal{H},\mathcal{S}_i)$ with respect to
    $r_i$~\pref{eq:ratio} is related to the intrinsic dimension of the data set:
 The lower the intrinsic dimension of a data set,
    the faster the convergence speed.
 For example (refer to Fig.~\ref{fig:HD}),
    the intrinsic dimension of the data set \emph{swiss roll} is $2$
    (the smallest of the four data sets),
    and the convergence speed of the diagram for the simplification procedure of \emph{swiss roll} is the fastest.
 The intrinsic dimension of the data set \emph{human face} is $3$,
    greater than that of \emph{swiss roll},
    and the convergence speed of its diagram is slower than that of \emph{swiss roll}.
 The intrinsic dimensions of the data sets \emph{MNIST} and \emph{CIFAR-10}
    are greater than those of \emph{swiss roll} and \emph{human face},
    and the convergence speeds of their diagrams are the slowest.

 \textbf{The determinant of covariance matrix} of a data set reflects
    its 'volume'.
 The metric $d_{COV}(\mathcal{H}, \mathcal{S})$~\pref{eq:covariance} measures
    the difference between the 'volume' of the original data set $\mathcal{H}$
    and that of the simplified data set $\mathcal{S}$.
 In Fig.~\ref{fig:CO},
    the diagram of $r_i$~\pref{eq:ratio} v.s.
    the normalized $d_{COV}(\mathcal{H},\mathcal{S}_i), i=1,2,\cdots,n$
    is illustrated.
 In the region with small $r_i$~\pref{eq:ratio},
    the simplified data set contains a small number of data points and
    the diagrams oscillate.
 This is because the 'volume' of a simplified data set
    with a small number of data points is heavily influenced by each of its points,
    and adding merely one point into the set can change its 'volume' significantly.
 However,
    with increasingly more data points inserted into
    the simplified data set (i.e., with increasingly greater $r_i$),
    the diagrams of all four data sets converge to zero.
 Considering the data set \emph{CIFAR-10} as an example,
    when the simplified data set contains $20\%$ of the data points of the original data set ($r_i = 0.2$),
    the metric $d_{COV}(\mathcal{H}, \mathcal{S})$~\pref{eq:covariance} is nearly zero,
    meaning that the 'volume' of the simplified data set has approached that of the original data very closely.

%
%
%

\subsection{Applications}

 In this section, two applications of the proposed data set simplification
    method are demonstrated; there are
    the speedup of DR
    and training data simplification in supervised learning.

  \textbf{Speedup of DR}:
 The computation of the DR algorithms
    for a high-dimensional data set is typically complicated.
 Given an original data set $\mathcal{H}$ ($Card(\mathcal{H})=n$),
    if we first perform the DR on the simplified data
    set $\mathcal{S}$ ($Card(\mathcal{S})=m$),
    and then employ the 'out-of-sample' algorithm~\cite{bengio2004out} to calculate the DR coordinates of the remaining data points,
    the computation requirement can be reduced significantly.
 Because the simplified data set $\mathcal{S}$
    captures the feature points of the original data set,
    the result generated by \emph{DR on simplified set + 'out-of-sample'} is similar as that by \emph{DR on original set}.

 \begin{figure}[!htb]
   \centering
   \subfigure[]
   {
        \label{subfig:face_10_simplify}
        \includegraphics[width=0.20\textwidth]{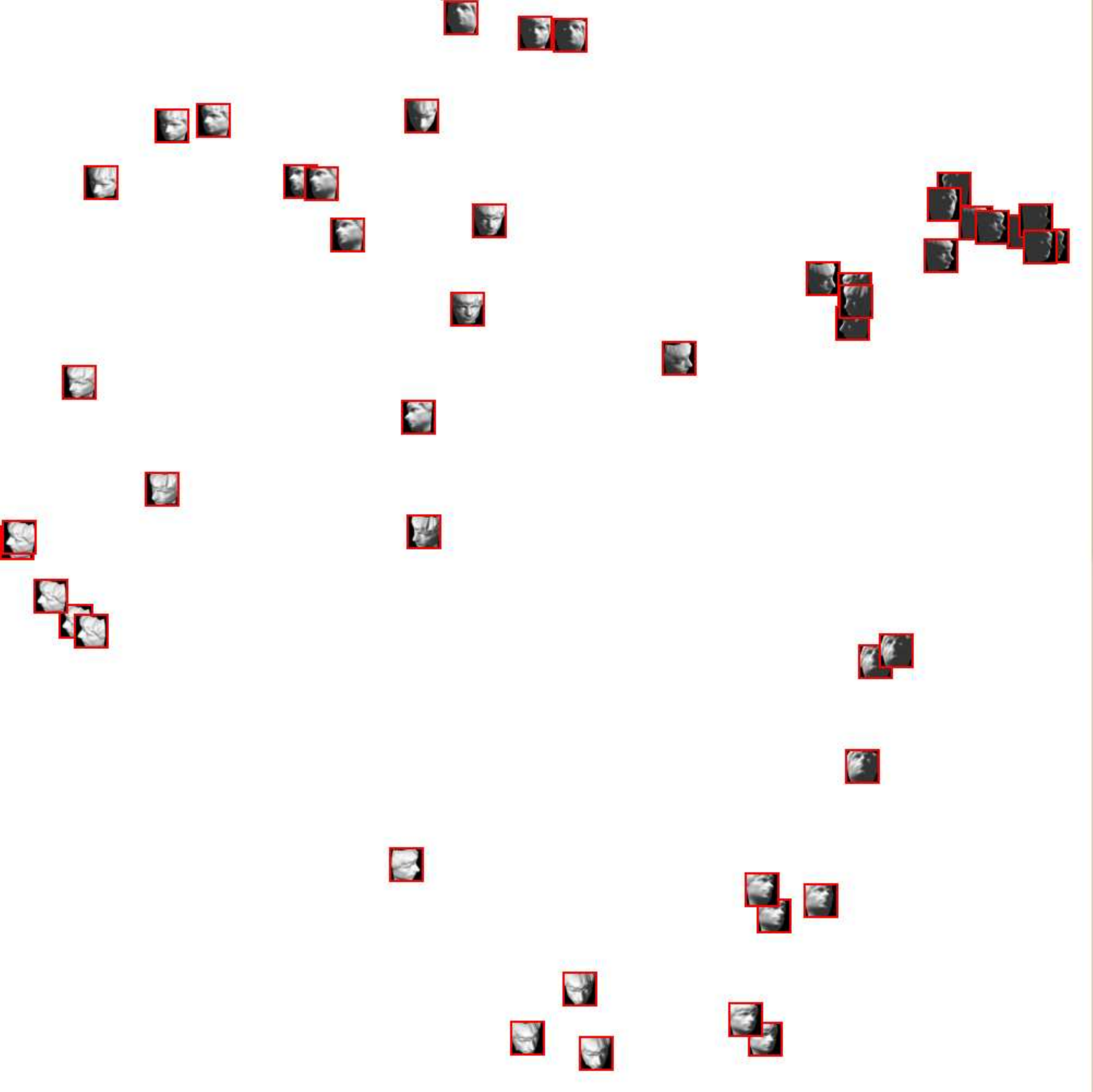}
   }
   \subfigure[]
   {
        \label{subfig:face_10_org}
        \includegraphics[width=0.20\textwidth]{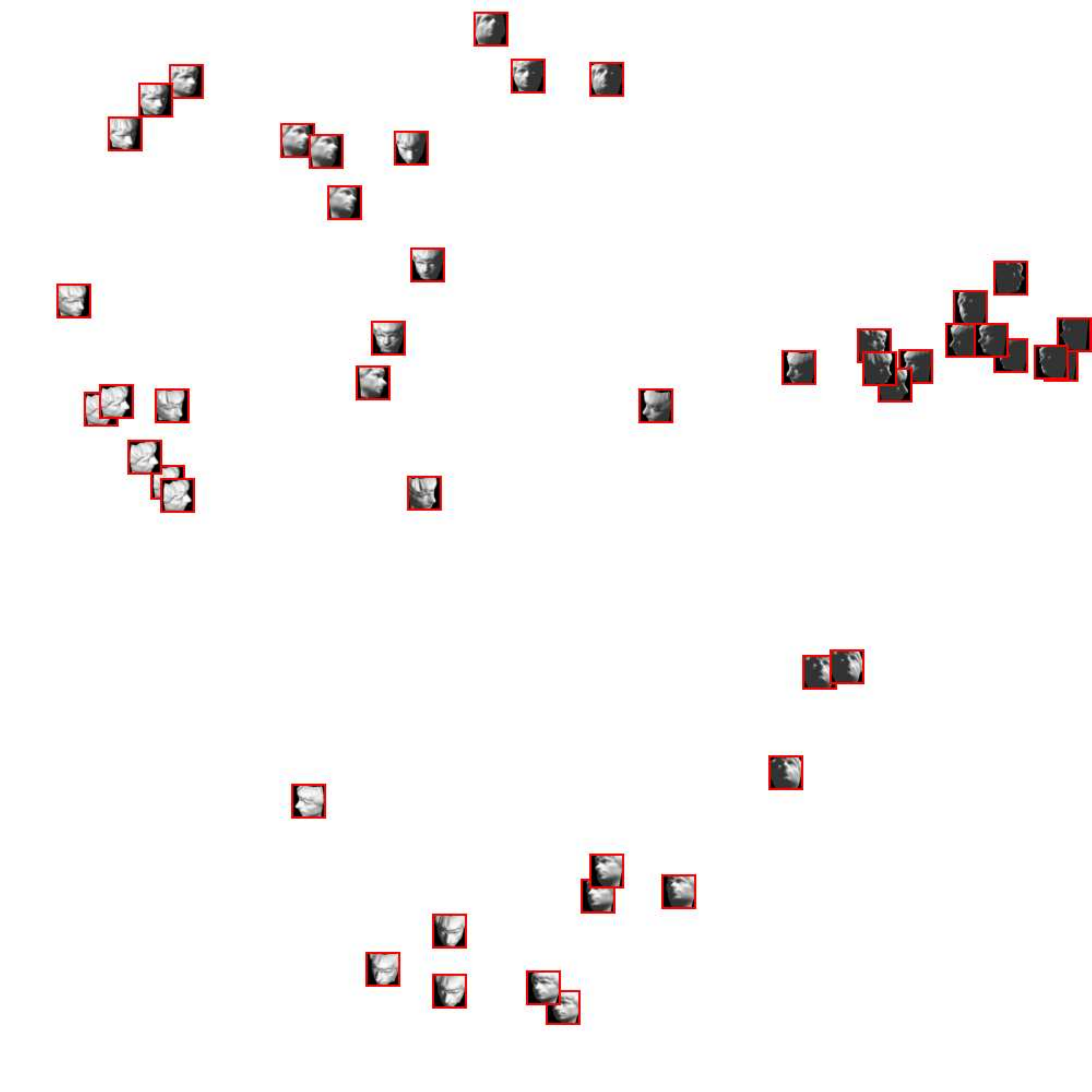}
   }
   \subfigure[]
   {
        \label{subfig:face_20_simplify}
        \includegraphics[width=0.20\textwidth]{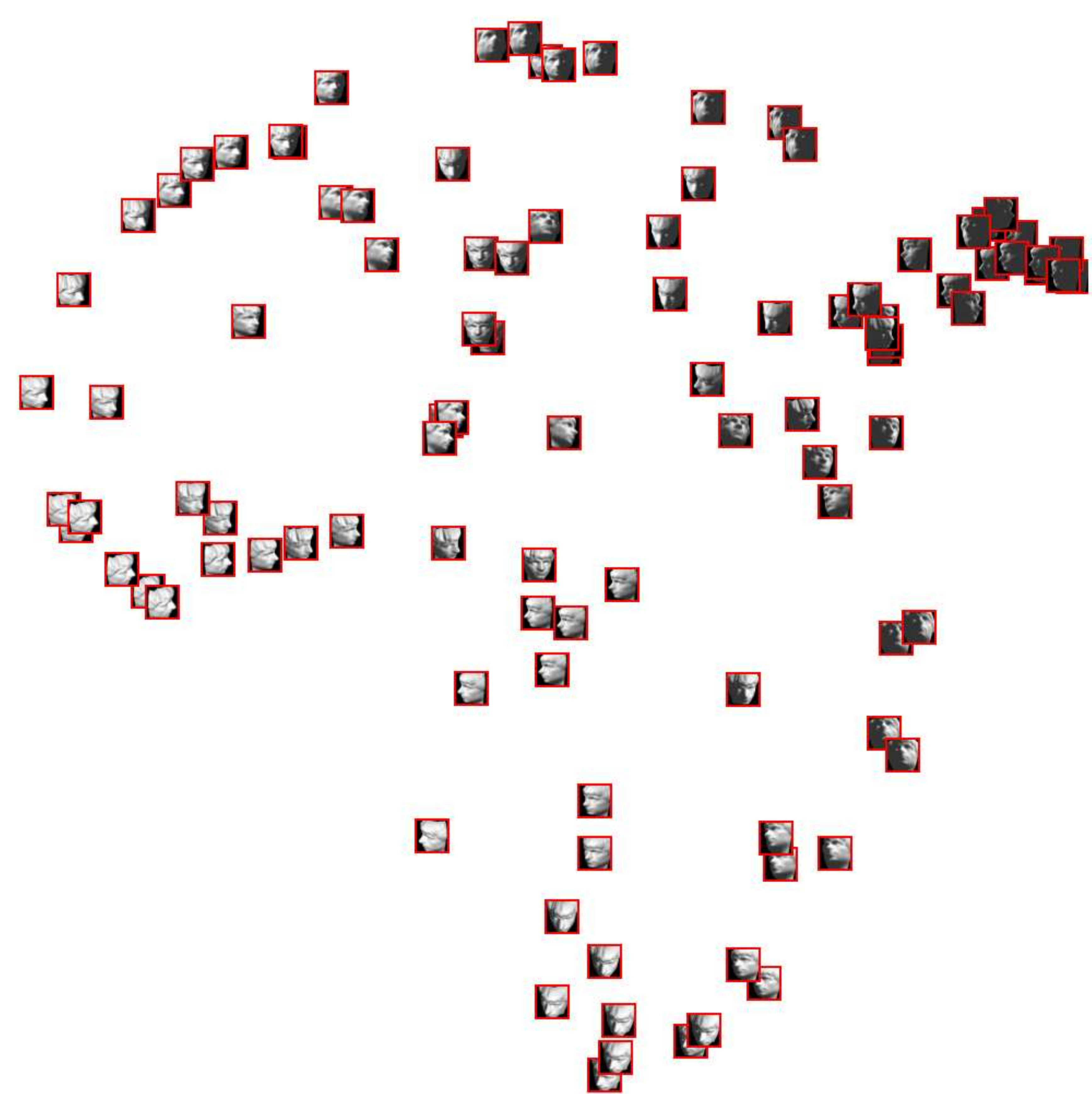}
   }
   \subfigure[]
   {
        \label{subfig:face_20_org}
        \includegraphics[width=0.20\textwidth]{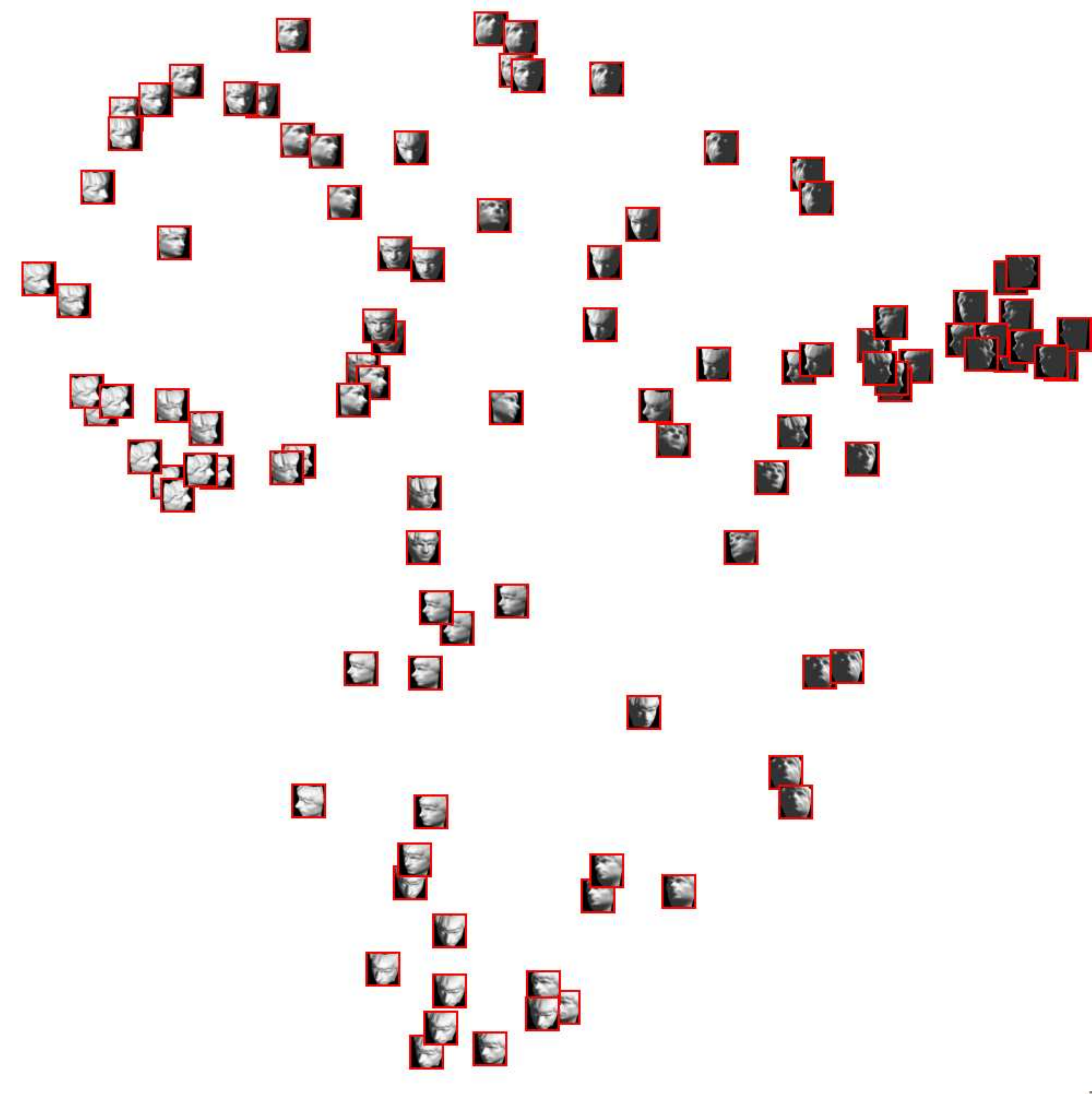}
   }
   \subfigure[]
   {
        \label{subfig:face_50_simplify}
        \includegraphics[width=0.22\textwidth]{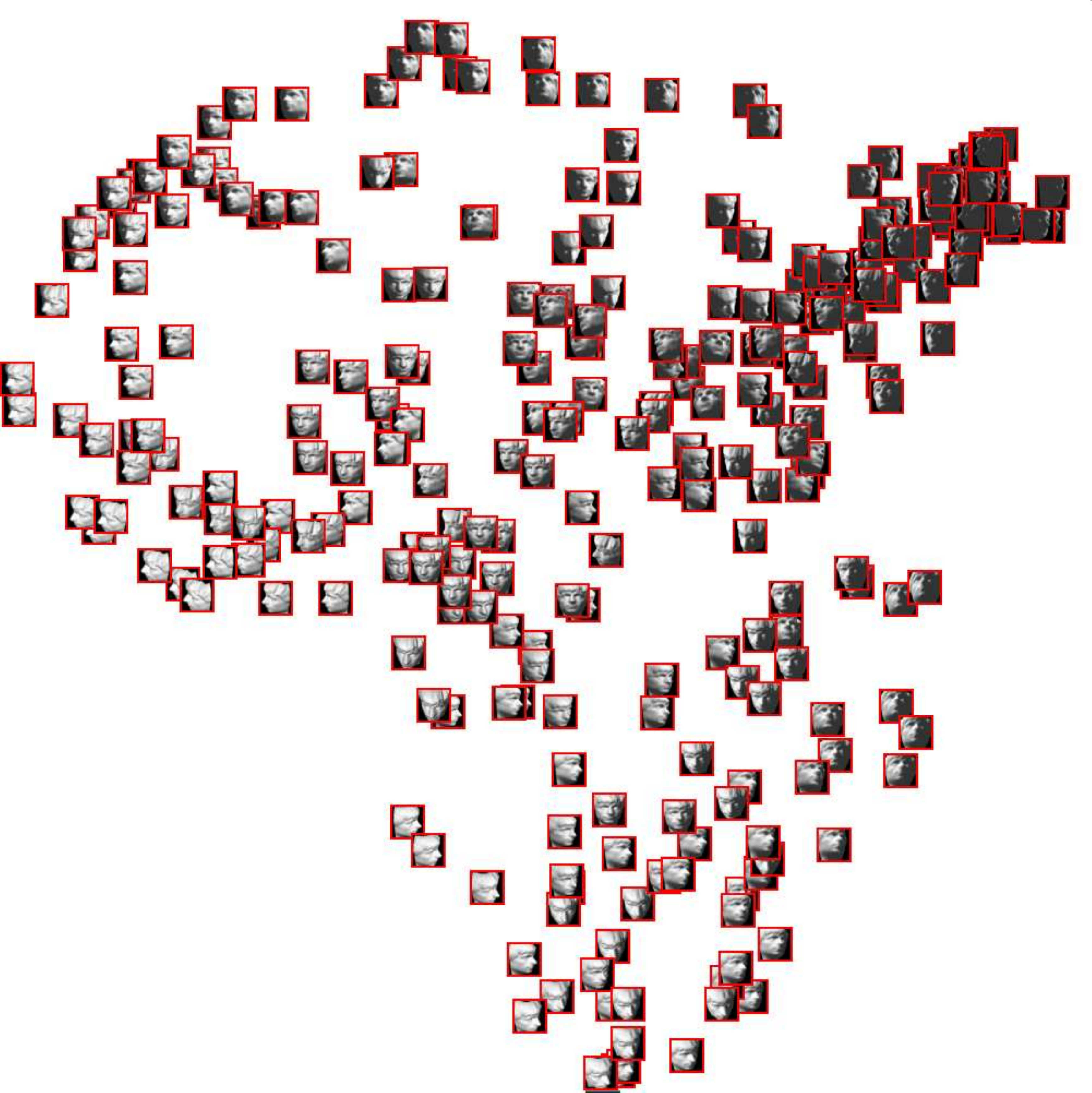}
   }
   \subfigure[]
   {
        \label{subfig:face_50_org}
        \includegraphics[width=0.22\textwidth]{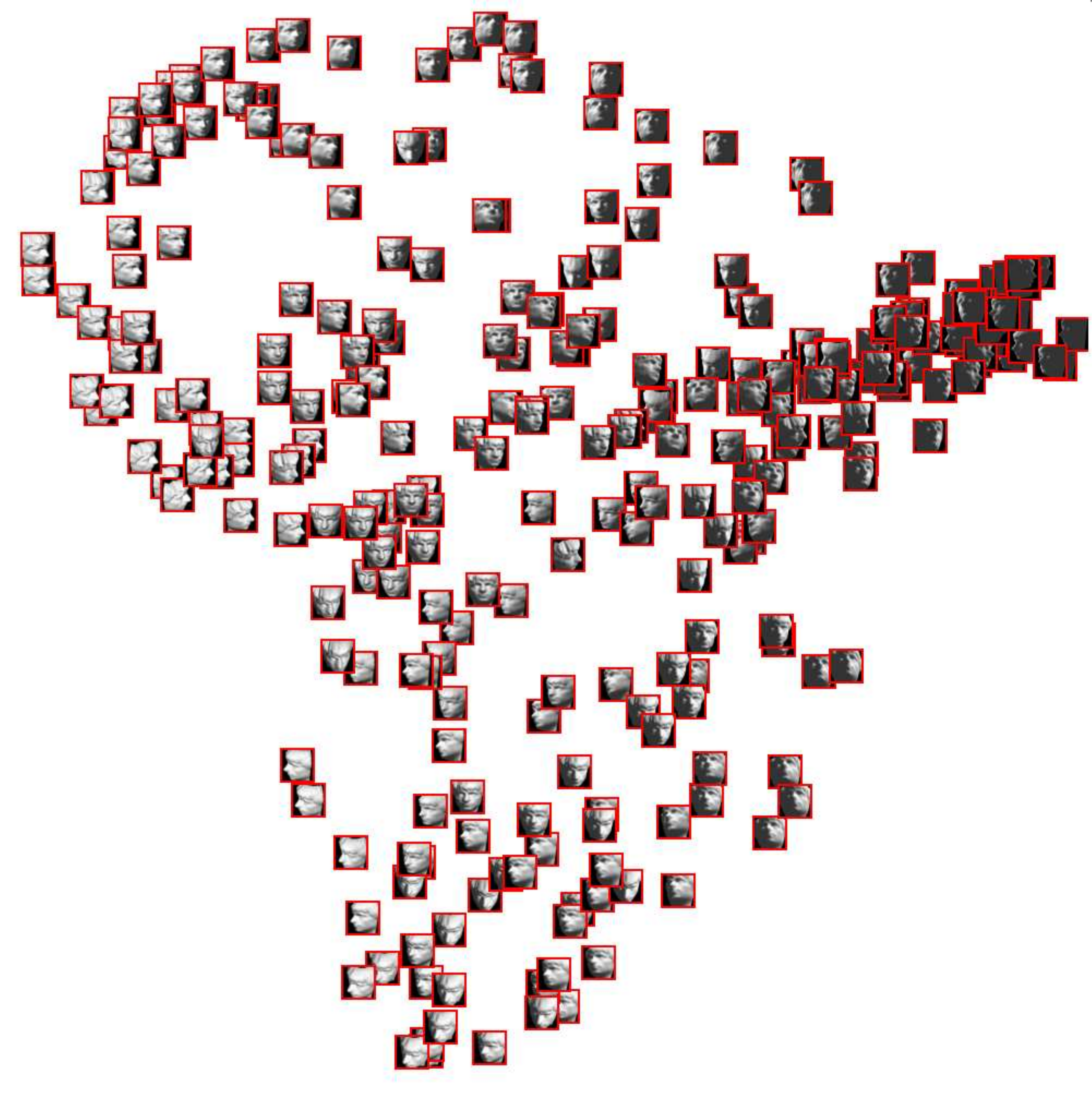}
   }
   \caption
   {
        Comparison between the DR result on the simplified data set and that
            on the original data set \emph{human face}.
        The DR results on the simplified
            data sets $\mathcal{S}_{10}$, $\mathcal{S}_{20}$, $\mathcal{S}_{50}$,
            are illustrated in (a, c, e), respectively.
        For comparison, the DR is performed on the
            original data set \emph{human face},
            and the subsets corresponding to the points in
            $\mathcal{S}_{10}$, $\mathcal{S}_{20}$, $\mathcal{S}_{50}$ are extracted and demonstrated in (b, d, f).
   }
   \label{fig:face_simplified}
\end{figure}

 First, we compare the results of DR on the simplified data sets
    and that of DR on the original data set.
 Here, the data set \emph{human face} is used as the original set
    and its simplified data sets are $\mathcal{S}_{10}, \mathcal{S}_{20}, \mathcal{S}_{50}$,
    consisting of the feature points of the first $10, 20, 50$ LBO eigenvectors, respectively.
 In Figs~\ref{subfig:face_10_simplify}, \ref{subfig:face_20_simplify},
    and~\ref{subfig:face_50_simplify},
    the DR results (by PCA) on the simplified data set $\mathcal{S}_{10}, \mathcal{S}_{20}, \mathcal{S}_{50}$ are presented, respectively.
 For comparison,
    the DR (by PCA) of the original data set \emph{human face} was performed
    and the subsets of the DR results corresponding to the points in $\mathcal{S}_{10}, \mathcal{S}_{20}, \mathcal{S}_{50}$ are extracted and demonstrated in
    Figs.~\ref{subfig:face_10_org},
    \ref{subfig:face_20_org},
    and \ref{subfig:face_50_org}.
 It can be observed that the shapes and key geometric structures
    in the DR results of the original data set
    hold in the DR results of the simplified data sets.

%

 \begin{figure}[!htb]
   \centering
   \subfigure[]
   {
        \label{subfig:face spcaout}
        \includegraphics[width=0.20\textwidth]{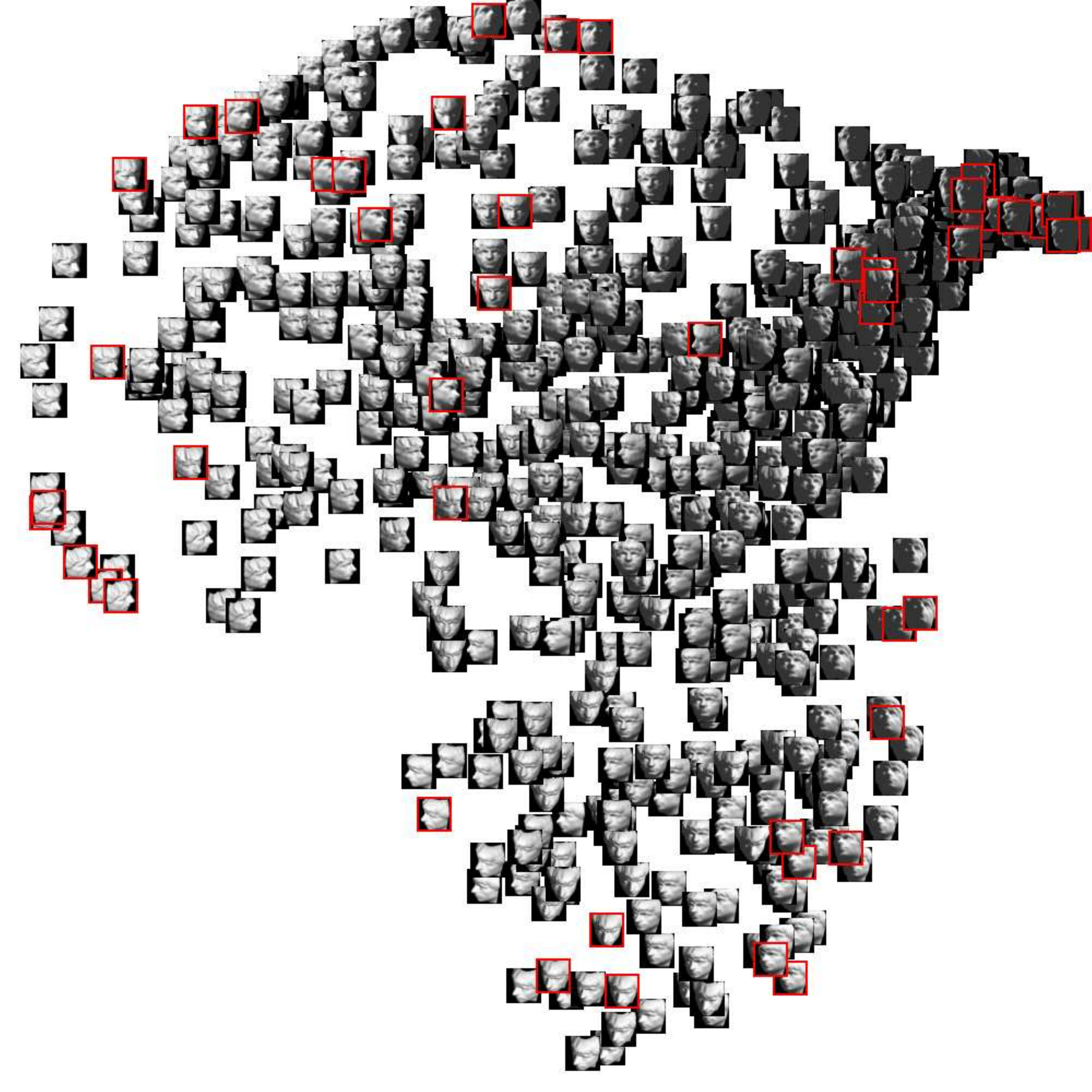}
   }
   \subfigure[]
   {
        \label{subfig:face pca}
        \includegraphics[width=0.20\textwidth]{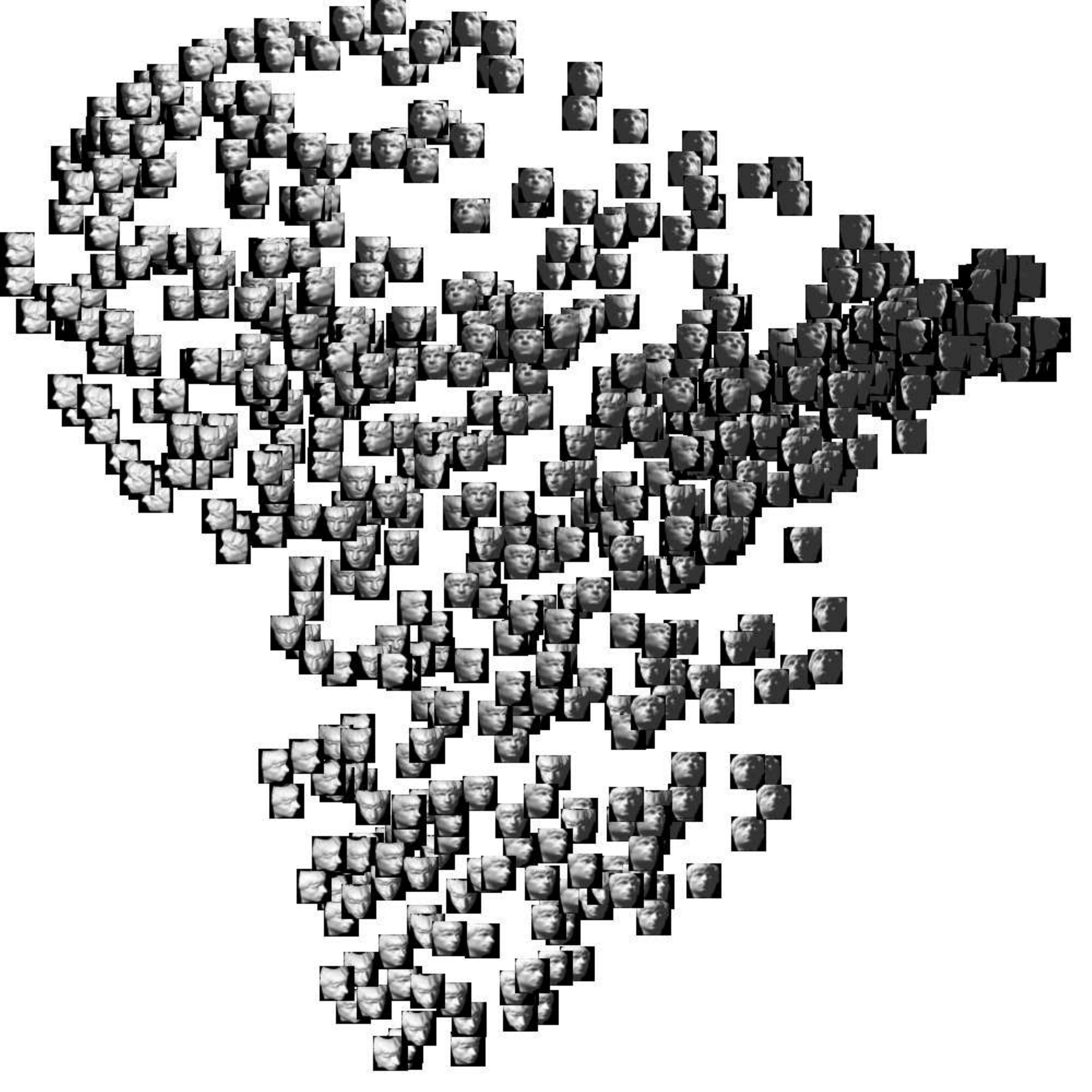}
   }
   \subfigure[]
   {
        \label{subfig:outoftable}
        \includegraphics[width=0.22\textwidth]{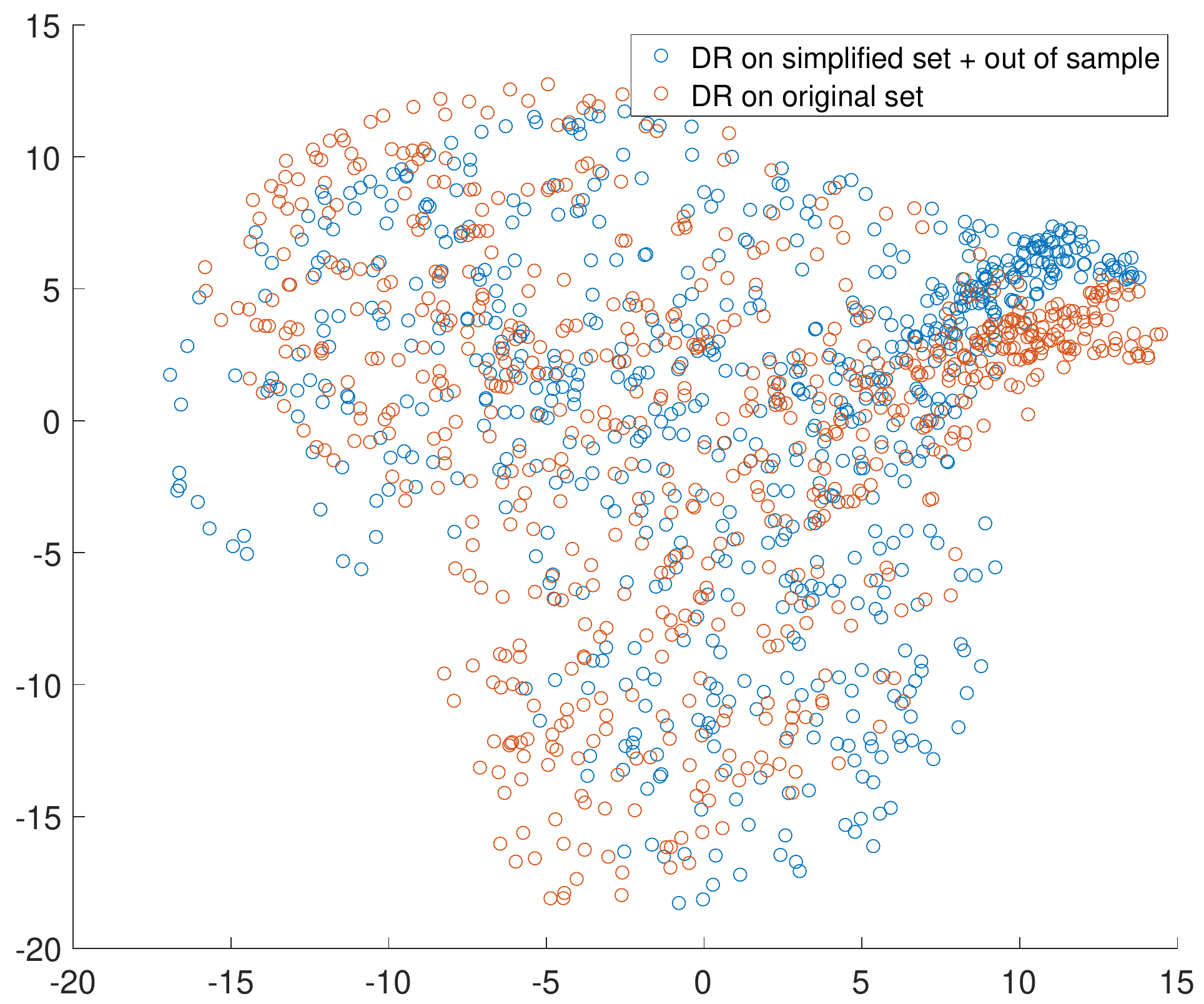}
   }
   \subfigure[]
   {
        \label{subfig:biases}
        \includegraphics[width=0.23\textwidth]{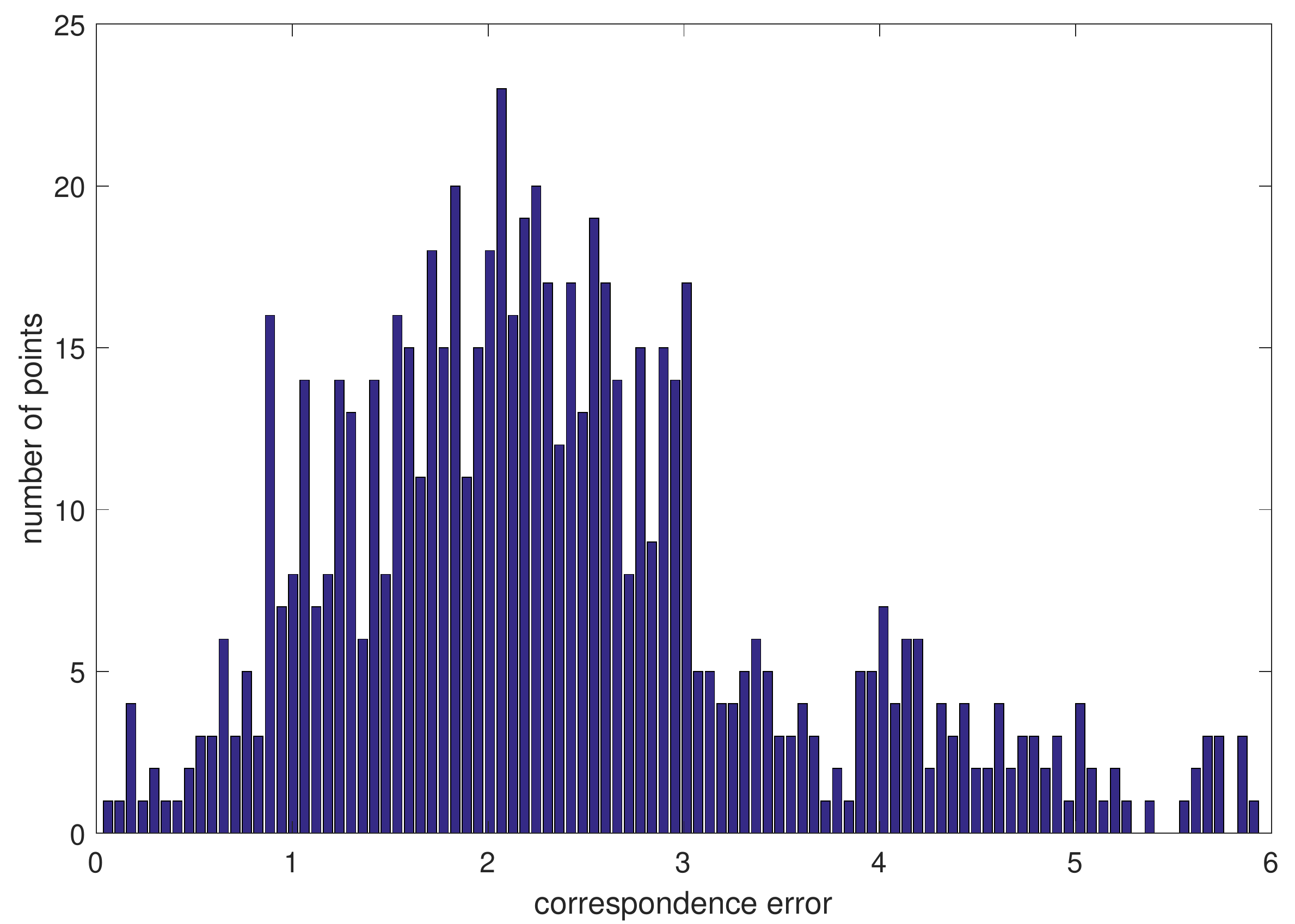}
   }
   \caption
   {
        Speedup of DR using the method of 'DR on simplified set + out of sample'.
      (a) Result generated by 'DR on simplified set + out of sample',
          where the images with red frame are the DR result on the simplified set $S_{10}$.
      (b) Result by 'DR on original set'.
      (c) The above two results are displayed in the overlapping manner.
      (d) The distribution of the correspondence error.
   }
   \label{fig:out of sample}
\end{figure}

 Beginning with the DR result on the simplified data set
    $S_{10}$ (Fig.~\ref{subfig:face_10_simplify}),
    the DR coordinates of the remaining points in the original data set \emph{human face} were calculated by the 'out-of-sample' algorithm~\cite{bengio2004out}.
 The DR result thus generated is illustrated in Fig.~\ref{subfig:face spcaout},
    where the images marked with red frames are the DR results of $S_{10}$,
    and the other image positions are calculated by the 'out-of-sample' algorithm~\cite{bengio2004out}.
 The DR result on the original set \emph{human face} is
    demonstrated in Fig.~\ref{subfig:face pca}.
 For clarity of visual comparison,
    the two results in Figs.~\ref{subfig:face spcaout} and~\ref{subfig:face pca} are displayed in an overlapping manner in
    Fig.~\ref{subfig:outoftable}.
 The correspondence error between the two DR results,
    i.e., the distance between the two DR coordinates of the same data point, were counted and
    displayed in Fig.~\ref{subfig:biases},
    where the $x$-axis is the interval of the correspondence error
    and the $y$-axis is the number of points with correspondence errors in the intervals.
 Because the simplified data set $S_{10}$ captures the feature
    points of the original data set,
    the DR result by 'DR on simplified set + out of sample' is similar to that by 'DR on original set'.
 They have a highly repeated distribution and small correspondence error.

 \begin{table}[htbp]
 \centering
 \caption
 {
    \label{tab:outofsample}
    Speedup of DR.
 }
 \begin{tabular}{ccccc}
    \toprule
    Data set         & DR (PCA)        & out-of-sample          &Total\\
    \midrule
    Human face       & 7.975           &    /                    & 7.975\\
    Simplified data  & 0.079           &    0.001               & 0.080\\
   \bottomrule
  \end{tabular}
 \end{table}

 In Table~\ref{tab:outofsample},
    the running time for the methods 'DR on simplified set + out of sample'
    and 'DR on original set' are listed.
 Whereas the method 'DR on original set' required $7.975$ seconds,
    the method 'DR on simplified set + out of sample' required only $0.080$ seconds,
    a run-time savings of two orders of magnitude.

\textbf{Training data simplification}:
  With the increase of the use of training data in supervised learning~\cite{dietterich1998approximate},
    the cost of manual annotation has become increasingly more expensive.
  Moreover, repeated samples or 'bad' samples can be added into
    the training data set,
    adding confusion and complication to the training data,
    and influencing the convergence of the training.
  Therefore, in supervised and active
    learning~\cite{johnson2008active},
    simplifying the training data set by retaining the feature
    points and deleting the repeated or 'bad' samples can
    \begin{enumerate}
        \item lighten the burden of the manual annotation,
        \item reduce the number of epochs, and,
        \item save storage space.
    \end{enumerate}



\begin{figure}[!htb]
      \centering
      \includegraphics[width=0.43\textwidth]{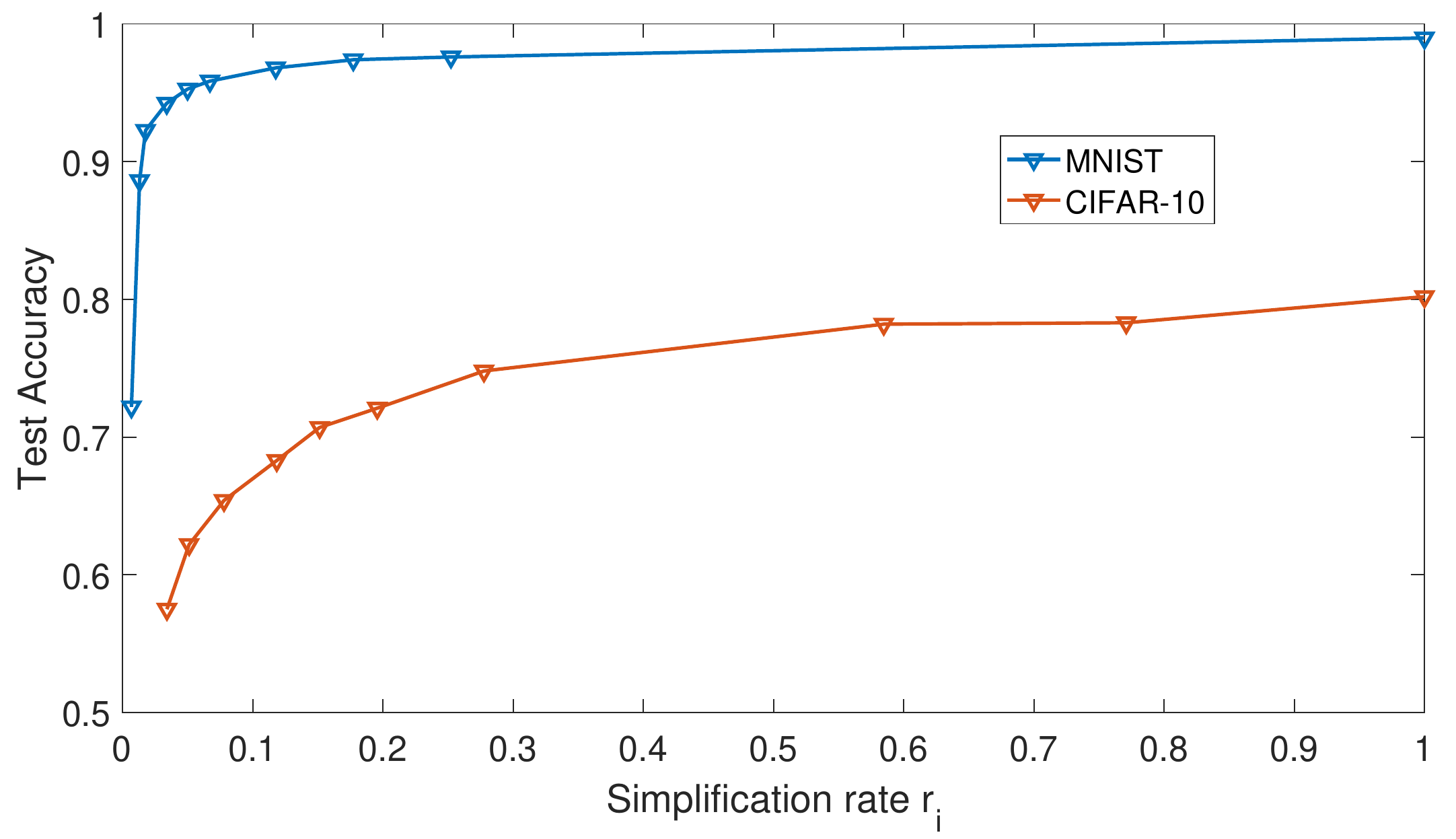}
      \caption
      {
         Diagram of simplification rate v.s. test accuracy.
      }
      \label{fig:classification}
\end{figure}

  To validate the superiority of the simplified data set in supervised
    learning,
    we constructed a CNN to perform an image classification task.
  The CNN consisted of two convolution layers,
    followed by two fully connected layers.
  \emph{Cross-entropy} was used as the loss function,
    and \emph{Adam}~\cite{kingma2014adam} was employed to minimize the loss function.

 First, we studied the relationship between the simplification
    rate~\pref{eq:ratio} and test accuracy with two experiments.
 In the first experiment,
    the data set \emph{MNIST} was employed,
    which contained $60000$ training images and $10000$ test images.
 In the second experiment,
    we used the data set \emph{CIFAR-10} with $50000$ training images
    and $10000$ test images.
 In the two experiments,
    the training image set was first simplified using
    the proposed method;
    then, the CNN was trained for $10000$ epochs using the simplified training sets.
 Finally, the test accuracy was calculated by the test images.
 In Fig.~\ref{fig:classification},
    the diagram of the simplification rate $r_i$ v.s. test accuracy is illustrated.
 For the data set \emph{MNIST},
     when $r_i$ = 0.177 (the simplified training set contained nearly $10000$ images),
     the test accuracy achieved $97.4\%$,
     which was near to the test accuracy of $98.98\%$ using the original training data ($60000$ images).
 For the data set \emph{CIFAR-10},
    when $r_i$ = 0.584 and
     the simplified training data set contained nearly $29000$ images,
     the test accuracy was $78.2\%$,
     whereas using the original training images ($50000$ images),
     the test accuracy was $80.2\%$.
 Therefore,
   with the proposed simplification method,
   the simplified data was sufficient for training in the two image classification tasks.

\begin{figure}[!htb]
      \centering
      \includegraphics[width=0.43\textwidth]{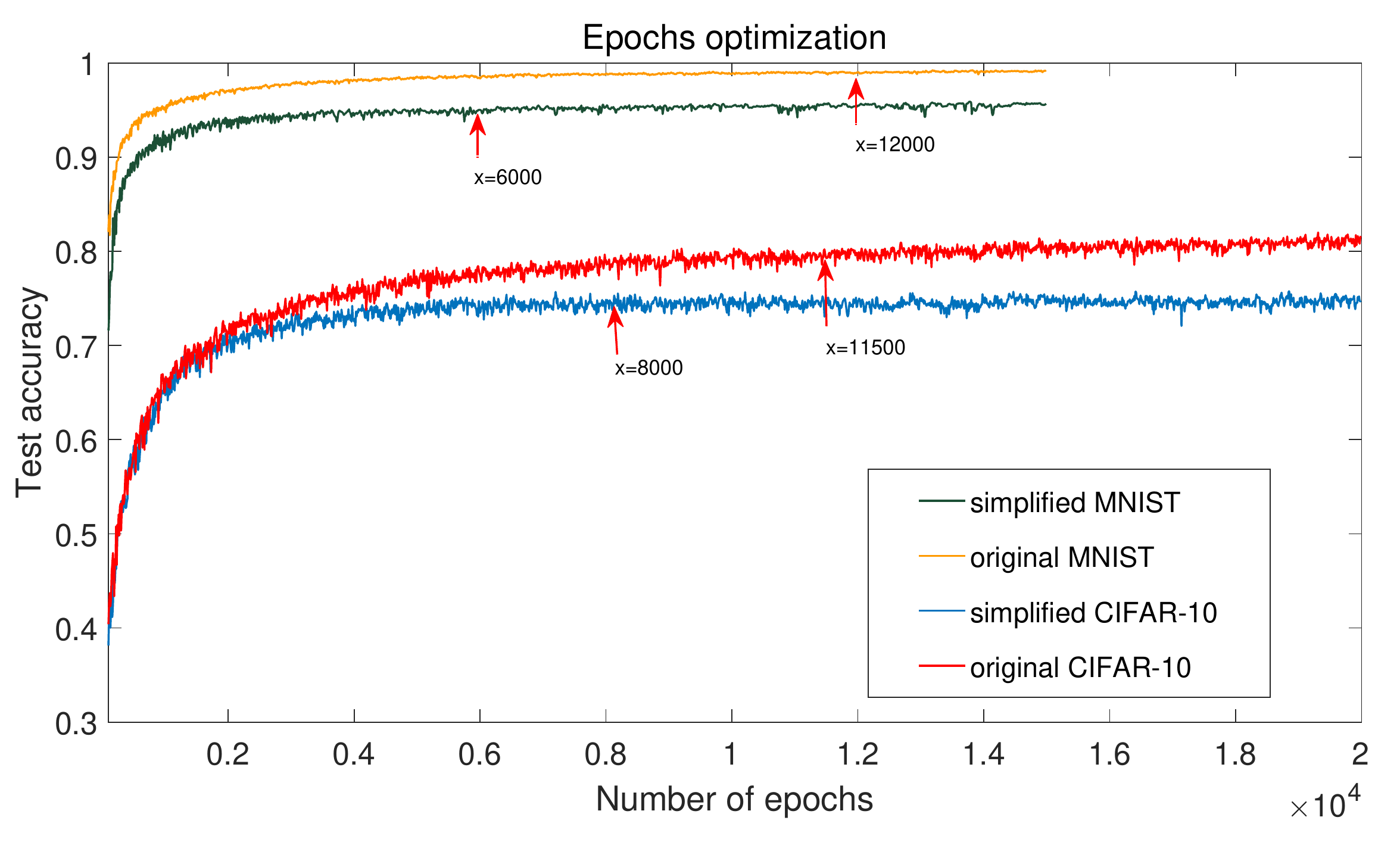}
      \caption
      {
        Diagrams of number of epochs v.s. test accuracy.
        The arrows indicate the places where the training processes stop.
      }
      \label{fig:epoch}
\end{figure}

 Furthermore, with the simplified training set,
    the training process can terminate earlier than that with the original training set,
    thus reducing the number of epochs.
 In Fig.~\ref{fig:epoch},
    the diagrams of the number of epochs v.s.
    test accuracy are illustrated.
 The 'early stopping' criterion proposed in ~\cite{prechelt1998automatic}
    was used to determine the stopping time (indicated by arrows).
 For the data set \emph{MNIST},
    the training process with the original training set (including $60000$ images) stopped at $12000$ epochs,
    with a test accuracy $99.07\%$;
    with the simplified training set (including $2991$ images),
    the training process stopped as early as at $6000$ epochs,
    with a test accuracy $95.25\%$.
 For the data set \emph{CIFAR-10},
    the training process stopped at $8000$ epochs with the simplified training set (including $10630$ images),
    whereas the training process with the original training set (including $50000$ images) stopped at $11500$ epochs.
 Therefore,
    with the simplified training data set,
    the training epochs can be reduced significantly.

\section{Conclusion}
\label{sec:conclusion}

 In this paper,
    we proposed a big data simplification method by detecting the feature points of the eigenvectors of an LBO defined on the data set.
 Specifically, we first clarified the close relationship between the feature
    points of the eigenvectors of an LBO and the scalar curvature of the manifold the LBO defined.
 Then, we proposed methods for detecting the feature points of an LBO
    defined on an unorganized high-dimensional data set.
 The simplified data set consisted of the feature points of the LBO.
 Moreover, three metrics were developed to measure the fidelity of the
    simplified data set to the original data set from three perspectives,
     i.e., information theory, geometry, and statistics.
 Finally, the proposed method was validated by simplifying some high-dimensional data sets,
 Applications of the simplified data set were demonstrated,
    including the speedup of DR
    and training data simplification.
 These demonstrated that the simplified data set can capture the main features of the
    original data set
    and can replace the original data set in data processing tasks,
    thus improving the data processing capability significantly.

%
%
%
%

\section*{Acknowledgement}

The authors would like thank to Zihao Wang for the helpful advices
and JiaMing Ai, JinCheng Lu for the help in dimensionality reduction.
This work is supported by the National Natural Science Foundation of China under Grant no. 61872316, and the National Key R\&D
Plan of China under Grant no. 2016YFB1001501.
\ifCLASSOPTIONcaptionsoff
  \newpage
\fi

\bibliographystyle{plain}

\bibliography{bibfile}

%
%
%

\end{document}